\documentclass{article}
\usepackage{preprint,times} %
\usepackage[T1]{fontenc} %
\usepackage{verbatim}
\usepackage{float} %
\usepackage{listings}
\newcommand{\wrapverb}[1]{\lstinputlisting{#1}}
\usepackage{microtype} %
\usepackage[hidelinks]{hyperref}
\usepackage{url}
\usepackage{graphicx}
\usepackage{booktabs}
\usepackage{array}
\usepackage{amsmath,amssymb}
\graphicspath{{figures/}}

\newcommand{\signpost}[1]{{\normalsize\bfseries #1}\hspace{1em}\ignorespaces}
\title{Credit Without Ground Truth:\\\mbox{Claim-Matched} Evaluation of \mbox{Step-Level}\\Credit with Finite Executed Replay}
\author{Haiyue Zhang\\
University of Southern California\quad\texttt{haiyuez@usc.edu}}

\begin{document}
\maketitle

\newcommand{\BankRank}{0.330}
\newcommand{\BankPearson}{0.736}
\newcommand{\BankMAE}{0.056}
\newcommand{\WorkerGPUHours}{28.16}

\newcommand{\RFSpearman}{0.330}
\newcommand{\RFPearson}{0.736}
\newcommand{\RFMAE}{0.056}
\newcommand{\RFSpearmanShort}{0.33}
\newcommand{\RFPearsonShort}{0.74}
\newcommand{\RFMedianWidth}{0.589}
\newcommand{\RFWorkerHours}{28.16}
\newcommand{\RFImplicitChange}{0.400}
\newcommand{\RFImplicitChangeLow}{-0.316}
\newcommand{\RFImplicitChangeHigh}{0.800}
\newcommand{\RFJudgeChange}{0.000}
\newcommand{\RFJudgeChangeLow}{-0.516}
\newcommand{\RFJudgeChangeHigh}{0.707}
\newcommand{\RFFactualGain}{8.5}
\newcommand{\RFFactualLow}{0.6}
\newcommand{\RFFactualHigh}{20.6}
\newcommand{\RFRandomGain}{3.3}
\newcommand{\RFRandomLow}{-2.1}
\newcommand{\RFRandomHigh}{13.3}

\newcommand{\PConfMean}{2.34}
\newcommand{\PConfLow}{-2.71}
\newcommand{\PConfHigh}{8.70}
\newcommand{\QConfMean}{5.94}
\newcommand{\QConfLow}{0.47}
\newcommand{\QConfHigh}{13.44}
\newcommand{\PImplMean}{2.34}
\newcommand{\PImplLow}{-2.80}
\newcommand{\PImplHigh}{8.65}
\newcommand{\QImplMean}{5.83}
\newcommand{\QImplLow}{0.36}
\newcommand{\QImplHigh}{13.28}
\newcommand{\PJudgeMean}{2.55}
\newcommand{\PJudgeLow}{-0.36}
\newcommand{\PJudgeHigh}{6.67}
\newcommand{\QJudgeMean}{3.96}
\newcommand{\QJudgeLow}{-0.21}
\newcommand{\QJudgeHigh}{9.48}
\newcommand{\IntRhoLow}{0.134}
\newcommand{\IntRhoHigh}{0.517}
\newcommand{\IntLocalAgree}{103}
\newcommand{\IntLocalDen}{150}
\newcommand{\IntTiedPairs}{108}
\newcommand{\IntLocalDefined}{29}
\newcommand{\IntLocalMedian}{0.600}
\newcommand{\SharedConf}{9}
\newcommand{\ShiftConf}{1.667}
\newcommand{\SharedImpl}{8}
\newcommand{\ShiftImpl}{1.667}
\newcommand{\SharedJudge}{9}
\newcommand{\ShiftJudge}{1.458}

\begin{abstract}
Step-level credit evaluation can conflate numerical association with ordering and branch recovery with position-selection value. We test these distinctions in ALFWorld with two independently seeded replay banks: $27{,}520$ continuations over $172$ fixed turns. At $16$ continuations per arm per bank, cross-bank replay-contrast associations are Pearson $0.736$ and pooled Spearman $0.330$. On $30$ historical failures, A-count-selected branches at confidence-/implicit-selected positions gain $8.54$ percentage points over same-position factual reruns, but their $3.33$-point advantage over frozen-random positions under the same alternative-selection rule remains unresolved. Numerical association alone does not establish reliable pooled ordering; branch recovery alone does not establish better score-informed position selection. Historical judge alignment is modestly positive with an unresolved archived-control increment. Same-cohort C/D information-source comparisons and a short, sparse-exposure training pilot provide scope checks; incremental hindsight training utility remains unestablished. Findings are conditional on the tested policies, intervention support and finite budgets.
\end{abstract}

\begin{figure}[H]
\centering
\includegraphics[width=\linewidth]{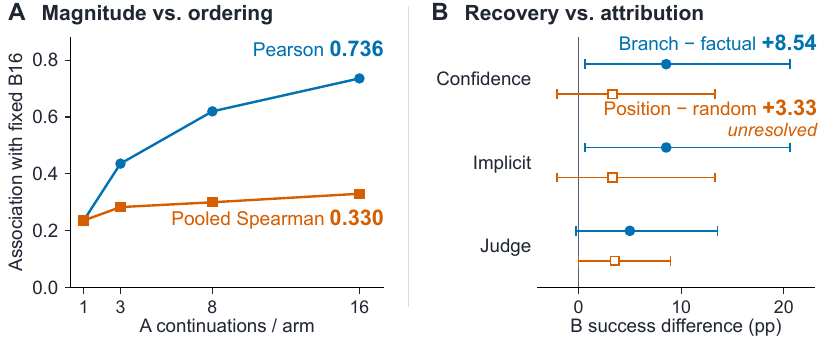}
\caption{Claim-matched checks distinguish numerical association from pooled ordering, and branch recovery from position-selection value. (A) Nested A depth versus fixed B16 on $172$ turns: Pearson reaches $0.736$, pooled Spearman $0.330$. (B) On $30$ failures, confidence/implicit gain $8.54$ pp over same-position factual reruns but $3.33$ pp over frozen-random positions, unresolved. Bars use the seven-comparison adjusted family in Figure~\ref{fig:recovery}; matching selector results are not independent confirmations.}
\label{fig:postsummary}
\end{figure}
\section{Introduction}\label{sec:intro}
Step-credit methods are credited with downstream improvements \citep{tarl2509,hcapo2603,ccpo2603}, but which decision did the score improve? Our audit independently remeasures finite replay and tests recovery-position selection against a frozen random baseline (Figure~\ref{fig:postsummary}).

This \textbf{claim-matched audit} finds stronger numerical than ordering agreement, and supported branch recovery without established location-selection value. Observed alignment and a separate training test define how far the evidence extends.

Executed replay restores a prefix, intervenes on an action and samples a frozen-policy continuation. It supplies a reference where step-level ground truth is unavailable, but replication budget, alternative support and ties affect that finite measurement. An audit must examine the reference before using it as a ruler for scores, then test the downstream choice directly.

Our ALFWorld study \citep{alfworld2021} makes \textbf{two core empirical contributions}:
\begin{enumerate}\renewcommand{\labelenumi}{\Alph{enumi}.}\setlength{\itemsep}{3pt}
\item \textbf{Independent replay separates magnitude from ordering.} Two independently seeded banks execute $27{,}520$ continuations on $172$ fixed turns. At $16$ continuations per arm, Pearson reaches $\mathbf{0.736}$, but pooled Spearman reaches $\mathbf{0.330}$. Independent ordering diagnostics materially qualify the numerical result (Section~\ref{sec:independent}; Figure~\ref{fig:postsummary}A).
\item \textbf{Recovery comparators separate branch and position claims.} On $30$ failures, confidence-/implicit-selected branches improve success by $\mathbf{8.54}$ percentage points over same-position factual reruns. Their $\mathbf{3.33}$-point advantage over frozen-random positions remains unresolved under the same branch rule. The comparator changes what the positive recovery result establishes (Section~\ref{sec:recovery}; Figures~\ref{fig:postsummary}B and~\ref{fig:recovery}).
\end{enumerate}
\textbf{Supporting boundary tests.} Historical judge alignment is modestly positive while the archived-control increment remains unresolved (Section~\ref{sec:signalinterpretation}). Fresh C/D evaluation under broader eligible position support leaves count and hindsight increments unresolved on the same historical failures (Section~\ref{sec:recovery}). A separate learned-policy F/N bridge has small, sign-changing differences and sparse realized exposure; it does not justify scaling that development comparison (Section~\ref{sec:trainingbridge}). These tests bound the interpretation of the two core results.

In this audit, the same system looks materially more convincing under partial comparisons than under comparisons matched to the claim being made.

The contribution is an empirical validation pattern for the tested reference and recovery decisions, rather than a new credit algorithm or a universal claim that hindsight credit fails. In this ALFWorld audit, numerical replay agreement and branch recovery gains do not establish reliable pooled ordering or better score-informed position selection.

\section{Validation targets in this step-credit audit}
\label{sec:auditsetup}
\signpost{Reference: a conditional intervention contrast.}
In ALFWorld \citep{alfworld2021,textworld2018}, restore history $h_t$ before an action, execute either factual action $a_t^f$ or an alternative, then continue under the same frozen policy $\pi$. If $p_t(a)$ is terminal-success probability after that intervention, the target for four sampled alternatives $S_t=\{a_{t,1},\ldots,a_{t,4}\}$ is
\begin{equation}
A_t(S_t)=p_t(a_t^f)-\frac14\sum_{k=1}^{4}p_t(a_{t,k}).
\label{eq:reframe_estimand}
\end{equation}
The factual action is re-executed rather than assigned its original outcome. Equal replication estimates a uniform mean over these four alternatives, not all possible actions. The policy is Qwen2.5-7B-Instruct \citep{qwen25}, with the original ALFWorld template and temperature $0.7$. Support stays fixed as depth increases. Historical turns lacking four distinct admissible alternatives were counted as undefined; a sampled zero is a finite-resolution observation, not a latent no-effect label.

\signpost{Signal: alignment with a measured ordering.}
We audit implicit credit adapted from Hindsight Credit Assignment Policy Optimization (HCAPO), using normalized and clipped hindsight-conditioned action likelihood \citep{hcapo2603}, and a Turn-level Adjudicated Reinforcement Learning (TARL) adapted Qwen2.5-72B judge with scores in $\{-1,0,1\}$ \citep{tarl2509}. Exact formulas, prompts and adaptations appear in Appendices~\ref{sec:method} and~\ref{app:instrument}. Alignment is within-trajectory Spearman against the sampled replay contrast, summarized over computable trajectories. Constant vectors yield undefined correlations, not zeros. Scores retain their original normalization when the replay mask is applied.

\signpost{Decision: the value of choosing a recovery position.}
Recovery uses archived scores, including hindsight, after a failure. The score chooses one of four positions; Bank A success counts choose one alternative there; independently seeded Bank B evaluates the frozen branch. This division separates the score's location choice from the action-selection rule. On the same B16 position and support, let $F$, $M$ and $S$ denote factual, mean-alternative and A-chosen success rates. Then $\widehat A^B=F-M$ and $Q=S-M$ imply $S-F=Q-\widehat A^B$. This accounting connects replay credit to branch gain, but a gain can come from the chosen alternative without establishing a better position selector.

The comparisons preserve the information available to the procedure: this is retrospective recovery, not online selection before an outcome is observed. Rank fidelity, calibrated magnitude, step correctness, recovery value and training utility remain distinct targets. A history-dependent additive baseline can preserve an expected policy gradient while changing score order \citep{sutton1999policy}; useful learning therefore need not require the ordering interpretation audited here.

\section{Can the replay reference reproduce itself?}
\label{sec:independent}
To test replay as a ranking reference, after four development and three structural exclusions we froze $43$ trajectories with four eligible position-quartile midpoints each. Factual actions and four archived alternatives receive $16$ continuations per bank:
$43\times4\times5\times16\times2=27{,}520$. Banks share $172$ turns, actions and policy; their continuation seeds are independent and neither is noiseless. Bank A fixes primary recovery choices before B collection. Runtime checks are in Appendix~\ref{app:postruntime}.

\begin{figure}[ht]
\centering\includegraphics[width=\linewidth]{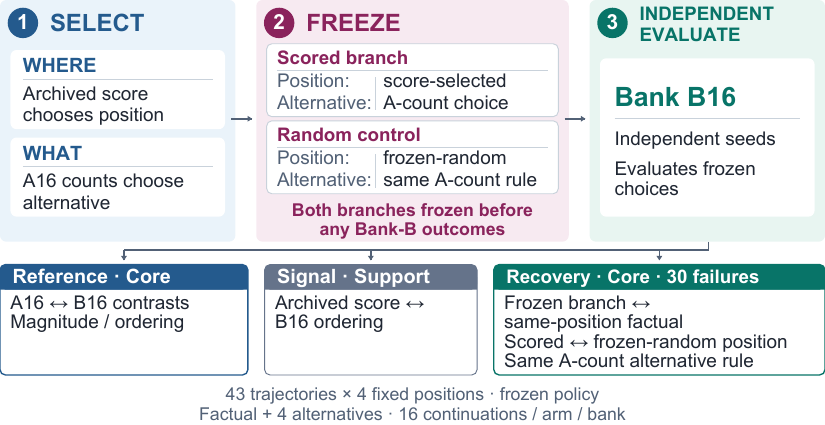}
\caption{Archived scores choose positions; A16 counts choose alternatives there and at frozen-random positions using the same rule. Both branches are frozen before B outcomes. A16 and independent B16 estimate reference contrasts; B16 also measures deeper Signal alignment and recovery. Historical Signal controls and the learned-policy pilot are separate.}
\label{fig:bankdesign}\label{fig:referenceorder}\label{tab:bankagreement}
\end{figure}

\subsection{Numerical agreement does not settle the ordering question}
At nested A depths $1,3,8,16$, Pearson association with B16 is $0.236,0.436,0.620,0.736$, whereas pooled Spearman is $0.236,0.283,0.300,0.330$ (Figure~\ref{fig:postsummary}A). Mean absolute error falls from $0.136$ to $0.056$. More replication makes numerical association and absolute error substantially better, while the pooled ordering diagnostic improves much less. These comparisons reuse B16 and nested A samples; the four depths are not independent experiments.

A ranking reference is used to distinguish which steps changed the outcome more. Pearson measures linear association and MAE measures absolute error; neither alone answers how consistently turns are ordered. The lower Spearman qualifies the ordering interpretation without erasing numerical improvement. Pearson is not a calibration test, and neither finite bank identifies latent effects.

Independent A/B remeasurement therefore tests the reference before either signal is compared against it. The local ordering analysis below qualifies the pooled diagnosis.

\subsection{Uncertainty and local support qualify the separation}
The post-hoc trajectory-cluster interval for pooled Spearman is $[0.134,0.517]$, versus Pearson's $[0.554,0.835]$. The paired Pearson--Spearman gap is $0.406$ $[0.202,0.601]$; Kendall $\tau_b=0.269$ $[0.096,0.438]$ retains the direction. The Spearman interval spans the frozen study-specific $0.5$ criterion. The data support a marked separation between the two diagnostics, not a definitive threshold failure or instability at arbitrarily large replay budgets.

Within trajectories, $103/150$ non-tied pairs agree, while $108/258$ pairs tie in at least one bank. Median within-trajectory Spearman is $0.600$ $[0.316,0.800]$ on $29/43$ computable trajectories, leaving $14$ undefined (Appendix~\ref{app:p0rank}). This stronger local agreement prevents a blanket claim of rank failure. It also shows why the population belongs beside the result: dropping tied or constant cases changes what the summary describes. Pooled agreement across $172$ turns is not the reliability of historical within-trajectory credit fidelity.

Individual effects remain weakly resolved: five-arm bounds exclude zero for $3/172$ B16 contrasts, with median width $\RFMedianWidth$ (simultaneous within a turn, not across turns). Historical-zero transitions and complete resolution summaries remain in Appendix~\ref{app:postprecision}. Ordering, ties and uncertainty therefore accompany claims made against the finite reference.

\section{What do the audited credit signals track?}\label{sec:signalinterpretation}
\begin{figure}[ht]
\centering\includegraphics[width=\linewidth]{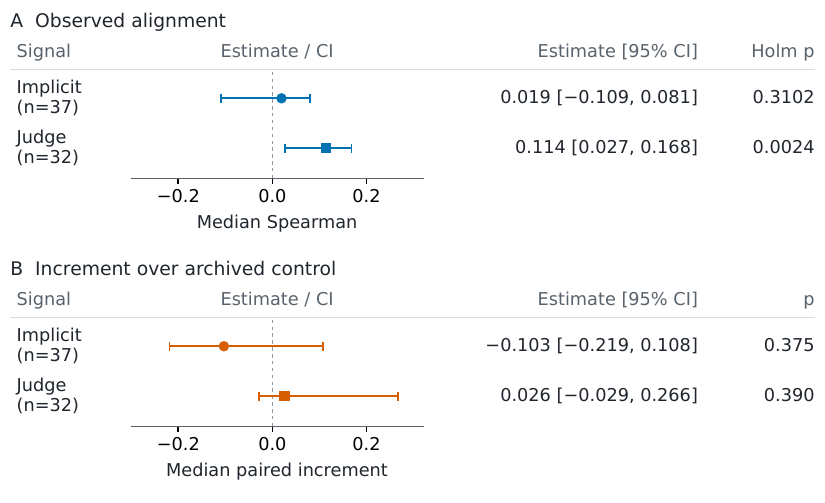}
\caption{On the judge's paired population, positive observed alignment coexists with an unresolved archived-control increment. (A) Median observed within-trajectory Spearman: descriptive $95\%$ trajectory-bootstrap intervals; Holm-adjusted positive permutation $p$. (B) Median paired increment over one archived full-domain shuffle: control-conditional $95\%$ intervals; paired sign-flip $p$. Populations differ; this is not a test of signal superiority.}
\label{fig:signal}
\end{figure}

The Signal layer asks whether a step score aligns with the measured replay ordering, and what that alignment adds over a specified control. These questions use the same signal-specific historical paired populations but different statistical objects. Figure~\ref{fig:signal} stacks the two comparisons so that the affirmative judge result and unresolved incremental result remain visible together.

\subsection{Observed alignment: is there a positive association?}
The historical audit uses three replay replicas per arm. Its post-hoc fixed-mask analysis holds replay eligibility and finite paired values fixed, then permutes retained scores without renormalizing them. If $r_i$ is the observed score--replay Spearman correlation for trajectory $i$, the summary is $T_{\rm obs}=\operatorname{median}_i r_i$. This tests alignment on the retained population, rather than calibration to a known latent contribution.

The adapted judge has median $0.114$, descriptive $95\%$ interval $[0.027,0.168]$, $n=32$, and Holm-adjusted positive permutation $p=0.0024$. Implicit credit has median $0.019$ $[-0.109,0.081]$, $n=37$, and adjusted $p\simeq0.310$. The test uses $10{,}000$ permutations per signal. The judge result is modestly positive and should be retained as affirmative evidence; a nondetection for implicit credit does not establish that it contains no contribution information. Significance for one signal and nondetection for another do not test their difference.

\subsection{Increment: what changes relative to the archived control?}
The archived paired statistic is instead
$D_{\rm arch}=\operatorname{median}_i(r_i-s_{i,\rm arch})$,
where $s_{i,\rm arch}$ is the correlation under one archived full-score-domain shuffle applied before replay masking. The implicit increment is $-0.103$ $[-0.219,0.108]$ and the judge increment is $0.026$ $[-0.029,0.266]$, with two-sided paired sign-flip $p=0.375$ and $0.390$. Neither paired increment is resolved; neither interval establishes equivalence to the control.

The two panels do not contradict each other. A median observed correlation asks about alignment itself; a median paired difference asks what remains after each trajectory's archived control is subtracted. In general, the median of paired differences is not a difference of medians. The intervals also condition differently: panel B holds one control realization fixed, whereas panel A's permutation test compares against a distribution of retained-score rearrangements. Treating these readings as interchangeable would change the claim being tested.

Their difference cannot be assigned to mask order alone. Permutation domain, statistic, null, and one-control versus distributional conditioning all differ. With unique scores and a fixed mask, uniform full-domain and retained-score shuffles yield the same retained-rank distribution; ties can alter that distribution and its undefined mass. The fixed-mask test assumes score assignments are exchangeable across retained positions. Alignment can reflect action type or trajectory stage, and fixing the mask does not validate that assumption. Full controls and uncertainty definitions are retained in Appendix~\ref{app:signaldetail}.

\subsection{Deeper replay retains a conditional population}
At B16, alignment is $0.200$ $[-0.316,0.600]$ for implicit credit on $23/43$ trajectories and $0.775$ $[0,0.775]$ for the judge on $13/43$. Membership is fixed across depths but differs by signal; ties and constant vectors restrict computability. These two-signal-adjusted intervals do not compare signal quality. Full depth curves, paired changes and the distinct historical resolution population remain in Appendix~\ref{app:signaldetail}.

These results support a limited positive alignment finding; decision value remains a separate target. Replay credit averages four alternatives, whereas recovery picks one alternative at one position. Agreement with the first ordering cannot by itself estimate the success of the second choice. We therefore evaluate that decision directly.

\section{Does credit improve a recovery decision?}
\label{sec:recovery}
A recovery procedure makes two decisions: where to intervene and which alternative to execute there. The score chooses the position; the A-count rule chooses the alternative. We evaluate this division through two comparators. A same-position factual rerun asks whether changing the action recovers. A frozen-random position under the same branch rule asks whether the score improves location choice. The central attribution question concerns this second comparison, even when the first is positive: finding a useful alternative does not by itself identify the value of the score-selected position.

On all $30$ historical failures, four selectors share positions, actions and budget:
minimum policy mean-token log-probability (\emph{confidence}), minimum implicit
credit, minimum judge credit, and identifier-hashed random position.

At each selected position, A chooses the alternative with greatest success count,
breaking ties by smallest frozen index; primary choices precede B collection. Independent B
outcomes compare the branch with its \textbf{same-position factual rerun}, then
compare success across \textbf{scored and frozen-random positions} under the same
rule. The latter uses success rates, not factual-adjusted gains. Random is one
identifier-hashed realization, not an average over random-selector seeds.

\begin{figure}[ht]
\centering\includegraphics[width=\linewidth]{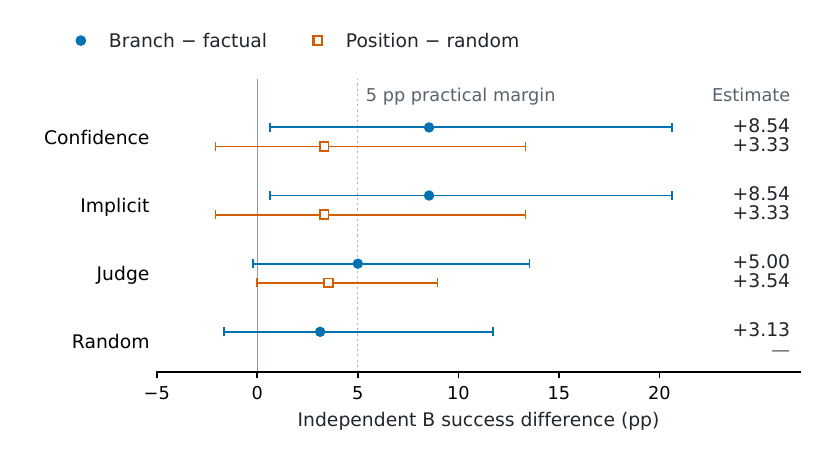}
\caption{Recovery under two comparators on $30$ failures. Filled markers compare selected branches with same-position factual reruns; open markers compare scored with frozen-random positions under the same branch rule. All seven contrasts use the frozen Bonferroni paired-bootstrap intervals ($10{,}000$ draws). The dashed line marks the $5$ pp practical margin. Confidence/implicit positions coincide on $23/30$ failures.}
\label{tab:recovery}\label{fig:recovery}
\end{figure}

\subsection{Two comparators isolate different claims}
\signpost{Can the frozen procedure find a branch that recovers?}
Confidence and implicit selectors each gain $8.54$ pp $[0.625,20.625]$ over their same-position factual reruns. This comparison holds position fixed and changes the executed action to the A-selected alternative. It supports the recovery value of the selected branch. A factual rerun is an essential comparator because the original trajectory's failure is not the success probability of repeating its action under a stochastic continuation.

\signpost{Did the score improve the location-selection decision?}
Under the same A-count branch rule, confidence and implicit success exceed frozen random-position success by $3.33$ pp $[-2.083,13.333]$. Judge's contrast is $3.54$ pp $[0,8.958]$. None resolves superiority; all leave positive gains compatible with the data. This comparison changes the position selector while retaining the branch rule, so it addresses the component attributed to the score. It compares branch success rates, not the difference between two factual-adjusted recovery gains.

The first result can therefore be positive while the second remains unresolved. The A-count rule can find a useful alternative at a position that a random selector can also reach; factual improvement alone cannot distinguish that possibility from valuable score-informed location choice. Confidence and implicit choose the same position on $23/30$ failures, so their matching aggregate results are not independent confirmations. Figure~\ref{fig:recovery} preserves both rows while making that dependence explicit.

\signpost{Which added information changes recovery?}
A matched ablation removes the outcome/final-observation hindsight bundle (H-off), or replaces count-based alternative choice with original default order. H-off retains its own full-trajectory normalization, clipping and ties; default retains the policy-generated candidate order. The two conditional effects hold the other input on.

\begin{table}[ht]
\centering\small
\begin{tabular}{@{}lcc@{}}
\toprule
Position support / outcome banks & Hindsight increment $\Delta_H$ & Count increment $\Delta_A$ \\
\midrule
Four archived positions, A/B & $0.000\;[-0.625,0.625]$ & $+7.500\;[0.833,16.667]$ \\
\begin{tabular}[c]{@{}l@{}}Full eligible position-choice\\support, C/D\end{tabular} & $-0.833\;[-5.000,3.125]$ & $+0.833\;[-1.250,3.750]$ \\
\bottomrule
\end{tabular}
\caption{Information-source increments on the same $30$ historical failures (pp [adjusted intervals]). Each study has a separate two-comparison paired-bootstrap family. A/B reuses archived outcomes post hoc; C/D uses fresh selection and independent evaluation after a prospective design freeze. The symbol $\Delta_A$ denotes the count switch, using C counts in row 2.}
\label{tab:informationablation}
\end{table}

The post-hoc four-position result supports consulting A counts over default order. The fresh study selects positions from the $1{,}324$ common eligible archived positions, selects alternatives with C16, then evaluates frozen choices with independent D16. Both fresh increments remain unresolved (Table~\ref{tab:informationablation}). New continuations remeasure the same cohort; they are not new-task replication. Position support, selection bank and evaluation bank change together, preventing attribution of the difference to support alone or a replication-failure claim. Neither study establishes equivalence, licenses removing hindsight, or tests which input is more valuable. Full cells, execution and uncertainty appear in Appendices~\ref{app:informationablation} and~\ref{app:freshcd}.

The earlier post-hoc $P$ (versus uniform positions) remains unresolved; $Q$
(versus uniform alternatives at the chosen position) supports the complete rule
for confidence/implicit, not judge (Appendices~\ref{app:exactuniform}--\ref{app:p0recovery}).

\section{A separate test of the training bridge}\label{sec:trainingbridge}
To test whether added hindsight information changes learning beyond recovery, we continued a public trained Qwen2.5-1.5B checkpoint for six updates with full hindsight (F) or matched scoring with hindsight fields withheld (N). Both share the micro-credit machinery. F adds terminal task/outcome/final-observation information and the next observation; N substitutes fixed unavailable-field strings. This independent learned-regime implementation differs from the frozen $7$B recovery policy and is not a native HCAPO reproduction \citep{hcapo2603}.

\begin{table}[ht]\centering
\begin{tabular}{lrrr}
\toprule
Same $128$ evaluation tasks & $t0$ & $t3$ & $t6$\\
\midrule
F: full hindsight & $77$ & $72$ & $70$\\
N: fields withheld & $77$ & $70$ & $72$\\
\textbf{F--N (pp)} & $0$ & $\mathbf{+1.56}$ & $\mathbf{-1.56}$\\
\midrule
\multicolumn{4}{l}{F realized exposure: \textbf{$7/96$ reward-variable groups}}\\
\bottomrule
\end{tabular}
\caption{Short learned-policy F/N bridge and realized exposure. Success counts at baseline and after three/six updates. F--N change contrasts equal raw differences because baseline outcomes coincide task by task. Exposure counts reward-variable groups in the realized F continuation.}\label{tab:trainingbridge}
\end{table}

The F--N change contrasts are $+1.56$ $[-6.25,9.38]$ pp at $t3$ and $-1.56$ $[-10.16,7.03]$ at $t6$ (post-hoc paired-task $95\%$ bootstrap). The predeclared replication criterion requires at least $5$ pp absolute change contrast at $t6$ and consistent nonzero raw signs; it was not met. The development comparison therefore did not justify scaling on evidence of incremental hindsight utility. Full protocol and execution checks are in Appendix~\ref{app:trainingbridge}.

Realized exposure limits the interpretation: F has only $7/96$ reward-variable groups, $1{,}121/24{,}597$ eligible turns ($4.56\%$), and $2{,}238/24{,}597$ turns with nonzero micro-credit ($9.10\%$). Eligibility and nonzero credit are different quantities. One continuation seed and six updates leave meaningful effects unresolved; this is a training boundary at the tested horizon and exposure, not equivalence or a general null effect of hindsight.

\section{Relation to replay-based credit audits}
\label{sec:related}
We build on replay-based credit auditing to measure the finite reference's independent reproducibility and the recovery-position value attributed to a score. Source-specific comparisons and scope boundaries appear below and in Appendix~\ref{app:concurrent}.

\signpost{Known causal targets and finite measurements.}
Causal Agent Replay (CAR) formalizes per-step interventions with re-execution and validates its estimator on synthetic structural causal models with planted effects \citep{car2606}. Our ALFWorld audit has no planted step-level truth. Independent banks instead measure how reproducibly finite replay orders the same fixed positions. Correct execution and a well-defined intervention target do not determine the resolution achieved by a finite allocation of replicas; neither bank becomes an oracle.

\signpost{Credit ranking and recovery decisions.}
Contextual Counterfactual Credit Assignment (C3) defines replay-based diagnostics for cooperative LLM protocols, including fidelity between assigned advantages and a reference advantage under a fixed continuation distribution \citep{c3exact2603}. We additionally remeasure the finite reference and carry frozen score-derived position choices into untouched Bank-B evaluation. This is independent continuation measurement on shared experimental objects, not external replication across tasks or policies. CARL relates resampled-continuation criticality to entropy and search expansion in multi-hop search, and already includes a random-fork training ablation \citep{carl2512}. Our factual and random-position comparisons isolate different claims about a recovery procedure; they do not evaluate CARL's search policy or generalize to its task regime.

\signpost{Scope relative to credit methods.}
HCAPO and TARL motivate the adapted instruments \citep{hcapo2603,tarl2509}. HCAPO already ablates hindsight weight in training; our recovery ablation and learned-policy pilot use adapted implementations and do not test its native update. Step correctness, policy-reference scoring and training outcomes address different targets (Appendices~\ref{sec:method} and~\ref{app:scope}). Our retrospective recovery can use archived hindsight, whereas an online selector must act before observing the outcome. The replay-hygiene and replica-noise constructs of \citet{survey2604} supply methodological context. The contribution is empirical: using existing scores and statistical tools, claim-matched comparisons change the reading of apparently positive results in this system.

\section{Discussion}\label{sec:postdiscussion}
\subsection{The comparator determines the claim tested}
Numerical agreement, observed judge alignment and branch recovery are positive findings. Table~\ref{tab:claimsynthesis} summarizes the central empirical contribution: apparently affirmative evidence can leave the stronger claim unresolved once the comparison is matched to that claim. Their interpretation changes when ordering, incremental signal evidence and position-selection value are measured separately.

Matched checks materially qualify the reading of positive results under shared experimental objects, giving established statistical distinctions a concrete empirical consequence. The historical Signal analyses and separate learned-policy pilot delimit how far that mainline extends.

\subsection{Consequences for evaluation and practice}
\signpost{If replay supplies rankings, remeasure the ordering.}
An independent numerical association is insufficient evidence for using replay as a ranking reference. Report independent rank agreement beside absolute error, together with depth, fixed alternative support, ties and undefined cases. Our stronger local agreement is useful, but applies to a computable subset; pooled and within-trajectory summaries describe different ordering targets. The practical response is to make the reference's measurement limits visible before evaluating a score against it, rather than treating a finite replay bank as ground truth.

\begin{table}[t]
\centering\small
\begin{tabular}{@{}>{\raggedright\arraybackslash}p{.14\linewidth}>{\raggedright\arraybackslash}p{.235\linewidth}>{\raggedright\arraybackslash}p{.285\linewidth}>{\raggedright\arraybackslash}p{.24\linewidth}@{}}
\toprule
\textbf{Stage} & \textbf{Partial evidence} & \textbf{Claim-matched check} & \textbf{Supported reading} \\
\midrule
\multicolumn{4}{@{}l}{\textsc{Core empirical results}} \\
\addlinespace[5pt]
\textbf{\textsc{Reference}} & Pearson $\mathbf{0.736}$; MAE $0.056$ & Independent A16--B16 pooled ordering & Spearman $\mathbf{0.330}$; local agreement conditional \\
\addlinespace[9pt]
\textbf{\textsc{Decision}} & $\mathbf{+8.54}$ pp versus same-position factual reruns & Same branch rule: scored versus frozen-random positions & $\mathbf{+3.33}$ pp; position advantage unresolved \\
\addlinespace[4pt]
\midrule
\multicolumn{4}{@{}l}{\emph{Supporting boundary tests}} \\
\addlinespace[2pt]
Information source (C/D) & A/B count increment: $+7.50$ pp & Full-support position choice; C selects, D evaluates frozen endpoints & $+0.83$ pp, unresolved; same historical cohort \\
\addlinespace
Training & Full-hindsight F arm & F/N continuation plus realized exposure & $+1.56\to-1.56$ pp; $7/96$ variable F groups; gate not met \\
\bottomrule
\end{tabular}
\caption{Two core comparator contrasts and their boundary tests. Unresolved comparisons do not establish equivalence. C/D changes support and both banks together; it does not isolate a support effect.}
\label{tab:claimsynthesis}
\end{table}

\signpost{If a score is credited with recovery, hold the branch rule fixed.}
Same-position factual reruns establish whether the selected alternative improves success there. To attribute improvement to location choice, compare scored and baseline positions under the same action rule and budget. A deployment decision could still value the complete procedure, but its recovery gain does not establish the score's position-selection advantage. The unresolved contrast permits meaningful gains and does not make the score dispensable.

\signpost{If the claim concerns learning, report realized exposure.}
A recovery branch never updates the policy. The learned-policy pilot directly tests that further target but does not meet its replication gate. Report reward-variable groups and nonzero micro-credit alongside nominal treatment assignment: these describe the opportunity for hindsight to affect learning. Sparse exposure limits interpretation; it is not a demonstrated cause of the sign-changing outcomes.

The fresh C/D intervals are conditional on the same historical cohort and realized C selection; new-task validation and repeated-selection uncertainty remain open. Historical training diagnostics remain separately inconclusive (Appendix~\ref{app:historicalcontext}). The replay-cost frontier is secondary and shares allocations and evaluation outcomes across depths; it does not identify an optimal budget or general compute efficiency (Appendix~\ref{app:postfrontier}).

\section{Conclusion and limitations}\label{sec:limitations}
The two core results contrast Pearson $0.736$ with pooled Spearman $0.330$ in independent replay, and $+8.54$ pp over factual reruns with an unresolved $+3.33$ pp over frozen random positions in independent recovery evaluation. Each positive comparison should support the claim it measures, with the decision baseline required by the stronger claim reported alongside it. In this ALFWorld audit, numerical replay agreement and branch recovery gains do not establish reliable pooled ordering or better score-informed position selection. Modest observed judge alignment coexists with an unresolved archived increment. Fresh C/D remeasurement with full position-choice support leaves the count and hindsight increments unresolved; a separate short training comparison leaves incremental hindsight utility unestablished.

The frozen-policy scope is one $7$B policy, one environment and selected historical prefixes. Finite B16 is not an oracle, and stronger conditional local agreement qualifies blanket rank-failure readings. Signal populations differ and exchangeability is assumed; confidence/implicit recovery selections overlap. The reconstructed runtime is not bitwise replay of the original stochastic runtime.

The training bridge uses a different $1.5$B policy, one continuation seed, six updates and sparse exposure. It establishes neither equivalence nor a general failure of hindsight, and supplies no evidence of fresh-task generalization. The practical contribution is a claim-matched evaluation pattern: report reference ordering when using rankings, isolate the decision credited to a score, and test learning utility with its realized exposure.\label{end:main}

\clearpage
\subsection*{Reproducibility statement}
The anonymous source package includes the existing numerical kit, original A/B scientific inputs and statistical scripts, frozen plans and archived results, and the matched information-ablation inputs and reproduction entrypoint. The artifacts include recorded provenance and checksums. They support numerical reconstruction and figure generation; a fresh environment replay or model-training run is a separate operation.

The completed C/D study is retained from the prior manuscript and its recorded validation (Appendix~\ref{app:freshcd}); that documentary source is included. The accompanying artifacts include documentary C/D evidence, but not the raw Attempt12 execution records; we report their original validation without claiming a new raw-record reconstruction.

The learned-policy bridge includes task outcomes and machine-readable audits of paired contrasts, realized exposure and recorded costs. Raw execution traces support the checks in Appendix~\ref{app:trainingbridge}. Some tensor-level gate checks remain receipt-backed because original state tensors and clip bits were not retained in the supplied gate bundle. The incremental trainer bundle also requires its original parent commit; a cold rebuild of that full trainer is not certified by this numerical package.

The retained historical replay archive has a write-time manifest: $29{,}402$ files were discovered, $29{,}401$ archived, and one unreadable file recorded. Trained adapters and the larger archival repository are staged separately for de-anonymization. The anonymous numerical kit accompanies the manuscript without exposing that repository's identifying link. Two historical common-support medians remain disclosed but unrecomputed because their generating procedure was not archived and their inputs predate an instrument correction. Runtime versions, reconstruction checks and historical corrections remain in the appendices. The supplied numerical inputs support the reconstruction scope stated above.

\section*{AI Use Statement}
AI systems assisted with experimental execution under fixed protocols, development
of computational plans within the authors' specified scope, code and prose
drafting, consistency checks, and generation of reproducibility records.
Human authors set the scientific scope and retained authority over interpretation
rules and final claims. Computational plans were fixed before their production
execution; the timing and non-primary status of later analyses are disclosed.
Training-bridge numerical results were independently validated against supplied raw records; the integration of replay results was checked against the baseline, included tables and recovered anonymous numerical artifacts;
historical values with unavailable inputs are not represented as regenerated.
In the independent-bank study, cohorts, endpoints, statistical tests and
interpretation rules remained fixed through scientific outcome review.
The policy and judge models studied here are distinct from AI assistance used in
preparing the paper. Errors involving AI-assisted pipelines and their corrections
are documented in Appendix~\ref{sec:integrity}.

\section*{Ethics Statement}
This work audits the validity of machine-generated training signals against executed
environment replay in a synthetic household-task environment (ALFWorld). It involves no human
subjects, no personal data, and no sensitive content; the audited models are publicly released
open-weight checkpoints used under their licenses; the anchor judge of the agreement control is a hosted API model (GPT-4.1), used under its provider's terms. We report control-sensitive evidence about widely used
credit signals so that training practice is not misdirected by overinterpreted per-step scores; we
see no ethical risks specific to this study beyond those general to research on agent training.

\bibliography{references}
\bibliographystyle{iclr2027_conference}

\newpage
\appendix
\renewcommand{\thesection}{A\arabic{section}}
\subsection*{Guide to supporting analyses}
\noindent The appendix connects the main displays to their detailed definitions, uncertainty and scope.
\begin{center}\small
\begin{tabular}{@{}ll@{}}
\toprule
Main claim / display & Detailed support \\
\midrule
Figure~\ref{fig:postsummary}A: Reference & \ref{app:postagreement}, \ref{app:postprecision}, \ref{app:p0rank} \\
Figure~\ref{fig:signal}: Signal & \ref{app:signaldetail}, \ref{app:r19audit} \\
Figures~\ref{fig:postsummary}B and~\ref{fig:recovery}: Recovery & \ref{app:postrecovery}, \ref{app:exactuniform}--\ref{app:p0recovery} \\
Table~\ref{tab:informationablation}: Information-source C/D & \ref{app:informationablation}--\ref{app:freshcd} \\
Table~\ref{tab:trainingbridge}: Learned-policy bridge & \ref{app:trainingbridge} \\
\bottomrule
\end{tabular}
\end{center}

\section{Learned-regime training bridge: evidence, uncertainty and costs}\label{app:trainingbridge}
\subsection{Protocol and reconstruction boundary}
This pilot is separate from the historical training diagnostics and the $7$B frozen-policy replay study. It resumes the public \texttt{chanyoungkim/grpo-qwen2.5-1.5b-alfworld-step140} checkpoint (revision \texttt{3bff836531bc3a1581751ad615f0ec0d0af3e2ca}) for updates 141--146, using continuation seed 91000. The publisher's original source commit and post-140 training stream are not known. This is an independent implementation on a learned policy, not an official or native HCAPO reproduction.

The frozen manifests specify 16 groups of eight episodes per update (128 episodes), six updates, and 128 validation-seen tasks with fixed environment seeds beginning at 1000. Evaluation samples at temperature $0.4$; training uses temperature $1$. Each arm has one recorded evaluation rollout per task/checkpoint. Hyperparameters include constant learning rate $10^{-6}$, one proximal policy optimization epoch, minibatch size 256, microbatch 32 split into four chunks of eight, maximum prompt/response lengths 2048/512, and Kullback--Leibler regularization coefficient $0.01$ against the base policy. F/N run on two A40 devices each. The O hardware receipt records two H100 devices, although the plan names A100; the observed hardware is used in cost reporting. Consequently O is not a hardware-matched estimate of adding the scoring machinery.

Independent reconstruction reads episode, timing, scorer and per-turn records rather than supplied headline summaries. It verifies task identities, ordering, environment seeds, binary outcomes, training manifests and finite metrics; recomputes the credit quantities from action log likelihoods; and checks their aggregates. All 160 implemented checks pass. The task-outcome table contains all $3\times3\times128=1{,}152$ records. These checks validate the supplied records, not an independent rerun of model training.

\subsection{Treatment and realized exposure}
F and N add a micro-credit term of weight one to the same macro group-relative policy optimization signal. F's scorer prompt includes terminal task/outcome/final-observation information and the next observation; N replaces both fields with the literal string \texttt{(not available)}. Both retain the same retrospective normalization machinery. History truncation is chosen using the N prompt in both arms. This treatment is not identical to the recovery hindsight ablation.

For mean command-token log likelihood $\ell_t$, form $s_t=\exp(\ell_t/5)$ and trajectory-normalized $\rho_t=\operatorname{clip}(s_t/\bar s,0.8,1.2)$. Then $Q_t=\rho_t\,0.95^{T-t}\,10\,\mathbb{1}\{\mathrm{won}\}$, with $Q_t^{s}=0.5Q_t+0.5Q_{t+1}$ except the final turn, which uses its own $Q_t$. At each group/turn, the active trajectories' smoothed values are centered and divided by their sample standard deviation plus $10^{-6}$; at least two values and nondegenerate standard deviation are required. The resulting micro term $M$ is clamped below at zero on successful trajectories; unsuccessful trajectories retain negative values. The audit reconstructs $\rho,Q,Q^s,M$ and eligibility from the raw log likelihoods and outcome records. An eligible turn is successful and belongs to a nondegenerate group/turn; it need not have nonzero $M$. Conversely failed turns can have nonzero negative $M$.

\begin{table}[ht]\centering\small\begin{tabular}{lrrrr}\toprule
Update & Variable groups & Turns & Eligible & $M\ne0$\\\midrule
141 & 0 & 3528 & 0 & 0\\
142 & 2 & 4512 & 336 & 732\\
143 & 2 & 4884 & 184 & 448\\
144 & 1 & 4021 & 39 & 312\\
145 & 1 & 4248 & 482 & 426\\
146 & 1 & 3404 & 80 & 320\\
Total & 7 & 24597 & 1121 & 2238\\
\bottomrule\end{tabular}

\caption{Realized F exposure by global update. Each update has 16 groups. Reward variability, eligibility and nonzero micro-credit measure different aspects of the realized signal; none is an experimentally isolated mechanism.}\label{tab:trainingexposure}\end{table}
N has $2/96$ reward-variable groups, 2,264 eligible turns and 1,540 nonzero micro terms among 25,520 turns. O has $4/96$ reward-variable groups and 23,851 turns. These arm-specific trajectories differ; sparse F exposure bounds interpretation but does not identify its cause or predict an effect at greater exposure.

\subsection{Paired outcomes and uncertainty}
Table~\ref{tab:trainingcontrasts} reports every within-arm checkpoint change and the raw and difference-in-differences (DID) F--N comparisons. DID subtracts each arm's own $t0$ before comparison. F and N have identical binary outcomes task by task at $t0$, so DID and raw contrasts coincide exactly at later checkpoints. Table~\ref{tab:trainingtransitions} preserves discordance even when aggregate totals are close.

Intervals are post-hoc, unadjusted 95\% paired-task percentile bootstrap intervals, using 100,000 common resamples of the 128 task indices and NumPy PCG64 seed 20260924. Each resample retains all arm/checkpoint outcomes for a task. These intervals describe task resampling conditional on the realized continuation and evaluations; they do not quantify training-seed variability or repeated rollout uncertainty. No equivalence test was performed. No task was removed on the basis of its outcome.
\begin{table}[ht]\centering\small\begin{tabular}{lrr}\toprule
Comparison & Change (pp) & 95\% interval\\\midrule
F\_t0\_to\_t3 & -3.91 & [-12.50, 3.91]\\
F\_t0\_to\_t6 & -5.47 & [-14.06, 3.12]\\
F\_t3\_to\_t6 & -1.56 & [-10.16, 7.03]\\
N\_t0\_to\_t3 & -5.47 & [-12.50, 1.56]\\
N\_t0\_to\_t6 & -3.91 & [-11.72, 3.91]\\
N\_t3\_to\_t6 & +1.56 & [-4.69, 7.81]\\
O\_t0\_to\_t3 & -8.59 & [-17.19, 0.00]\\
O\_t0\_to\_t6 & -4.69 & [-12.50, 3.12]\\
O\_t3\_to\_t6 & +3.91 & [-3.91, 11.72]\\
N\_to\_F\_t0 & +0.00 & [0.00, 0.00]\\
N\_to\_F\_t3 & +1.56 & [-6.25, 9.38]\\
N\_to\_F\_t6 & -1.56 & [-10.16, 7.03]\\
DID\_FN\_t3 & +1.56 & [-6.25, 9.38]\\
DID\_FN\_t6 & -1.56 & [-10.16, 7.03]\\
\bottomrule\end{tabular}

\caption{Paired changes in percentage points. All intervals use the same task-resampling design and are descriptive, not multiplicity-adjusted tests.}\label{tab:trainingcontrasts}\end{table}
\begin{table}[ht]\centering\small\begin{tabular}{lrrrr}\toprule
Comparison & 00 & 01 & 10 & 11\\\midrule
F\_t0\_to\_t3 & 39 & 12 & 17 & 60\\
F\_t0\_to\_t6 & 38 & 13 & 20 & 57\\
F\_t3\_to\_t6 & 41 & 15 & 17 & 55\\
N\_t0\_to\_t3 & 43 & 8 & 15 & 62\\
N\_t0\_to\_t6 & 40 & 11 & 16 & 61\\
N\_t3\_to\_t6 & 48 & 10 & 8 & 62\\
O\_t0\_to\_t3 & 39 & 10 & 21 & 58\\
O\_t0\_to\_t6 & 38 & 11 & 17 & 62\\
O\_t3\_to\_t6 & 45 & 15 & 10 & 58\\
N\_to\_F\_t0 & 51 & 0 & 0 & 77\\
N\_to\_F\_t3 & 44 & 14 & 12 & 58\\
N\_to\_F\_t6 & 42 & 14 & 16 & 56\\
\bottomrule\end{tabular}

\caption{Paired task transitions. For ordered comparison $X\to Y$, columns give counts $(X,Y)=(0,0),(0,1),(1,0),(1,1)$, where 1 denotes success. Each row totals 128.}\label{tab:trainingtransitions}\end{table}

\subsection{Development gate and unresolved rule overlap}
The frozen rule has three clauses. A triggers replication when the absolute $t6$ DID is at least 5 pp, raw F--N signs at $t3/t6$ agree and are nonzero, and the run is healthy. B stops when absolute $t6$ DID is below 3 pp with no stable raw separation of at least 3 pp. C extends the same pair to $t10$ when absolute $t6$ DID is between 3 and 5 pp \emph{or the signs disagree}. The observed $+1.5625/-1.5625$ pp contrasts fail A, satisfy B, and also satisfy C through sign disagreement. The frozen text supplies no precedence. There is no supplied $t10$ result or adjudication that resolves this overlap. The evidence therefore supports failure to meet the replication criterion, but not an unambiguous preregistered stop. Assigning precedence after seeing outcomes would change the rule.

\subsection{Execution gates: what can be independently checked}
The gate release records a checkpoint-resume match for 338/338 weights and 29/29 optimizer states, Adam step 2871 and scheduler step 140. Raw continuation traces reach Adam step 2885 and scheduler 141 after 14 optimizer minibatches; the arithmetic is consistent. Original state tensors are absent, so this audit cannot independently repeat the tensor comparison.

The split-versus-native check preserves recorded sample ordering, masks, advantages, step count and accumulation boundaries. Independent recomputation from objective and gradient-norm traces gives maximum relative objective differences 0.50845 for native repetition and 0.05007 for split execution; gradient-norm differences are 0.06086 and 0.02203. The five recorded threshold comparisons pass. Parameter-update relative norms/cosines and clip mismatch are receipt-backed because the underlying parameter tensors and clip bits are absent. The recorded pass is an implementation gate under its stated tolerance, not exact mathematical or scientific equivalence of training procedures.

\subsection{Cost accounting}
\begin{table}[ht]\centering\small\begin{tabular}{lrrr}\toprule
Quantity & F (A40) & N (A40) & O (H100)\\\midrule
Training episodes & 768 & 768 & 768\\
Evaluation episodes & 384 & 384 & 384\\
Environment steps & 35,667 & 37,001 & 34,892\\
Batched scorer calls & 6 & 6 & 0\\
Scorer tokens & 22,982,478 & 22,475,817 & 0\\
Rollout device-s & 7,399.70 & 7,868.89 & 2,899.45\\
Scorer device-s & 1,310.20 & 1,292.63 & 0.00\\
Optimizer device-s & 10,271.70 & 10,738.64 & 855.70\\
Evaluation device-s & 3,611.51 & 3,673.15 & 1,477.18\\
Accounted device-s & 22,593.11 & 23,573.31 & 5,232.33\\
Elapsed wall-s & 12,129.88 & 12,589.05 & 3,149.94\\
\bottomrule\end{tabular}

\caption{Recorded training-bridge costs. Phase times are occupied-device seconds (two devices times phase wall time), not utilization or billing measurements. O uses H100, F/N A40; cross-device totals do not estimate relative efficiency.}\label{tab:trainingcost}\end{table}
Each arm records 768 training and 384 evaluation episodes. Optimizer time includes update, old-policy and reference-log-probability phases. Accounted totals exclude setup, idle time, failed launches, pilots and checkpoint overhead; elapsed wall time is also reported separately. Exact generated-token counts cannot be recovered from the supplied records. The O cost summary's generated-token field is an approximation based on padded mean response length and is not treated as an exact count.

F and N each make six batched scorer calls, one per update, but equal call counts do not mean equal work. F uses 506,661 more scorer tokens and 17.57 more occupied-device seconds than N. This is an observed difference across each arm's own trajectories, not the isolated causal price of adding hindsight fields on identical inputs. The scorer totals include the batched turn sequences, not six individual model forward passes. The pilot does not establish a favorable cost--benefit tradeoff for hindsight.

\section{Detailed signal comparisons}
\label{app:signaldetail}
Direct step-credit evidence is conditional. The historical fixed-mask audit finds
modest positive judge alignment and no detectable implicit alignment, while
neither signal has a detectable archived paired increment. Deeper-bank estimates
use small common computable populations and have broad intervals. None
calibrates latent contribution or establishes signal superiority.

\subsection{Conditional alignment at greater replay depth}
This predeclared deeper-bank precision analysis fixes membership at the
intersection of computable old3 and B3/B8/B16 correlations, separately for each
signal: $23/43$ implicit and $13/43$ judge trajectories. Fixed membership prevents
population drift across depths, but not selection into computability: score and
replay vectors must be nonconstant. The signals' intersections differ, so their
medians are not a paired comparison of signal quality.

\begin{table}[H]
\centering\small
\begin{tabular}{lrrr}
\toprule
Signal & Common members & Median Spearman & Adjusted interval \\
\midrule
Implicit & 23/43 & 0.200 & $[-0.316, 0.600]$ \\
Judge & 13/43 & 0.775 & $[0.000, 0.775]$ \\
\bottomrule
\end{tabular}

\caption{\textbf{Credit alignment remains conditional on computability.}
Median within-trajectory Spearman at B16 on common computable members,
with the frozen two-signal adjusted bootstrap intervals.
\textbf{Rows use different signal-specific populations; they are not a paired
comparison of signal quality.}
These are different populations from the historical $37/32$-trajectory audit.
Complete depth and membership summaries are in Appendix~\ref{app:postprecision}.}
\label{tab:signal}
\end{table}

The implicit old3-to-B16 change is $\RFImplicitChange$, with descriptive paired
$95\%$ interval $[\RFImplicitChangeLow,\RFImplicitChangeHigh]$; the judge change is
$\RFJudgeChange$, $[\RFJudgeChangeLow,\RFJudgeChangeHigh]$.
The deepest-alignment intervals and the depth-change intervals belong to different
prespecified summaries: the former adjust across two signals, while the latter
are descriptive paired intervals. They do not establish a depth-driven change,
absent alignment, or a difference between the two signals.

\signpost{What the conditional populations mean.}
With four positions per trajectory, a few ties or a constant vector can make a
rank statistic coarse or undefined. The judge has only three score levels;
implicit credit is clipped after normalization. Undefined values are not repaired with zeros. The judge's high point estimate
therefore travels with its small population and broad interval. The pooled
reference diagnostic cannot supply within-trajectory fidelity reliability
(Section~\ref{sec:independent}).

\signpost{Why the historical support matters.}
The historical three-replica design estimated contrasts only when four distinct
admissible alternatives were obtained. At that budget, $95\%$ Newcombe intervals
excluded zero for just $24/1{,}768$ complete turns \citep{newcombe1998}.
This provides resolution context, not a before/after comparison with $3/172$:
the historical interval construction and population differ from the new five-arm
bounds (Appendix~\ref{sec:map}). Sampled zeros and undefined turns cannot certify unimportant actions.
Exclusions change the support of alignment; deeper replication of fixed
alternatives does not undo that selection.

\subsection{Two historical comparisons, two statistical objects}
\begin{figure}[H]
\centering
\includegraphics[width=\linewidth]{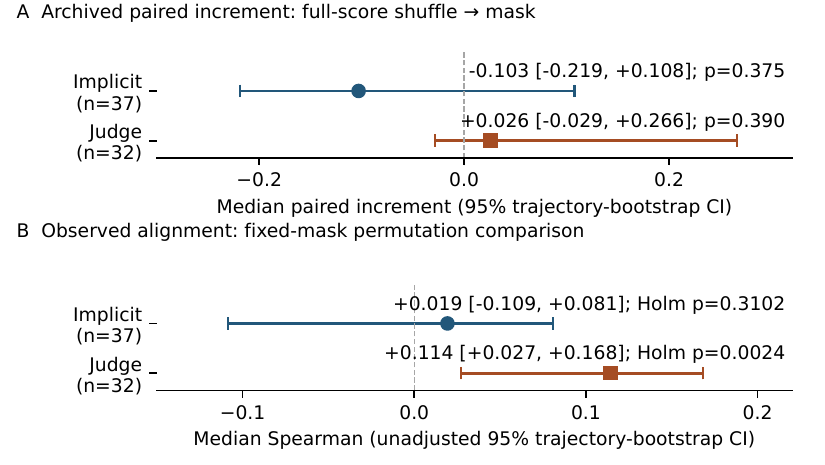}
\caption{\textbf{Historical alignment and paired increment answer different questions.}
(A) Increment over one full-score shuffle followed by masking;
$95\%$ trajectory-bootstrap intervals condition on that control, with two-sided
paired sign-flip $p$ values. (B) Observed alignment, with unadjusted $95\%$
trajectory-bootstrap intervals and positive fixed-mask permutation tests
(Holm over two signals). Intervals are not inversions of those tests.
Neither panel tests equivalence or differences between signals.
Full evidence: Appendices~\ref{sec:audit} and~\ref{app:paired}.}
\label{fig:verdict}
\end{figure}

The historical audit covers more positions, with only three replicas per arm.
Figure~\ref{fig:verdict} shows two questions on the same signal-specific
paired populations: observed alignment (panel B) and increment over an archived
control (panel A).

\signpost{Observed alignment under a fixed-mask null.}
A post-hoc analysis plan, frozen before this calculation, holds replay eligibility and finite
paired values fixed, then permutes retained scores without renormalizing them.
The statistic in Figure~\ref{fig:verdict}B is $T_{\rm obs}=\operatorname{median}_i r_i$, where $r_i$ is a
trajectory's observed score--replay Spearman correlation.
The adapted judge has median $0.114$, $95\%$ trajectory-bootstrap interval
$[0.027,0.168]$, $n=32$, and Holm-adjusted permutation $p=0.0024$.
Implicit credit has median $0.019$, $[-0.109,0.081]$, $n=37$, and Holm-adjusted
$p\simeq0.310$. The positive test uses $10{,}000$ permutations per signal.
Significance for one signal and nondetection for the other do not establish a
cross-signal difference; interval meanings are separated in Figure~\ref{fig:verdict}.

\signpost{Increment over one archived control.}
The archived statistic in Figure~\ref{fig:verdict}A is instead
$D_{\rm arch}=\operatorname{median}_i(r_i-s_{i,\rm arch})$:
each trajectory's observed correlation is paired with the correlation from its
one archived full-score-domain shuffle, applied before replay masking.
The implicit increment is $-0.103$, $95\%$ interval $[-0.219,+0.108]$,
with two-sided paired sign-flip $p=0.375$; the judge increment is $+0.026$,
$[-0.029,+0.266]$, $p=0.390$.
These intervals condition on that particular control realization. Neither
increment is detectably different from zero, and neither is an equivalence result
(Appendix~\ref{app:paired}). Thus a modest positive judge association and an
archived paired nondetection can coexist without contradiction.

\signpost{The control domain is part of the comparison.}
Implicit likelihood normalization precedes clipping and replay filtering;
both instruments score beyond retained replay pairs. Full-domain shuffling can
change the evaluated numeric multiset, but that alone need not change a rank null.
With unique scores and a fixed mask, uniform full-domain and retained-score
shuffles yield the same retained-rank distribution. Ties can change that
distribution and its undefined mass. Exact checks and counts remain in
Appendix~\ref{app:postcontrol}.

The readings differ in permutation domain, one-control versus distributional
conditioning, statistic and test. Their detection difference cannot be attributed
to mask order alone. The fixed-mask null also assumes that score assignments are exchangeable
across retained positions; holding the mask fixed does not validate that assumption.
Alignment can reflect action type or trajectory stage, rather than information
beyond those structures. Known-probability simulations check the named tests under
their generators, not real-data exchangeability or latent fidelity
(Appendix~\ref{sec:audit}).

\signpost{From uncertain alignment to a direct decision test.}
The reference contrast averages four alternatives; recovery chooses one
A-selected alternative at one score-selected position. Ranking the former does
not estimate the value of the latter. Because reference reproducibility and
direct alignment remain resolution- and support-limited, downstream decision
value requires independent evaluation. Section~\ref{sec:recovery} supplies it,
with supported branch gains and inconclusive position-selection value.

\begin{table}[ht]\centering\small
\begin{tabular}{lrrr}
\toprule
Estimate vs. B16 & Spearman & Pearson & MAE \\
\midrule
old3 & 0.063 & 0.439 & 0.099 \\
A1 & 0.236 & 0.236 & 0.136 \\
A3 & 0.283 & 0.436 & 0.092 \\
A8 & 0.300 & 0.620 & 0.071 \\
A16 & 0.330 & 0.736 & 0.056 \\
\bottomrule
\end{tabular}

\caption{Complete fixed-turn agreement summaries retained from the baseline. Nested A depths share B16.}\label{tab:agreementdetail}
\end{table}
\section{Independent-bank extension: frozen design, full results and validation}
\label{app:postexperiment}

\subsection{Design and chronology}
The historical analyses and canonical numerical kit remain unchanged. The primary B/C
design was frozen before formal outcomes; the secondary frontier and four-case
reporting rule were frozen after collection but before scientific outcome review.
Neither is described as an original confirmatory endpoint. Integrity validators
had read raw Bank-B files to verify completeness, schemas and accounting; only engineering counts and costs, not terminal-effect summaries, had been
reviewed for scientific decisions when the secondary plan was fixed.

The $50$ historical trajectories yielded $43$ formal members after four development
exclusions (seeds 011, 013, 041, 045) and three structural exclusions (010, 020, 025).
The structural rule requires at least four complete archived turns with five distinct
admissible actions. A known missing historical alternative replica at seed001,
turn033, alternative2 was excluded structurally before collection; it was not
silently supplied or selected after new outcomes.
Trajectory ordering used deterministic round-robin strata of pre-existing replay-map
values and implicit/judge full-score-mean terciles, with SHA256 ordering.
The measured cost gate admitted all $43$ eligible members; no outcome-based
cohort reduction occurred. At each trajectory, four position-quartile midpoint
indices among sorted eligible turns fixed the decision positions.
Factual actions and all four alternatives were copied from the archive, not
regenerated at any depth.

Replica namespaces were $200000+i$ for A and $300000+i$ for B,
$i=0,\ldots,15$, passed to the historical seed derivation with trajectory, turn,
arm kind and alternative index. The original replicas $0,\ldots,2$ and pilot
$100000,\ldots,100007$ are separate. Validation found no seed collisions.
The primary nested depths use $1,3,8,16$ replicas from each new bank.
The formal gate was created at 2026-09-15 00:54:25 UTC.
Bank-A decisions were frozen at 15:03:18 UTC; Bank B started at 15:03:19 UTC.
The $640$-continuation pilot was development only and does not enter any formal
estimate. Its predefined engineering and opportunity gates enabled the fixed design;
its scientific effects did not choose selectors, endpoints, directions or thresholds.

\subsection{Complete reference agreement}\label{app:postagreement}
\begin{table}[H]
\centering\small
\begin{tabular}{lrrrrr}
\toprule
Estimate vs. B16 & Spearman & Pearson & MAE & RMSE & Undefined /43 \\
\midrule
old3 & 0.063 & 0.439 & 0.099 & 0.182 & 19 \\
A1 & 0.236 & 0.236 & 0.136 & 0.275 & 21 \\
A3 & 0.283 & 0.436 & 0.092 & 0.176 & 18 \\
A8 & 0.300 & 0.620 & 0.071 & 0.124 & 14 \\
A16 & 0.330 & 0.736 & 0.056 & 0.100 & 14 \\
\bottomrule
\end{tabular}

\caption{\textbf{Complete reference summary, including undefined ordering counts.}
All 172 fixed turns contribute to pooled agreement. Within-trajectory correlations
use four positions; constant vectors remain undefined. No rows or undefined
members are removed to improve the result.}
\label{tab:fullbankagreement}
\end{table}

\subsection{Primary precision analysis}
\label{app:postprecision}
For bank $b$ and depth $d$, the same sampled-set contrast is
\[
\widehat A_{b,d}(t)=\widehat p_{b,d,f}(t)
  -\frac14\sum_{k=1}^{4}\widehat p_{b,d,k}(t).
\]
Across all $172$ positions we report pooled Spearman, Pearson, MAE and RMSE
against independent B16, plus within-trajectory ordering and its undefined count.
Table~\ref{tab:bankagreement} reports every primary A depth and the old3 comparator.
The $0.5$ ordering criterion, $0.2$ alignment magnitude and $0.1$ absolute
depth-change thresholds were fixed before pilot outcome review. They are
interpretation thresholds, not newly introduced significance or equivalence tests.

\begin{table}[htbp]
\centering\small
\begin{tabular}{lrrrrl}
\toprule
Signal ($n$/43) & old3 & B3 & B8 & B16 & Adjusted B16 interval \\
\midrule
Implicit (23/43) & -0.200 & 0.200 & 0.236 & 0.200 & $[-0.316, 0.600]$ \\
Judge (13/43) & 0.775 & 0.447 & 0.272 & 0.775 & $[0.000, 0.775]$ \\
\bottomrule
\end{tabular}

\caption{\textbf{Same-member credit--replay medians.}
Membership is the intersection of computable old3 and B3/B8/B16 correlations,
separately for each signal; it is not the full cohort or the historical $37/32$ population.
B16 intervals use the frozen two-signal bootstrap adjustment. All conditional
per-depth membership lists, including exclusions, are in the accompanying evidence.}
\label{tab:postfidelity}
\end{table}

Primary trajectory bootstrap uses $10{,}000$ draws, seed $2026091502$.
The deepest two-signal intervals take quantiles $0.0125,0.9875$; depth curves and
paired changes have descriptive $95\%$ intervals. An interval is withheld for fewer
than eight contributing trajectories. A constant score or reference produces an
undefined correlation with an explicit reason; there is no zero filling or redraw
until a statistic becomes defined. The common implicit set has $23$ members and
the judge set $13$, out of $43$. These are conditional complete-case summaries,
not an outcome-independent population representative of all trajectories.

\begin{table}[htbp]
\centering\small
\begin{tabular}{lrrrrr}
\toprule
Estimate & Resolved /172 & Positive & Negative & Median width & Old zeros resolved /105 \\
\midrule
old3 & 0 & 0 & 0 & 1.658 & 0 \\
A1 & 0 & 0 & 0 & 1.990 & 0 \\
A3 & 0 & 0 & 0 & 1.658 & 0 \\
A8 & 0 & 0 & 0 & 1.005 & 0 \\
A16 & 1 & 1 & 0 & 0.613 & 0 \\
B1 & 0 & 0 & 0 & 1.990 & 0 \\
B3 & 0 & 0 & 0 & 1.658 & 0 \\
B8 & 0 & 0 & 0 & 0.969 & 0 \\
B16 & 3 & 2 & 1 & 0.589 & 0 \\
\bottomrule
\end{tabular}

\caption{\textbf{All prespecified depth resolution summaries on 172 fixed positions.}
For each turn, five exact Clopper--Pearson arm intervals use $\alpha=0.05/5$.
The contrast lower bound subtracts the mean alternative upper bounds from the
factual lower bound; the upper bound reverses the endpoints. Exclusion of zero
defines resolution. This is simultaneous coverage across five arms at one turn,
not a family-wise guarantee across all 172 turns.}
\label{tab:postresolution}
\end{table}

The new bank narrows the median interval relative to the same-turn historical
three-replica measurement, but only three B16 contrasts exclude zero.
Among $105$ old zeros, $31$ acquire a nonzero point estimate and none resolves.
The archived full-design Newcombe intervals and this exact five-arm construction
have different definitions and denominators; their resolved fractions cannot
identify a pure replication-depth effect.

\subsection{Primary recovery analysis}
\label{app:postrecovery}
The population is all $30$ formal trajectories whose historical terminal outcome
was failure. Selector positions use only the frozen archived scores: minimum
policy token-mean log-probability, minimum implicit credit, minimum judge credit,
and a fixed identifier-hashed random position. Score ties choose the earliest
selected turn. Hindsight implicit/judge information is available because the
setting is post-trajectory recovery; this is not an online causal selector.
No new judge scoring was needed.

At the selected position, the chosen alternative maximizes its A16 success count;
ties choose the smallest frozen alternative index. The factual arm remains a
control, not a candidate in that alternative maximization.
The primary estimate is the trajectory mean of independent Bank-B selected-branch
success minus its same-position factual probability; three additional comparisons
use the random selector's chosen-branch success. They are not differences between
the selectors' factual-adjusted gains.
All seven comparisons appear in Table~\ref{tab:recovery}.
Bootstrap resamples the $30$ trajectories jointly across selectors, uses
$10{,}000$ draws and seed $2026091503$, and takes quantiles
$0.05/14$ and $1-0.05/14$ for approximate Bonferroni-adjusted intervals.
The practical success margin is $0.05$. No equivalence test was added.

Policy confidence and implicit credit share $23/30$ selected positions.
Their per-trajectory selected probabilities coincide on $28/30$, as do their
factual probabilities, but the full vectors are not identical.
The equal aggregate effects therefore cannot be counted as two independent
confirmations. The historical credit-fidelity audit and this recovery study target
different quantities on different conditional populations.

\begin{table}[H]
\centering\small
\begin{tabular}{lrr}
\toprule
Comparison & Difference (pp) & Adjusted interval (pp) \\
\midrule
Random $-$ factual & 3.125 & $[-1.667, 11.727]$ \\
Confidence $-$ factual & 8.542 & $[0.625, 20.625]$ \\
Implicit $-$ factual & 8.542 & $[0.625, 20.625]$ \\
Judge $-$ factual & 5.000 & $[-0.208, 13.542]$ \\
Confidence $-$ random & 3.333 & $[-2.083, 13.333]$ \\
Implicit $-$ random & 3.333 & $[-2.083, 13.333]$ \\
Judge $-$ random & 3.542 & $[0.000, 8.958]$ \\
\bottomrule
\end{tabular}

\caption{\textbf{Primary recovery estimates before main-text rounding.}
All seven comparisons retain the original three-decimal percentage-point display.
Machine-readable evidence retains the underlying full-precision numbers.
The interval family and 30-trajectory denominator are unchanged.}
\label{tab:fullrecovery}
\end{table}

\subsection{Four-case interpretation rule}
\label{app:postcases}
The reporting rule, fixed before scientific outcome review, first returns
Case 4 if a required primary quantity is undefined or recovery has fewer than
eight trajectories. Otherwise an A16/B16 Spearman below $0.5$ returns Case 3.
For stable replay, Case 1 requires a scored selector's adjusted lower bounds to
exceed $0.05$ against both factual and random controls. Case 2 requires
recoverability headroom (some selected-alternative versus factual lower bound
above $0.05$) and all scored-versus-random upper bounds below $0.05$.
Remaining configurations return Case 4. Case 2 would still not constitute a newly
validated equivalence test.

Here \textbf{Case 3} is selected: independent rank agreement is $\BankRank<0.5$.
Case 1 is not established, and Case 2 is not established.
Precision remains inconclusive for useful selector advantage over random and
broad credit-alignment claims, even though the finite-depth rank criterion fails.
No uncomputed confidence interval for ordering stability is asserted.
All formal outputs passed the specified runtime, restoration and output-integrity
checks; no formal engineering failure was detected.

\subsection{Secondary nested verification frontier}
\label{app:postfrontier}
The single secondary plan was fixed at 2026-09-16 10:49:56 UTC, after both banks
were collected but before primary scientific analysis and scientific outcome
review. It adds no rollouts. For each A arm, a SHA256 order of compact JSON
containing seed $2026091601$, bank, trajectory, turn, kind, alternative index,
replica and continuation seed defines nested prefixes of $1,3,8,16$ replicas.
There are $350$ distinct arm orders across $70$ distinct selector positions.
All $480$ selector--trajectory--depth choices were serialized from A before the
secondary evaluator read Bank-B scientific outcomes.
The alternative rule and tie break are unchanged.

Each depth evaluates the chosen alternative and same-position factual arm using
all $16$ B replicas. Verification costs include all five A arms, hence
$5,15,40,80$ continuations per position. The full fixed B allocation is
$80$ continuations per position, reported separately rather than charged as a
depth-dependent verification cost. Costs are observed generated tokens, model
calls and environment steps including restoration; historical score-computation
cost is not remeasured or included in this continuation-cost frontier.
The horizontal axis is not a prospective token cap: equal rollout allocations
can consume different numbers of tokens.

\begin{figure}[htbp]
\centering
\includegraphics[width=\linewidth]{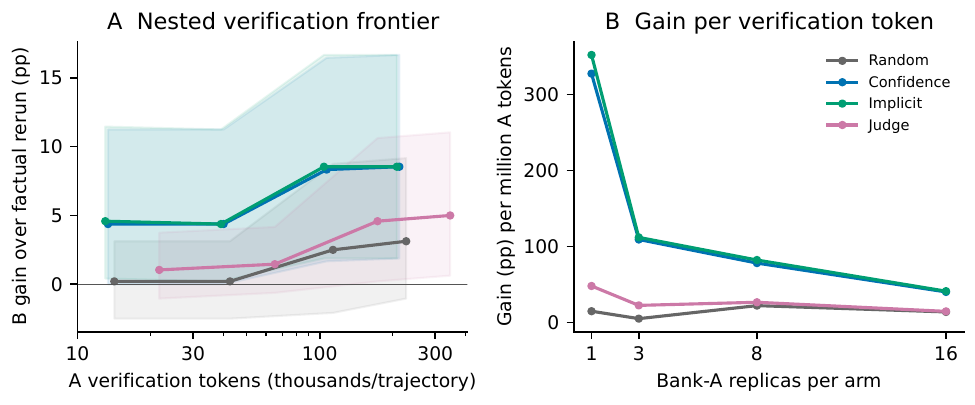}
\caption{\textbf{All fixed secondary operating points; no selected optimum.}
Left: mean independent B success gain over factual reruns versus measured A
verification tokens, with descriptive $95\%$ paired bootstrap shading.
Right: gain per million A verification tokens (point estimates; numerical
intervals are in the evidence). Each line follows depths $1,3,8,16$;
the same full B bank is reused at all depths. Shared selector positions and
nested subsets make these dependent comparisons, not independent experiments.}
\label{fig:postfrontier}
\end{figure}

\begin{table}[htbp]
\centering\small
\begin{tabular}{lrrrrr}
\toprule
Selector & Depth & A tokens & B tokens & B success (\%) & Gain (pp) \\
\midrule
Random & 1 & 14,192 & 229,406 & 3.958 & 0.208 \\
Confidence & 1 & 13,366 & 213,170 & 6.042 & 4.375 \\
Implicit & 1 & 13,025 & 208,195 & 6.250 & 4.583 \\
Judge & 1 & 21,758 & 347,503 & 6.458 & 1.042 \\
Random & 3 & 42,595 & 229,406 & 3.958 & 0.208 \\
Confidence & 3 & 40,119 & 213,170 & 6.042 & 4.375 \\
Implicit & 3 & 39,125 & 208,195 & 6.042 & 4.375 \\
Judge & 3 & 65,283 & 347,503 & 6.875 & 1.458 \\
Random & 8 & 113,467 & 229,406 & 6.250 & 2.500 \\
Confidence & 8 & 106,839 & 213,170 & 10.000 & 8.333 \\
Implicit & 8 & 104,131 & 208,195 & 10.208 & 8.542 \\
Judge & 8 & 173,318 & 347,503 & 10.000 & 4.583 \\
Random & 16 & 227,598 & 229,406 & 6.875 & 3.125 \\
Confidence & 16 & 213,299 & 213,170 & 10.208 & 8.542 \\
Implicit & 16 & 208,166 & 208,195 & 10.208 & 8.542 \\
Judge & 16 & 345,796 & 347,503 & 10.417 & 5.000 \\
\bottomrule
\end{tabular}

\caption{\textbf{Complete secondary curve and separate evaluation costs.}
Costs are mean generated tokens per historical failed trajectory.
Every row uses the same $30$ trajectories. B tokens charge the full five-arm
allocation; A tokens charge the nested verification subset. Gains are percentage
points over same-position factual reruns. Allocations overlap across selectors
and depths and must not be summed as physical experimental cost.}
\label{tab:postfrontier}
\end{table}

\begin{table}[htbp]
\centering\small
\begin{tabular}{rlrr}
\toprule
Depth & Comparison & Difference (pp) & Adjusted interval (pp) \\
\midrule
1 & Random $-$ factual & 0.208 & $[-3.958, 5.223]$ \\
1 & Confidence $-$ factual & 4.375 & $[-0.417, 16.265]$ \\
1 & Implicit $-$ factual & 4.583 & $[0.000, 16.473]$ \\
1 & Judge $-$ factual & 1.042 & $[-1.890, 5.833]$ \\
1 & Confidence $-$ random & 2.083 & $[-3.750, 14.390]$ \\
1 & Implicit $-$ random & 2.292 & $[-3.750, 14.598]$ \\
1 & Judge $-$ random & 2.500 & $[-0.625, 8.557]$ \\
3 & Random $-$ factual & 0.208 & $[-3.958, 5.223]$ \\
3 & Confidence $-$ factual & 4.375 & $[-0.417, 16.265]$ \\
3 & Implicit $-$ factual & 4.375 & $[0.000, 16.057]$ \\
3 & Judge $-$ factual & 1.458 & $[-1.667, 6.265]$ \\
3 & Confidence $-$ random & 2.083 & $[-3.750, 14.390]$ \\
3 & Implicit $-$ random & 2.083 & $[-3.750, 14.390]$ \\
3 & Judge $-$ random & 2.917 & $[-0.625, 9.375]$ \\
8 & Random $-$ factual & 2.500 & $[-3.958, 13.750]$ \\
8 & Confidence $-$ factual & 8.333 & $[0.000, 21.920]$ \\
8 & Implicit $-$ factual & 8.542 & $[0.208, 22.113]$ \\
8 & Judge $-$ factual & 4.583 & $[-1.042, 15.223]$ \\
8 & Confidence $-$ random & 3.750 & $[-0.640, 15.655]$ \\
8 & Implicit $-$ random & 3.958 & $[-0.625, 16.057]$ \\
8 & Judge $-$ random & 3.750 & $[-0.417, 10.833]$ \\
16 & Random $-$ factual & 3.125 & $[-1.875, 14.167]$ \\
16 & Confidence $-$ factual & 8.542 & $[0.208, 22.113]$ \\
16 & Implicit $-$ factual & 8.542 & $[0.208, 22.113]$ \\
16 & Judge $-$ factual & 5.000 & $[-0.625, 15.417]$ \\
16 & Confidence $-$ random & 3.333 & $[-2.500, 15.417]$ \\
16 & Implicit $-$ random & 3.333 & $[-2.708, 15.208]$ \\
16 & Judge $-$ random & 3.542 & $[-0.208, 10.208]$ \\
\bottomrule
\end{tabular}

\caption{\textbf{All 28 secondary contrasts.}
Approximate Bonferroni-adjusted paired trajectory-bootstrap intervals use
$10{,}000$ draws, seed $2026091602$, and quantiles $0.05/56,1-0.05/56$.
These non-primary intervals are distinct from the descriptive $95\%$ plot shading
and the seven-comparison primary family.}
\label{tab:postsecondary}
\end{table}

At depth $16$, all choices and seven point estimates exactly reproduce the
primary recovery endpoints. Similar values at depths $8$ and $16$ do not
establish saturation, an optimal budget, or a new stopping rule.
Compute-normalized ratios and their descriptive intervals, in tokens, calls and
environment steps, are retained with the complete curves.
No seed sweep or Pareto-envelope selection was performed.

\subsection{Control-domain attribution and numerical-kit reconciliation}
\label{app:postcontrol}
An exact-enumeration check distinguishes a numeric multiset from a rank null.
With unique scores over the full domain, a fixed mask, and a uniform permutation,
every retained ordering is equally likely. Uniformly shuffling only the retained
scores also produces each ordering equally often. Thus the Spearman null against
a fixed nonconstant reference is identical even though full-domain shuffling
can change the retained multiset. Tied scores can alter the retained tie pattern
and the probability of a constant vector; those changes can alter the rank null
and its undefined mass. This is an explanatory validation, not a novelty claim
or a real-data exchangeability proof.

The bounded archived-data analysis fixed $37$ trajectories per signal and
$2{,}000$ PCG64 controls per trajectory before formal outcome review.
It retains undefined statistics separately without replacement.
The matched decision comparison uses a pre-outcome guaranteed-computable
population: $37$ implicit and $24$ judge trajectories.
It holds population, statistic and tail fixed within each comparison,
using $128$ control pairs: two-sided bootstrap decisions disagree on
$1/128$ implicit and $22/128$ judge controls; positive one-sided decisions
disagree on $2/128$ and $43/128$.
Independent control realizations still contribute variation, so these frequencies
do not isolate mask order as the sole cause.

\begin{table}[htbp]
\centering\small
\begin{tabular}{lrrrr}
\toprule
Signal & Multiset changed & Ties changed & Full undefined & Retained undefined \\
\midrule
Implicit & 70,968/74,000 & 45,392/74,000 & 0/74,000 & 0/74,000 \\
Judge & 57,901/74,000 & 56,472/74,000 & 1,419/74,000 & 10,000/74,000 \\
\bottomrule
\end{tabular}

\caption{\textbf{Additional archive attribution, with its own fixed populations.}
Each denominator is $37\times2{,}000=74{,}000$ trajectory--control draws.
The judge population includes five constant retained original vectors;
their $10{,}000$ retained-shuffle undefined values are counted.
This table is not the canonical kit's $37/32$-trajectory diagnostic.}
\label{tab:postcontrol}
\end{table}

The canonical numerical kit separately reproduces $70{,}964/74{,}000=95.9\%$
implicit and $49{,}183/64{,}000=76.8\%$ judge multiset changes.
All four release-artifact hashes were verified before use. The kit's judge
population has $32$ members and its controls use Philox; the archive attribution
uses $37$ and PCG64. Their denominator and random-stream differences were
reconciled without changing or rerunning the completed archive analysis.
Conditional control-distribution distances, per-trajectory tie/undefined counts,
single-control variability and matched decision rows remain in the evidence.
Neither multiset-change percentages nor independently drawn control differences
by themselves prove that mask order caused the historical detection difference.

\subsection{Runtime, failures, compute and independent validation}
\label{app:postruntime}
The retained CPU environment was not the historical GPU runnable.
The isolated replay environment is labeled \texttt{RECONSTRUCTED\_RUNTIME}:
Python 3.11.13, Torch 2.11.0+cu130 (built against CUDA 13.0),
Transformers 5.14.1, vLLM 0.26.0, tokenizers 0.22.2,
ALFWorld 0.4.2 and TextWorld 1.7.0. All $221$ historical package pins matched;
two bootstrap extras and OS, glibc, path and bootstrap differences were recorded.
Historical wheel hashes were unavailable. Driver CUDA capability was not used
to infer the Torch build.
Qwen2.5-7B-Instruct revision
\texttt{a09a35458c702b33eeacc393d103063234e8bc28}
was verified across four weight shards and seven metadata files.
BF16, temperature $0.7$, the historical tokenizer/template, history length $2$,
maximum generation $512$ tokens and horizon $50$ were retained.
One A100-SXM4-80GB ran the frozen vLLM continuous-batching policy with at most
$16$ CPU workers; no model, environment family, judge or training method was added.

The smoke restored an archived prefix and ran factual and already-frozen
alternative continuations, checking exact restored state, admissibility, prompt
identity and terminal mapping. Two deterministic fixture runs matched exactly.
Full seeded stochastic traces were not bitwise identical under the retained
continuous-batching contract; this limitation was recorded before formal execution.
Structural and semantic validations passed before data were admitted.

All $13{,}760$ continuations in each bank validated, with no formal engineering
failure or missing/corrupted result. Generated inadmissible actions were retained
under the historical policy: $40{,}578$ in A and $41{,}702$ in B.
These are action counts, not failed restoration counts or dropped continuations.
An earlier socket-path smoke failure and a prelaunch executable-path error
produced no scientific data; their attempts and receipts remain preserved.
There was no outcome-selective retry.

Formal workers used $\WorkerGPUHours$ aggregate A100 hours:
$14.039$ for A and $14.118$ for B, producing $82{,}040{,}853$ tokens,
$638{,}891$ model calls and $1{,}262{,}571$ environment steps.
This is measured worker time, excluding pilot/setup and idle rental time;
it is not a billing total.
Shared allocations are counted once in physical compute.
GPU collection ended after both banks; no deeper follow-up was executed.

The primary analysis ran unchanged. The first independent validator encountered
an integer-type API requirement on mathematically integral historical JSON floats.
Its isolated replacement asserted finite integral counts within the valid range
before casting, without changing data, formulas, seeds, endpoints or thresholds.
Independent validation then passed. A second numerical audit verified all
$37$ primary bootstrap intervals using frequency-weighted resampling and explicit
order-statistic quantiles, plus all agreement rows, $350$ nested arm orders,
$480$ secondary choices, $16$ curve points, $28$ secondary contrasts and cost
deduplication. Engineering failures are recorded separately from scientific evidence.

The accompanying reproducibility archive preserves the original validated primary
and interpretation bundles byte-for-byte and adds manuscript sources, rendered
tables/figures, input hashes and a claim-to-evidence map. It does not replace
the original canonical numerical kit or claim that a new public release has occurred.
\clearpage
\subsection{Post-hoc exact uniform-position sensitivity}
\label{app:exactuniform}
\paragraph{Question and status.}
The primary recovery experiment uses one frozen identifier-hashed random position
(Appendix~\ref{app:postrecovery}). We additionally integrate the intended uniform
choice over the same four positions to assess dependence on that one realization.
This sensitivity was conceived after the primary results were observed; its
analysis plan was frozen before this calculation. It is post-hoc and secondary,
distinct from the verification frontier fixed before initial outcome review
(Appendix~\ref{app:postfrontier}). Neither primary endpoints nor collected data change.

\paragraph{Estimand.}
For each of the same $30$ historical failures $i$, let $j$ index its four frozen
positions and $k\in\{0,1,2,3\}$ its frozen alternatives. At every position, choose
$k^*_{ij}=\arg\max_k c^A_{ijk}$ from A16 success counts, breaking ties by the
smallest alternative index. The factual arm is excluded. With $c^B_{ij,k^*_{ij}}$
successes among the $16$ B continuations of that branch, define
\[
 y^B_{ij}=\frac{c^B_{ij,k^*_{ij}}}{16},\qquad
 u_i=\frac14\sum_{j=1}^{4}y^B_{ij},\qquad
 \Delta_s=\frac1{30}\sum_{i=1}^{30}\bigl(y^B_{i,j_s(i)}-u_i\bigr).
\]
The confidence, implicit and judge positions $j_s(i)$ are the archived primary
choices; no score or position is recomputed. This is \textbf{exact over position
randomization, finite-sample in Bank-B outcome measurement}: an exact conditional
average over four slots, not oracle success, an expectation over latent outcomes
or a new independent evaluation bank. The primary comparator is one deterministic
identifier-hashed realization, not an empirical average over random selectors.

\paragraph{Uncertainty and results.}
The statistical unit remains trajectory, $n=30$, with no exclusions. We reuse
the primary $10{,}000$ joint bootstrap index draws, seed $2026091503$, original
trajectory order and linear percentile interpolation. The three scored-selector
contrasts form the complete sensitivity family; approximate two-sided
family-wise $95\%$ Bonferroni intervals use quantiles $0.05/6$ and $1-0.05/6$.
The original primary family remains seven, so interval-width changes cannot
be attributed solely to integrating position randomization.

\begin{center}
\small
\begin{tabular}{lrr}
\toprule
Scored selector minus exact uniform & Difference (pp) & Adjusted interval (pp)\\
\midrule
Confidence & 2.344 & [-2.708, 8.698]\\
Implicit credit & 2.344 & [-2.796, 8.646]\\
Judge credit & 2.552 & [-0.365, 6.667]\\
\bottomrule
\end{tabular}
\end{center}
All three unrounded intervals contain zero. The exact-uniform mean empirical
B success is $7.865\%$; the frozen-random mean is $6.875\%$.
Their frozen-random-minus-uniform difference is $-0.990$ percentage points,
reported descriptively without a new test. Full-precision values and all
per-trajectory terms are retained in the accompanying analysis artifacts.
Confidence and implicit select the same position on $23/30$ trajectories;
their equal aggregate effects are not independent confirmations.

\paragraph{Interpretation and limits.}
Under this additional comparison, scored position-selection value remains
unresolved: all three adjusted intervals include zero, while positive gains
remain compatible with the measurements. This is not equivalence, absence of
value or validation of a general random policy. The positive primary branch-versus-factual
findings and inconclusive primary frozen-random comparisons remain unchanged.
The four-position support, selected historical cohort, one environment and frozen
policy remain limiting. Finite A selection and B16 measurement are noisy; this
bootstrap does not integrate all continuation noise. Reusing B does not supply
an independent replication, a training result or new-task generalization.
Freezing this calculation does not retroactively make the question prospective
or adjust away the post-hoc choice to ask it.

\paragraph{Validation.}
The A-only stage froze all $120$ branches before B evaluation without reading B
outcomes and reproduced all four selectors' primary choices. All seven primary
estimates and adjusted intervals reproduced exactly. Independent implementations
agreed on uniform averages, contrasts and intervals; population, position and
action identity, complete outcomes without filtering, row-order invariance and
two clean runs also passed. The analysis package retains per-trajectory terms,
input hashes, leakage checks and deterministic resampling artifacts. No rollout,
model call, environment replay, seed search or additional hypothesis test was performed.

\clearpage
\subsection{Post-hoc recovery decomposition and alternative-selection value}
\label{app:p0recovery}
\paragraph{Status and question.}
These existing-data analyses were conceived after the primary results; all three
plans (this subsection and Appendices~\ref{app:p0rank}--\ref{app:p0power}) were frozen
before their calculations. They are post-hoc and secondary, not amendments of the
primary endpoints. No rollout, model inference or environment interaction was
added. The same $30$ historical failures, four positions, frozen actions, A16/B16
counts and archived selector positions are retained without exclusions.

\paragraph{An accounting identity, not causal mediation.}
At position $j$ in trajectory $i$, select the alternative with greatest A count,
using the smallest frozen index for ties and excluding the factual arm. Let
$S_{ij}$ denote its B success rate, $F_{ij}$ the factual B rate, and $M_{ij}$ the
mean B rate over the four alternatives. Define $U^S_i=\frac14\sum_j S_{ij}$ and
$U^F_i=\frac14\sum_j F_{ij}$. For archived selector position $j_s(i)$, write
$S_{is}=S_{i,j_s(i)}$ and $F_{is}=F_{i,j_s(i)}$. Then, exactly,
\[
 S_{is}-F_{is}=
 \underbrace{S_{is}-U^S_i}_{\text{position component }P_{is}}+
 \underbrace{U^S_i-U^F_i}_{\text{generic recovery }H_i}+
 \underbrace{U^F_i-F_{is}}_{\text{factual-baseline shift }L_{is}}.
\]
Only $P$ measures the selected branch's value relative to uniform position choice
under the same branch rule. $H$ includes generic alternative-versus-factual
headroom as well as any value of A-based branch selection; it is not exclusively
a search effect. $L$ records which factual baseline the selector encounters.
Terms may have either sign; their magnitudes are not causal shares.

\begin{center}\small
\begin{tabular}{lrrrrrr}
\toprule
Selector & $S$ (\%) & $F$ (\%) & $S-F$ & $P$ & $H$ & $L$\\
\midrule
Frozen random & 6.875 & 3.750 & 3.125 & -0.990 & 3.750 & 0.365\\
Confidence & 10.208 & 1.667 & 8.542 & 2.344 & 3.750 & 2.448\\
Implicit & 10.208 & 1.667 & 8.542 & 2.344 & 3.750 & 2.448\\
Judge & 10.417 & 5.417 & 5.000 & 2.552 & 3.750 & -1.302\\
\bottomrule
\end{tabular}
\end{center}
Differences are percentage points. Uniform-position selected-branch success is
$7.865\%$ and factual success $4.115\%$. Confidence/implicit therefore have
$8.542=2.344+3.750+2.448$ points of gain. The position term reproduces
Appendix~\ref{app:exactuniform}; its adjusted interval still includes zero.
The primary seven-comparison estimates and intervals also reproduce exactly.

\paragraph{Uncertainty.}
We reuse the original $10{,}000$ paired trajectory-bootstrap draws, seed
$2026091503$, and linear percentile interpolation. Component/rate intervals are
\emph{descriptive, unadjusted} $95\%$ intervals, not a new family of significance
claims. The common $H$ interval is $[0.885,6.979]$ points. Selected-position factual
shifts have intervals $[-2.240,2.812]$ for random, $[0.781,4.375]$ for confidence
and implicit, and $[-4.167,1.302]$ for judge. Full component intervals and exact
per-trajectory rational terms are in the analysis package. The primary seven
and exact-uniform three comparison families are unchanged.

\paragraph{Which alternative, holding position fixed?}
The separate contrast $Q_{is}=S_{is}-M_{i,j_s(i)}$ compares A's chosen alternative
with a uniform alternative at that same position, whose mean includes the chosen
alternative. Confidence, implicit and judge
form a new, explicitly post-hoc three-contrast interval family: two-sided
Bonferroni-adjusted $95\%$ trajectory-bootstrap intervals use tails
$0.05/6$ and $1-0.05/6$. Random and all-position $Q$ are descriptive only.
The $P$ and $Q$ families are separate, not simultaneous coverage of six contrasts;
their different detection status is not a test that $Q>P$.
\begin{center}\small
\begin{tabular}{lrr}
\toprule
Selector & $Q$ (pp) & Adjusted interval (pp)\\
\midrule
Confidence & 5.938 & [0.469, 13.438]\\
Implicit & 5.833 & [0.365, 13.281]\\
Judge & 3.958 & [-0.208, 9.479]\\
\bottomrule
\end{tabular}
\end{center}
These intervals support the complete frozen alternative rule at confidence/implicit
positions in this conditional cohort; they do not establish credit-score quality,
position-selection superiority or judge inferiority. Judge $Q$ is unresolved.
All-position $Q$ is $3.372$ points, descriptive interval $[1.133,5.977]$;
random-position $Q$ is $2.188$ points. Neither is promoted to an additional test.

\clearpage
\paragraph{Exact alternative-label calibration and its assumption.}
At each selector's fixed position, hold the four observed B alternative counts
fixed and choose a label uniformly, independently across trajectories. Exact
integer convolution of these $30$ four-point distributions gives the conditional
reference distribution. The one-sided inclusive tail is the probability that a
uniform-alternative policy's count sum is at least the A-selected sum. Factual
arms are never permuted. Holm adjustment covers exactly the three scored selectors.
\begin{center}\small
\begin{tabular}{lrr}
\toprule
Selector & Uniform-label tail & Holm-three tail\\
\midrule
Confidence & 0.000707 & 0.002121\\
Implicit & 0.000979 & 0.002121\\
Judge & 0.005974 & 0.005974\\
\bottomrule
\end{tabular}
\end{center}
These are exact \emph{reference-policy exceedance probabilities} on the observed
B payoff table. They are calibrated $p$-values only under the stronger model
that A's selected-label mapping is uniformly exchangeable over the four B
alternatives, independently across trajectories. Collection randomized continuation
seeds, not semantic action labels; deterministic alternative order and tie-breaking
need not satisfy that model. Thus these tails are not an unconditional test of
zero average $Q$, a five-arm exchangeability claim, a causal best-of-four correction,
or an integration over B-outcome noise. Judge's small conditional tail and its
zero-crossing trajectory interval answer different questions; both are retained.
The adjusted $Q$ intervals, not the label calibration alone, support the bounded
branch-selection statement above.

\paragraph{Ties, overlap and limits.}
Confidence/implicit select the same position on $23/30$ trajectories, so their
similar effects are not independent confirmations. Minimum-score ties occur on
$0/30$ confidence, $1/30$ implicit and $27/30$ judge trajectories. A's maximum
alternative count ties on $85/120$ cells, so $Q$ does not isolate the value of
A-count information beyond deterministic action order. The position rule takes
the earliest turn among minimum-score positions, not necessarily the earliest
of all four; the judge result concerns this coarse-score implementation.
Archived positions and the original smallest-index alternative rule are
unchanged. This is post-failure, hindsight-
conditioned recovery with a frozen policy, not a training result or external
replication. Finite A selection and B measurement, four-position/action support
and the selected historical cohort bound every interpretation. The post-hoc
question does not become prospective because its calculation was frozen.

\paragraph{Validation.}
Every trajectory-level and cohort-level decomposition holds exactly in rational
arithmetic. An independent count-dictionary implementation, integer-polynomial
reference distribution and frequency-weight/bootstrap-order-statistic calculation
agree with the primary implementation. Reversing input rows and re-keying produces
identical clean-run artifacts. No primary choice, interval, denominator or data
record changes. Full distributions, component intervals, hashes and all validation
records accompany the analysis package.

\clearpage
\subsection{Post-hoc clustered and tie-aware ordering diagnostics}
\label{app:p0rank}
\paragraph{Targets and resampling.}
The same $43$ trajectories and $172$ fixed positions are retained. A/B contrasts
use integer numerators $(4c_{\mathrm{factual}}-\sum_k c_{\mathrm{alt},k})/64$;
ties are exact, without jitter or outcome-conditioned filtering. We resample
$43$ whole four-position clusters with replacement in each of $10{,}000$ draws
(seed $2026091702$), recomputing average-rank Spearman, tie-aware Kendall $\tau_b$
and Pearson on the identical draw. Intervals below are descriptive $95\%$ linear
percentile intervals, not a simultaneous family or new hypothesis tests. The
Pearson--Spearman difference is a paired descriptive summary, not a universal
comparison between correlation scales.
\begin{center}\small
\begin{tabular}{lrr}
\toprule
Summary & Estimate & Cluster interval\\
\midrule
Pooled Spearman & 0.330 & [0.134, 0.517]\\
Pooled Kendall $\tau_b$ & 0.269 & [0.096, 0.438]\\
Pooled Pearson & 0.736 & [0.554, 0.835]\\
Pearson minus Spearman & 0.406 & [0.202, 0.601]\\
Non-tied pair concordance & 0.687 & [0.598, 0.768]\\
Median defined within-trajectory Spearman & 0.600 & [0.316, 0.800]\\
\bottomrule
\end{tabular}
\end{center}
\paragraph{Six position pairs per trajectory, with ties visible.}
For each unordered pair compare the signs of its A and B contrast differences.
All $43\times6=258$ pairs remain in the table; conditional concordance uses only
the $150$ non-tied pairs and is always reported with that denominator.
\begin{center}\small
\begin{tabular}{lr}
\toprule
Pair category & Count\\
\midrule
Concordant, neither tied & 103\\
Discordant, neither tied & 47\\
A tie only & 16\\
B tie only & 15\\
Both tied & 77\\
\bottomrule
\end{tabular}
\end{center}
Thus $103/150$ non-tied pairs agree, while $108/258$ pairs tie in at least one
bank. Per-trajectory Spearman is defined for $29/43$ trajectories; its median and
interval condition on that computability. All $14$ undefined IDs and individual
values remain in the package. Every reported bootstrap summary is defined in
all $10{,}000$ draws; there is no silent resample deletion.

\paragraph{Qualified interpretation.}
The positive paired metric gap survives trajectory clustering, but pooled
Spearman's interval spans the existing study-specific $0.5$ rule. The frozen
post-hoc classification is therefore \emph{partial support}, not an uncertainty-
level confirmation of the original point-based threshold failure. The original
Case~3 rule remains a point-estimate adjudication. Conditional within-trajectory
agreement is more informative than blanket rank failure would suggest, while its
ties, undefined populations and broad interval remain consequential. These targets
are distinct from pooled ordering and from historical credit-fidelity estimator
reliability. Kendall does not replace Spearman or make a different scale ``correct.''
No nonzero-union mask, additional metric, alternate seed or new test was used.

\paragraph{Validation.}
An independent weighted-midrank and explicit pair-count implementation checks all
$60{,}000$ bootstrap statistics, every interval and the complete pair table, with
maximum absolute discrepancy below $10^{-12}$. Raw identities, undefined values,
row-order invariance and two clean runs pass; full draws and code are archived.

\clearpage
\subsection{Post-hoc recovery design sensitivity}
\label{app:p0power}
\paragraph{Two goals under an explicit residual model.}
Use the original $30$ paired score-minus-frozen-random B differences $d_{is}$.
Resample whole three-selector residual rows $r_{is}=d_{is}-\bar d_s$ and add a
hypothesized true effect $\delta$. This preserves the joint empirical residual
distribution, not all constraints of the physical replay design. It is not
prospective power, evidence that an
effect exists, or a generative model for fresh tasks. Goal C1 requires the
seven-comparison adjusted lower interval bound to exceed $0$; C2 requires it to
exceed $5$ percentage points. At $\delta=5$ points, C2 is a boundary/null event,
not a conventional $80\%$-power target. The original point-estimate-above-$5$-point
flag is a different, weaker condition.

\paragraph{Frozen grid and bootstrap procedure.}
The sample-size grid is $30,40,50,60,80,100,125,150,200$ and the effect grid is
$0,2.5,5,7.5,10,12.5,15$ points. We use $5{,}000$ outer experiments per size,
sharing paired rows and draws across effect scenarios. All shifted residual
supports stay in $[-1,1]$; no clipping or exclusions occur. Outer streams use
seed components $(2026091703,n)$; selector-specific inner streams use
$(2026091704,n,s)$ in confidence/implicit/judge order.

The primary $10{,}000$-resample linear-percentile procedure and seven-comparison
lower tail $q=0.05/14=1/280$ are preserved \emph{in distribution}. For each outer
sample, integer differences $D=16d$ yield the conditional bootstrap-sum
probability polynomial raised to the sample size. Its finite lattice CDF is
computed by zero-padded convolution. Since $(10{,}000-1)q=35+199/280$, the lower
endpoint interpolates the $36$th and $37$th inner order statistics with weights
$81/280$ and $199/280$. Their joint uniform order statistics are drawn through
Dirichlet$(36,1,9964)$, inverted through the sum CDF, then rescaled and shifted.
This avoids a large nested simulation without replacing the CI with a t-test,
a reduced inner sample or an infinite-bootstrap percentile limit. Exact integer
threshold comparisons handle equality; numerical ambiguity invokes integer
convolution rather than a stochastic reroll. Only marginal selector power is
reported, not joint-family power from independent inner couplings.

\paragraph{Current-$n$ sensitivity.}
For a true $5$-point advantage, the chance of a positive adjusted lower bound is:
\begin{center}\small
\begin{tabular}{lrrr}
\toprule
Selector & Power at $n=30$ (\%) & MC $95\%$ interval (\%) & First $n$ with $80\%$\\
\midrule
Confidence & 45.96 & [44.58, 47.34] & 80\\
Implicit & 45.38 & [44.00, 46.76] & 80\\
Judge & 99.18 & [98.89, 99.39] & 30\\
\bottomrule
\end{tabular}
\end{center}
The Monte Carlo intervals are Wilson intervals for simulation error, not
uncertainty about the empirical residual model or future populations. The same
powers apply to a true $10$-point effect and C2 because only effect minus claim
threshold changes in this additive model. At the C1 null (or C2 boundary),
success rates are $0.04\%$, $0.02\%$ and $0.00\%$ respectively.

\paragraph{Shared-endpoint scope check.}
A separate post-hoc diagnostic, planned before its calculation with the power
results already known, verifies the actual B outcome-record identities.
Confidence/implicit/judge choose the same endpoint as frozen random on
$\SharedConf/30$, $\SharedImpl/30$, and $\SharedJudge/30$ trajectories; those
differences are structurally zero. At $\delta=5$ points, the existing shift
$d_{is}+\delta-\bar d_s$ moves all of these zeros by
$\ShiftConf$, $\ShiftImpl$, and $\ShiftJudge$ points, respectively.
All shifted values stay in $[-1,1]$ but leave the raw $1/16$ lattice. A continuous
shift may leave that lattice; the consequential restriction is that it does not
preserve shared-endpoint equality. Thus this is an abstract sensitivity, not a
physical replay generator. The diagnostic does not establish why judge power
is $99.18\%$ or invalidate the declared model's arithmetic. Full partitions,
record lineage and independent exact checks accompany the analysis package.

\clearpage
\paragraph{Effect and sample-size requirements are conditional.}
At $n=30$, the smallest effect on the frozen grid reaching $80\%$ power is:
\begin{center}\small
\begin{tabular}{lrr}
\toprule
Selector & C1: lower bound $>0$ & C2: lower bound $>5$ pp\\
\midrule
Confidence & 7.5 pp & 12.5 pp\\
Implicit & 7.5 pp & 12.5 pp\\
Judge & 5.0 pp & 10.0 pp\\
\bottomrule
\end{tabular}
\end{center}
For C1 at a true $5$ points (equivalently C2 at a true $10$ points), the first
sample sizes on the evaluated grid reaching each target power are:
\begin{center}\small
\begin{tabular}{lrrr}
\toprule
Selector & $50\%$ power & $80\%$ power & $90\%$ power\\
\midrule
Confidence & 40 & 80 & 80\\
Implicit & 40 & 80 & 80\\
Judge & 30 & 30 & 30\\
\bottomrule
\end{tabular}
\end{center}
There is no interpolation or extrapolation. The full $378$-cell power grid,
all $50/80/90\%$ required-$n$ rows, support checks and Monte Carlo intervals
remain in the analysis package, including every null/boundary and high-power
scenario. For confidence/implicit the first $80\%$ crossing is $n=80$, whose
estimated powers are $92.12\%$ and $90.96\%$; no claim is made about the smallest
integer $n$ between grid points.

\paragraph{Implications and limits.}
The model supports a precision limitation for confidence/implicit at a true
$5$-point advantage, but not a blanket claim that all selectors are underpowered:
judge reaches about $99\%$ modeled power under its own residual distribution.
That is not evidence that judge is superior or that a $5$-point effect exists.
Confidence/implicit overlap is not independent replication. Residual-shape
estimation error, different future trajectories, alternative support and changed
bank-selection noise are outside this sensitivity model. Frozen random remains
the primary comparator; exact uniform does not replace it here.

Reaching $n=80$ would require $50$ additional same-design failed trajectories,
about $32.74$ productive A100-hours at the archived mean throughput before
overhead. This extrapolation assumes eligible prefixes and unchanged throughput;
neither is established for a new cohort. Untested sizes $61$--$79$ are not ruled
out. Judge already meets the modeled target at $n=30$. No additional data were
collected for this sensitivity.

\paragraph{Validation.}
Independent arbitrary-precision integer polynomials validate $54$ prespecified
outer-sample quantile cases. Million-resample original-data fixtures check the
conditional CDF, and synthetic literal-resampling fixtures check the joint
order-statistic law. Every power cell and required-$n$ row is independently
reconstructed from saved integer quantiles and exact thresholds. Reversed-row
clean reruns agree. Full checks, numerical diagnostics, seeds and all outputs
accompany the frozen analysis plans. No scientific method or result-selection
rule changed.

\clearpage
\subsection{Post-hoc matched information-source ablation}
\label{app:informationablation}
\paragraph{Question, status and inputs.}
Does the hindsight bundle add value to recovery-location selection, and do A
success counts add value beyond default alternative order? The plan was frozen
before these calculations, after historical outcomes were known. This is a new
post-hoc question on existing data, not a new independent evaluation bank or
external preregistration. All $30$ historical Qwen failures, their original four
positions and four ordered alternatives, and A16/B16 remain; no trajectory or
zero difference is excluded. The $30$ exact game identities are distinct, but
the historical cohort is selected and task families can share structure.

\paragraph{Matched information switches.}
For $h\in\{0,1\}$, let $\ell^h_{it}$ be mean command-token log likelihood and
$D_i$ the same original $50$-turn domain. Recompute
\[
 q^h_{it}=\exp(\ell^h_{it}/5),\qquad
 \rho^h_{it}=\operatorname{clip}_{[.8,1.2]}
 \left(\frac{q^h_{it}}{|D_i|^{-1}\sum_{u\in D_i}q^h_{iu}}\right).
\]
Each condition uses its own numerical denominator over the unchanged domain.
$h=1$ includes the appended SUCCESS/FAILURE and first $500$ characters of final
\texttt{obs\_after}; $h=0$ removes that whole line. The rest of the prompt, command
mask, history, temperature transform and clipping are unchanged. Both select
the earliest exact minimum among the original four positions, denoted $j_h(i)$.
Removing H changes conditioning and prompt length; without a matched placebo
this isolates the implemented bundle, not semantics separately from formatting.
Full-trajectory normalization still makes H-off retrospective.

At position $j$, $k_1(i,j)$ takes the smallest original index among maximum
A counts, excluding factual; $k_0(i,j)$ takes the smallest original alternative
index without reading A. The default retains policy-generated candidate order,
not an information-free baseline. With $Y^B$ denoting B16 success rate, define
\[
 V_{ha}=\frac1{30}\sum_iY^B_{i,j_h(i),k_a(i,j_h(i))},\qquad
 \Delta_H=V_{11}-V_{01},\qquad \Delta_A=V_{11}-V_{10}.
\]
These are two simple effects with the other input on; no interaction is tested.

\paragraph{All results and uncertainty.}
The four descriptive means are $V_{11}=V_{01}=49/480=10.208333\ldots\%$,
$V_{10}=13/480=2.708333\ldots\%$, and $V_{00}=12/480=2.5\%$.
The original full-H scores, choices and $V_{11}$ reproduce exactly.
The $10{,}000$ saved paired trajectory-bootstrap draws (seed $2026091503$) give
a separate two-contrast Bonferroni family with linear percentile endpoints at
$0.0125$ and $0.9875$:
\begin{center}\small
\begin{tabular}{lrrl}
\toprule
Simple effect & Difference (pp) & Adjusted interval (pp) & Reading \\
\midrule
Added H bundle, $\Delta_H$ & 0.000 & $[-0.625,0.625]$ & Unresolved \\
A counts versus default, $\Delta_A$ & 7.500 & $[0.833333\ldots,16.666666\ldots]$ & Positive increment \\
\bottomrule
\end{tabular}
\end{center}
Intervals are approximate and conditional on the selected cohort, realized A
selection and finite B measurements, not project-wide multiplicity control or
fresh-task coverage. Different classifications do not establish
$\Delta_A>\Delta_H$. No loss margin, equivalence or removal criterion was added.

\paragraph{Decision dependence, not identical scores.}
H removal changes $7/30$ positions; A counts change $9/30$ actions at full-H
positions. A-max actually matches default on $21/30$ at either H condition;
this differs from counting tied A maxima. Across all $1{,}500$ domain turns,
$256$ clipped scores are exactly equal; maximum absolute difference is
$0.3521428799$. Clipping expands the minimum tie set on $1/30$ trajectories in
each condition. Full score vectors and tie sets accompany the per-trajectory
four-cell records.

\clearpage
\paragraph{Shared endpoints and empirical ties.}
An endpoint binds trajectory/prefix, executable action, model/runtime/decoding
contract and B replica lineage. Identical endpoints reuse the same B outcomes;
different endpoints with equal observed counts are only empirical ties.
\begin{center}\small
\begin{tabular}{lrrrr}
\toprule
Contrast & Shared-endpoint zero & Distinct-endpoint zero & Positive & Negative \\
\midrule
$\Delta_H$ & 23 & 5 & 1 & 1 \\
$\Delta_A$ & 21 & 4 & 5 & 0 \\
\bottomrule
\end{tabular}
\end{center}
All $30$ rows remain in both bootstrap contrasts. The four cells use $47$ unique
endpoints and $752$ existing B continuation records. Shared-endpoint zeros
describe this exact support; an all-zero resampled subset would not establish
equality on unseen tasks. Equal cell means do not imply equal decisions.

\paragraph{Provenance and validation.}
All $60$ likelihood dumps cover the complete domains. Archived producer code,
checkpoint verification, token lengths and original H-score reproduction support
source matching. Original dumps did not capture prompt strings, token IDs or
per-request revisions. Reconstructed prompts/token identities use the pinned
tokenizer and archived producer; they are not claimed as captured execution
tensors or fresh model inference. Original H scores reproduce with maximum
absolute error zero and all original position/action choices match.

Separate scoring processes see only their own likelihood inputs; default action
choice does not read A counts; neither choice process reads B values. All choices
and input hashes are serialized before evaluation. B-value mutation leaves
choices unchanged, H-input mutation leaves H-off unchanged, and A-count mutation
leaves default choices unchanged. Live synthetic fixtures activate each switch
and check clipping ties, missing-domain rejection and endpoint reuse. Independent
count-dictionary/rational arithmetic and weighted-bootstrap order statistics
reproduce all means and interval endpoints; reversed-row clean runs agree.
The complete plans, full-precision rows, score domains, hashes and deterministic
draws are provided in the analysis package. These checks validate computation,
not a new restoration experiment or a training positive control.

\paragraph{Interpretation and relation to existing controls.}
The A-count increment supports using those recorded counts over the original
default order at full-H locations in this cohort. It is a different comparison
from $Q$, which evaluates the complete rule against uniform alternatives.
The H bundle changes some decisions but has unresolved average recovery value
here; nondetection is not equivalence, absence of credit information, or evidence
that H can safely be removed. No deployable-policy cost saving was measured.
Neither comparison establishes score-informed position superiority, general
random-policy performance, latent causal credit or training benefit. Old
confidence/implicit overlap and their non-independence remain unchanged.

HCAPO already varies hindsight weight in a training ablation
\citep[Appendix D]{hcapo2603}; its successful-final-state conditioning and native
optimization differ from this adapted failed-trajectory recovery test. CARL
already compares guided and random forking during training
\citep[Section 6.3]{carl2512}. Our two switches neither reproduce nor refute those
training algorithms or TARL's annotation-dependent judge. C3 already recognizes
finite Monte Carlo variance in its replay reference \citep{c3exact2603}.
The contribution here is the measured effect of removing particular inputs from
one frozen decision pipeline, not inventing ablation or declaring those methods
uncontrolled. No model call, rollout, new task or fresh policy execution was added.

\clearpage
\subsection{Fresh independent measurement on the same historical cohort}
\label{app:freshcd}
The completed continuation evaluates the same $30$ historical Qwen failures with position choice over $1{,}324$ common archived eligible positions ($36$--$50$ per trajectory). Scores come from the same archived likelihood dumps as Appendix~\ref{app:informationablation}. Fresh continuations were collected at the $73$ selected positions: $43$ full-support choices under the two hindsight conditions and $30$ bridge positions. The original full $50$-turn normalization domains, clipping, earliest-minimum position ties and alternative ordering are retained.

Fresh C16 outcomes choose alternatives by maximum success count, with smallest-index ties. Independent D16 outcomes evaluate the frozen endpoint union; shared executable endpoints reuse the same D records. The design preceded these formal outcomes, while historical outcomes were already known. The disjoint engineering pilot is excluded.

H-off removes the implemented outcome/final-observation bundle; full-trajectory normalization still makes the procedure retrospective. Default chooses the first original alternative without reading C counts. The retained symbol $\Delta_A$ denotes the alternative-count switch, now using C rather than historical A counts. For hindsight mode $h$ and count/default choice $a$, $V^{CD}_{ha}$ averages the D success rate over all $30$ trajectories. The two conditional effects are $\Delta_H^{CD}=V_{11}^{CD}-V_{01}^{CD}$ and $\Delta_A^{CD}=V_{11}^{CD}-V_{10}^{CD}$; neither an interaction nor a direct comparison of effects is tested.

\begin{table}[H]\centering
\begin{tabular}{lrr}
\toprule
D-evaluated cell & Successes / $480$ & Mean (\%)\\
\midrule
$V_{11}$ & $31$ & $6.458$\\
$V_{01}$ & $35$ & $7.292$\\
$V_{10}$ & $27$ & $5.625$\\
$V_{00}$ & $28$ & $5.833$\\
\midrule
Simple effect & Estimate (pp) & Adjusted interval (pp)\\
$\Delta_H$ & $-0.833$ & $[-5.000,3.125]$\\
$\Delta_A$ & $+0.833$ & $[-1.250,3.750]$\\
\bottomrule
\end{tabular}
\caption{Fresh full-support information-source ablation. Each cell has $480=30\times16$ D replicas. The separate two-comparison family uses the prospectively bound $10{,}000$ paired trajectory draws and linear percentile limits at $0.0125$ and $0.9875$. Intervals are approximate and conditional on the historical cohort and realized C selection.}
\end{table}

H removal changes $13/30$ positions; C counts change $8/30$ full-H actions. For $\Delta_H$, the endpoint partition is $17$ shared-endpoint zeros, $10$ distinct-endpoint empirical zeros, one positive and two negative pairs. For $\Delta_A$, it is $22$, $3$, $3$ and $2$, respectively. All $30$ pairs remain in each estimate and interval.

The full-H four-position bridge chooses among each trajectory's original four positions and evaluates the C-selected alternative on D. Its mean success is $46/480=9.583\%$. Full-support $V_{11}^{CD}$ minus this bridge is $-3.125$ pp, descriptive only: no interval, test or H-off counterpart was specified. The bridge shares H-on scoring and C/D banks with the full-support cells but not the historical A/B outcomes. It does not isolate a support-by-hindsight interaction and is not compared with historical $V_{11}=10.208\%$.

\paragraph{Execution and provenance.}
The completed-study record reports $4{,}672$ C and $1{,}344$ D continuations, counting each shared endpoint once ($84$ D endpoints over $73$ positions). The reconstructed runtime uses the frozen policy revision beginning \texttt{a09a3545}, bfloat16, temperature $0.7$, $512$ new tokens, history length $2$ and the $50$-step horizon. Replica namespaces are C $400100$--$400115$ and D $500100$--$500115$, with structured SHA256 seed derivation and no collision with the $55{,}527$ historical seeds or the pilot.

The uncertainty method was bound on 2026-09-18 at 10:18 UTC, before the first C record at 10:59 UTC. The D receipt is at 20:38 UTC. The worker session (10:56--20:38 UTC) used $9.70$ A100-hours wall time on one A100-SXM4-80GB, excluding model load and including the $175$-second pilot; C and D phase times were $7.56$ and $2.06$ hours. The original completed-study validation reports schema, prefix-identity, seed and hash checks for all C/D raw records, reconstruction of C-only choices before D outcomes existed, and agreement among the producer and two independent reconstructions on cells, effects, intervals, partitions and the bridge.

\paragraph{Interpretation.}
The A/B four-position count increment is positive; the C/D full-support count increment and both hindsight increments remain unresolved. Support, selection bank and evaluation bank all change between studies. They are not pooled or tested against each other, and no replication-failure or support-only effect is inferred. The fresh count interval's upper limit ($3.750$ pp) is below the archived count point estimate ($7.500$ pp), while their adjusted intervals overlap; this descriptive comparison is not an equivalence or margin test. New seeds provide new outcome measurement on the same historical games, not new-task validation or uncertainty averaged over repeated C selection. These results do not establish scored-position superiority, latent causal credit, native training utility or a criterion for removing hindsight.

\clearpage
\section{Historical reference, signal and training analyses, retained}
\label{app:historicalaudit}
\label{app:historicalcontext}
The historical definitions and analyses below are retained from the validated
baseline, with headings demoted for this placement. They include the original
control-order figure and complete synthetic detector table.
The main text now interprets these results alongside independent reference
reproducibility and recovery evaluation.

\subsection{A sampled interventional reference from executed replay}
\label{sec:method}

The historical instrument is executed replay: at each decision point of a collected trajectory we re-execute the environment under sampled alternative actions and measure how the outcome distribution shifts. We instantiate it in ALFWorld \citep{alfworld2021}, a replayable single-agent tool environment built on TextWorld \citep{textworld2018}, with Qwen2.5-7B-Instruct \citep{qwen25} as the policy: 50 trajectories on a task list fixed before collection, sampled at temperature 0.7 under the HCAPO method's published ALFWorld agent template \citep{hcapo2603}, extracted unchanged from its appendix. Environment determinism was verified before collection and re-verified on every host that touched the data, enforcing the re-feed versus checkpoint-restore distinction of \citet[\S2.3]{survey2604} (the verification design: Appendix~\ref{sec:integrity}).
A concurrent formalization validates the same instrument under planted effects \citep{car2606}; the contrast is in
Appendix~\ref{app:concurrent}.

\signpost{The two audited families.}%
We audit HCAPO-style implicit credit and an adapted TARL-style judge. The implicit scorer is
the policy checkpoint. For action-command tokens $y_{t,1:m_t}$, its executed score is
\begin{equation}
q_t=\exp\!\left(\frac{1}{5m_t}\sum_{i=1}^{m_t}
\log\pi_\theta(y_{t,i}\mid y_{t,<i},h_t,s_{\rm final})\right),\qquad
\rho_t=\operatorname{clip}_{[0.8,1.2]}\!\left(\frac{q_t}{|D|^{-1}\sum_{u\in D}q_u}\right).
\label{eq:score}
\end{equation}
Here $D$ contains all trajectory turns with nonempty action-token dumps, before replay
filtering; action tags are masked out. This follows HCAPO's score normalization
\citep[Eqs.~6--7]{hcapo2603}, with our own hindsight-injection line and no training-stage
mask or smoothing applied to the audited score. The judge uses TARL's prompt adapted to
ALFWorld ($\{1,0,-1\}$; Qwen2.5-72B-Instruct) \citep{tarl2509}. Prompts, conditioning and
regime deviations are in Appendix~\ref{app:instrument}.
These two parameterizations are what this paper audits: other implicit-PRM constructions --- a
policy-reference log-likelihood ratio trained from outcome labels alone, for instance
\citep{freeprm2412} --- are out of scope, and these findings do not extend to them.

Rank agreement concerns contribution ordering, not magnitude calibration, and is
\emph{not necessary} for a score to be useful in training; the policy-gradient invariance argument behind that
distinction is in Appendix~\ref{app:scope}.
The replay reference at a turn is a contrast between two outcome distributions, never a single continuation; it is the interventional reference defined in Appendix~\ref{app:scope}. In the historical collection, at every action turn $t$ we re-execute the factual action at least three times and sample $K{=}4$ distinct admissible alternatives from the same policy snapshot at the collection temperature (within 300 seeded draws), rolling each alternative to terminal at least three times; the replay advantage is
$A_{\mathrm{replay}}(t) = \operatorname{mean}(\text{outcome} \mid \text{factual replays at } t) - \operatorname{mean}(\text{outcome} \mid \text{alternative rollouts at } t)$.
The realized continuation is never used as the factual estimate. The alternative aggregate is
the \emph{uniform mean over the sampled alternative set} --- each of the $K$ alternatives
replicated equally --- not a policy-weighted expectation; the estimand this choice fixes is
stated in Appendix~\ref{app:scope}.
When four distinct admissible alternatives cannot be sampled within the budget, the turn has
insufficient distinct alternatives under the sampling budget and the contrast is left
\emph{undefined} there; such turns are excluded and counted, not imputed. Under Qwen2.5-7B this leaves 1{,}768 complete turns of 2{,}034 intervened; the exclusion is itself a finding (Section~\ref{sec:map}). The following determinism statement concerns only the E010 checkpoint, not the full executed record count.
Prefix-restore determinism held across the full sweep: zero divergences over the 20{,}538 replay rollouts archived at the E010 checkpoint. That is the checkpoint's divergence coverage; the executed run is $27{,}318$ records, $2{,}034{\times}3$ factual and $1{,}768{\times}4{\times}3$ alternative.

\begin{figure}[!htb]
\centering
\includegraphics[width=\linewidth]{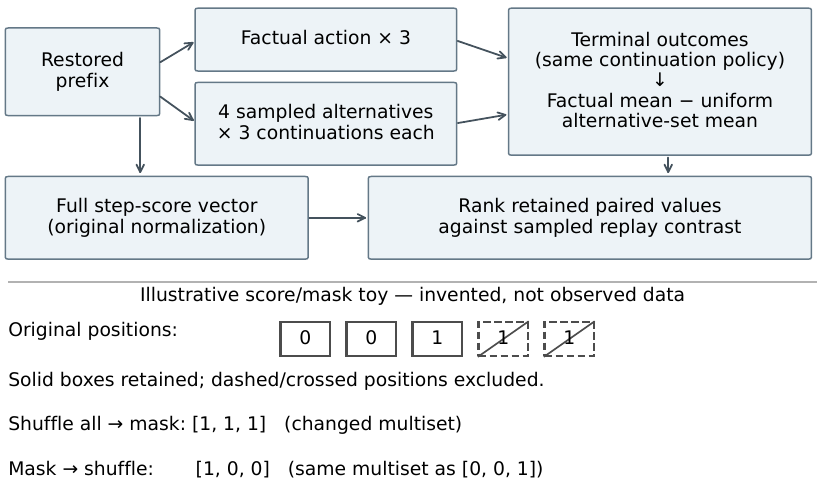}
\caption{\textbf{Executed replay and two distinct control orders.}
The factual action is re-executed, not represented by the original continuation.
Each of four policy-sampled alternatives receives three continuations under the same policy;
their uniform sampled-set mean is subtracted from the factual mean.
Scores keep their original normalization. Eligibility requires at least four complete replay
turns, then three finite pairs and nonconstant retained vectors.
The invented score/mask toy below is not an observation: full-vector shuffling can change the
retained multiset, whereas shuffling retained scores preserves it. Dashed, crossed boxes mark
excluded positions independently of color.}
\label{fig:replaydesign}
\end{figure}

The estimand this estimator targets, the three choices that fix it, and the scope statement
placing this audit on causal contribution are in Appendix~\ref{app:scope}. The same factual replicas give the estimator its noise floor and its resolution; the construct is the replica noise floor of \citet{survey2604}. Per turn, $\sigma_{\mathrm{floor}}(t)$ is the standard deviation of the outcome across factual replays, and the \emph{nonzero replay-contrast label} (called ``pivotal'' in the frozen protocol) holds exactly when $A_{\mathrm{replay}}(t) \neq 0$ under the literal stored value --- no significance filter, no floor threshold. Because outcomes are discrete and replicas few, exact zeros are common: 69.5\% of complete turns carry $A_{\mathrm{replay}} = 0$, and for the 1{,}184 turns where both arms are all-zero the data exclude only $|\Delta p| > 0.632$ at one-sided 95\% confidence. Every zero in this paper is therefore a resolution-bounded statement --- ``indistinguishable from the factual action at the achieved sampling resolution'' --- and never a claim of no causal effect.

\signpost{Where fidelity is scored.}%
\looseness=-1 Fidelity is scored where credit is consumed: within trajectories. For each family we compute the within-trajectory Spearman correlation between the credit values and $A_{\mathrm{replay}}$, aggregated as the median across trajectories with a 10{,}000-resample bootstrap interval; trajectories failing the registered completeness or constant-reference rules (fewer than four complete turns, or constant $A_{\mathrm{replay}}$) are excluded from this computation only, symmetrically across families, and counted. Random, uniform, and within-trajectory shuffled credit --- the shuffle preserves the full scored vector's marginal, before the replay mask --- run through the identical pipeline in the same batch.
The archived decision order was fixed in advance and its controls gate comes first: a family whose
interval overlaps its own shuffled control's earns outcome class H3 --- no detectable ranking
advantage over its own control --- regardless of its point estimate. Three structural bias tests accompany the verdicts under a Holm correction over all six family-level hypotheses; one of them, T3, carries a registered directional prediction --- positive for the implicit family --- that fluent, high-probability actions receive inflated credit regardless of causal effect.

The original protocol fixes thresholds, exclusions and the decision order; amendments and
post-hoc analyses are identified separately. Recoverable results are regenerated from archived
inputs; explicitly unrecoverable historical values are disclosed, not reconstructed.
Checkpoints are content-verified shard-by-shard before any stage loads them
(Appendix~\ref{sec:integrity}).
\begingroup
\interfootnotelinepenalty=10000
Because HCAPO's scorer \emph{is} the policy, the swap moves the measured object, the instrument
and the ruler together, so the cross-family run tests the system-level claim and not the scoring
instrument apart from the policy (Appendix~\ref{app:protocol}). The Llama arm collected 28 trajectories, 27 of which enter the replay-agreement analysis set after one exclusion; the fidelity estimate retains $21$ of the $28$ under the pre-registered exclusion rules (Section~\ref{sec:audit}). Its replay layer was re-executed in full under a corrected instrument after a chat-template defect was found and quantified (Appendix~\ref{sec:integrity}); its coverage is stated once here\footnote{\label{fn:coverage}Coverage 88.3\% (1{,}082/1{,}225 parseable turns). Missingness is concentrated exclusively in long ($\geq$44-turn) trajectories (13/28 partially covered; short/medium buckets 100\%). The 143 turns outside the analysis set --- 142 never generated at the cap trip $+$ 1 unparseable --- were not re-run, by pre-registered precedent ($\sim$1\% expected conversion). Length-sensitive quantities inherit this caveat (gate disclosure: Appendix~\ref{sec:integrity}).} and inherited by every Llama-side quantity.
\par\endgroup

\subsection{Coverage and resolution of the replay reference}
\label{sec:map}

\subsubsection{The measurability map}

The interventional design specifies a conditional contrast; finite replay estimates it only
where the required alternatives can be sampled. Under Qwen2.5-7B, $30.5$\% of complete
turns ($540$ of $1{,}768$) carry a nonzero replay contrast at the achieved sampling resolution.
That count is a resolution-bounded label, not a measurement of how sparse causal contribution
is: under the counting-structure null of Appendix~\ref{app:divisibility}, which holds each
turn's pooled outcomes fixed and randomizes which arm they land on, $509$ of the $540$ nonzero
labels are forced by arithmetic alone, the expected count is $536.80$, and the observed excess
is $0.862$ null standard deviations (two-sided $p{=}0.3884$). Factual outcomes show no
variation at $86.0$\% of complete turns ($\sigma_{\mathrm{floor}} = 0$). For the $1{,}184$
both-arms-all-zero turns, the data exclude only $|\Delta p| > 0.632$ at one-sided 95\%
confidence (Section~\ref{sec:method}); that bound does not cover every exact-zero contrast. These labels describe the sampled reference;
they do not determine which signals are useful for training.

\paragraph{How much of that reference a single turn resolves.}
Very little, and we measure it rather than assert it. At the historical full design --- $n{=}3$ replicas of
the factual action and of each of $K{=}4$ alternatives, $15$ rollouts per decision point --- a
95\% Newcombe score interval for the difference of two proportions \citep{newcombe1998} excludes
zero at $24$ of $1{,}768$ complete turns, $1.36$\% (Wilson 95\% $[0.91, 2.01]$\%;
\citealp{wilson1927}), and at none of the $1{,}204$ turns whose outcome pool is constant --- where
it is forced to zero by arithmetic and no interval could exclude it (Table~\ref{tab:divisibility},
Appendix~\ref{app:divisibility}).
Per-turn contrasts are therefore rarely individually resolved at this budget, and no claim in
this paper is made at the turn level; the fidelity estimates of Section~\ref{sec:audit} are
read uncorrected, the retracted correction standing as a sensitivity analysis
(Appendix~\ref{app:simval}). Why trajectory-level rank statistics remain readable when single
turns are not, the full $(n, K)$ grid, the subsampling-sensitivity figure and four cautions
are in Appendix~\ref{app:resolution}.

\subsubsection{Measurability is model-dependent --- in both directions}

The cross-family comparison below is descriptive and underpowered by its own amendment, which
gives the measurability map no confirmatory endpoint.
Both divergences, their Wilson 95\% intervals and their denominators are drawn in
Figure~\ref{fig:map} in Appendix~\ref{app:supfigs}, beside the rates in prose.
The measured inability to obtain four distinct admissible alternatives depends on the
collection protocol and action distribution at the fixed budget. Concentrated action mass
is a plausible explanation, not an isolated cause; this statistic does not itself measure
task capability. Coverage belongs to the (policy, environment, intervention support,
measurement budget) tuple, not an environment constant or a uniform ordering of audit
 difficulty (Appendix~\ref{app:scope}). Concurrent work reports a similar sparsity pattern
\citep{carl2512}; how ours differs in kind is in Appendix~\ref{app:concurrent}.
The exclusions shift the distribution of policy log-probability for Qwen; that shift
does not determine the direction of bias in rank fidelity. The measured difference and
the historical absence of a Llama counterpart are in
Appendix~\ref{app:limits} and Appendix~\ref{app:superseded}.
The analysis sets are therefore right-truncated on the fluency axis, with the direction of the truncation measured for Qwen, and Section~\ref{sec:audit} states what this range restriction does to the fidelity estimates.

\begin{figure}[h]
  \centering
  \includegraphics[width=\linewidth]{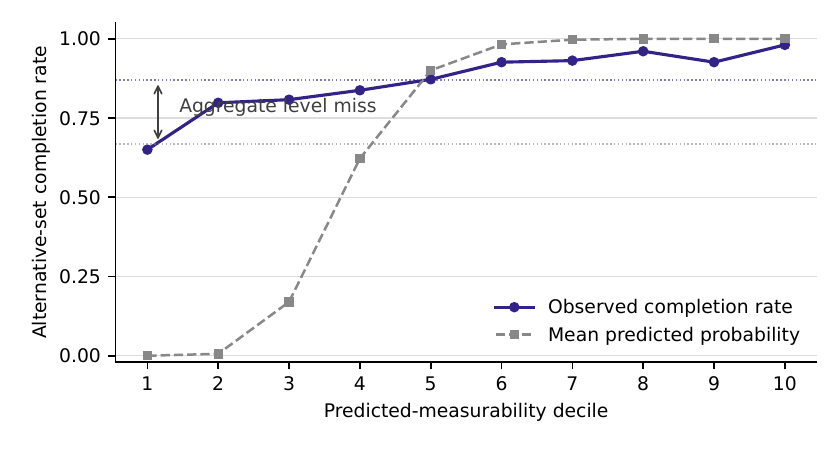}
  \caption{\textbf{Recorded completion differs across predicted deciles, with an aggregate
  level mismatch.} KS1/Qwen only, with no cross-family replicate or out-of-sample validation.
  Observed completion is $0.650$ and $0.981$ in the lowest and highest predicted deciles,
  respectively, with a decline from decile eight to nine. The aggregate level miss is about
  20 percentage points (Section~\ref{sec:map}). $n{=}2{,}034$
  intervened turns; data \texttt{runs/ks1xfam/measurability\_predictor.json} (\texttt{ks1}
  key).}
  \label{fig:s2}
\end{figure}

On the recorded Qwen turns, observed completion is higher in the highest than the lowest
predicted decile, but does not increase at every decile; the aggregate prediction also misses
the observed level. This is an in-sample relationship, not validated out-of-sample ranking
or calibration (Figure~\ref{fig:s2}, Appendix~\ref{app:supfigs}).
\subsection{Historical ranking conclusions depend on the comparison}
\label{sec:audit}
In the unchanged historical audit, the adapted judge has modest positive alignment with the sampled replay reference under a
fixed-mask permutation test; implicit credit is not detectably aligned.
Figure~\ref{fig:verdict} places this result beside the archived paired comparison.

\signpost{Observed alignment under a fixed-mask null.}%
A post-hoc plan fixed before execution permutes only retained scores, preserving their original
normalization. Eligibility is fixed first: at least four complete replay turns, then at least
three finite paired values and nonconstant vectors. The resulting $37/32$ implicit/judge
trajectories match the old paired sets; five of the $37$ replay-eligible judge trajectories
have constant retained original scores.
Writing $r_i$ for observed within-trajectory Spearman, the statistic is
$T_{\mathrm{obs}}=\operatorname{median}_i r_i$.
Each signal receives $10{,}000$ independent within-trajectory permutations; shuffled medians
form the reference distribution for a positive-alignment test, with add-one $p$ values and
Holm adjustment across two signals. The judge estimate is $0.114$ ($95\%$ trajectory-bootstrap
CI $[0.027,0.168]$, $n{=}32$, Holm $p{=}0.0024$); implicit credit is $0.019$
($[-0.109,0.081]$, $n{=}37$, Holm $p{=}0.310$). The CIs are unadjusted sampling intervals,
not inversions of the permutation test. Significance for one signal and not the other
is not evidence of a difference between them.

The null conditions on each trajectory's score and replay-reference values: the assignments
of scores to retained positions are exchangeable under the allowed permutations.
This is stronger than zero Spearman correlation and is not a causal no-effect null.
Average-rank ties preserve the zero mean shuffled correlation for fixed nonconstant vectors;
this algebraic property does not establish real-data exchangeability.
Overall alignment can reflect action type or trajectory stage; it is not residual alignment
beyond those structures.

\signpost{Why the archived comparison can remain a nondetection.}%
The archived statistic is $D_{\mathrm{arch}}=\operatorname{median}_i(r_i-s_{i,\mathrm{arch}})$,
where one full-score shuffle precedes the replay mask. Its intervals condition on that one
control and contain zero for both signals; two-sided paired sign-flip $p$ values are
$0.375/0.390$ (implicit/judge). The older registered interval-overlap rule also returns H3.
These results do not establish equivalence. Four components differ: the permutation domain,
conditioning on one random control versus a permutation distribution, the statistic, and the
test. Their different readings do not identify a gain caused by mask order alone.
A separate $2{,}000$-draw full-shuffle diagnostic changes evaluated score multisets in
$95.9\%/76.8\%$ of trajectory--permutation draws in this audit (implicit/judge);
judge controls can become undefined. The denominators and undefined counts are recorded in
Appendix~\ref{app:r19audit}. Exact numerical-kit reconciliation and the distinct bounded
archive attribution are reported in Appendix~\ref{app:postcontrol}.
Variation of $D_b=\operatorname{median}_i(r_i-s_{ib})$ over retained-score shuffles is control
sensitivity, not sampling uncertainty (Table~\ref{tab:r19control};
Appendices~\ref{app:paired} and~\ref{app:r19audit}).

\signpost{Known-answer checks at the executed sampling budget.}%
Known factual and alternative success probabilities generate independent $3+4\times3$
Bernoulli outcomes on the fixed Qwen design. Table~\ref{tab:r20validation} compares two
specified detectors, not the historical H3 overlap gate. Selected synthetic support is
typically $47$ trajectories, not the real $37/32$; thus these rates do not calibrate real-data
latent fidelity. Observed zero contrasts are not imposed as latent no-effect labels.
\begin{table}[!htbp]
\centering\small
\caption{\textbf{Detector behavior depends on achieved latent association.} Each cell reports a per-emulator rejection percentage with a Wilson $95\%$ Monte Carlo interval, from 1,000 datasets per scenario. Achieved latent correlation is the dataset median of the new-detector-selected trajectory median, not a generator parameter. Fixed-mask: positive test, 999 permutations; archived: two-sided paired sign-flip, 10,000 signs. Both apply Holm over the two scores per dataset. This comparison does not isolate mask-order power gain.}
\label{tab:r20validation}
\setlength{\tabcolsep}{3pt}
\begin{tabular}{llrcc}
\toprule
Scenario & Score & \shortstack{Achieved latent\\correlation} & Fixed-mask (\%) [MC95] & Archived (\%) [MC95] \\
\midrule
Null & Clipped & $0.002$ & $3.0$ $[2.1,4.3]$ & $2.2$ $[1.5,3.3]$ \\
Null & 3-level & $0.002$ & $2.7$ $[1.9,3.9]$ & $2.2$ $[1.5,3.3]$ \\
Weak & Clipped & $0.426$ & $92.4$ $[90.6,93.9]$ & $61.4$ $[58.3,64.4]$ \\
Weak & 3-level & $0.397$ & $89.8$ $[87.8,91.5]$ & $55.7$ $[52.6,58.8]$ \\
Moderate & Clipped & $0.798$ & $100.0$ $[99.6,100.0]$ & $99.4$ $[98.7,99.7]$ \\
Moderate & 3-level & $0.751$ & $100.0$ $[99.6,100.0]$ & $98.9$ $[98.0,99.4]$ \\
\bottomrule
\end{tabular}
\end{table}

The joint family-wise false-positive rate, counting either score's Holm rejection once per
association-null dataset, is $39/1000$ ($95\%$ Wilson interval $[0.029,0.053]$) for the
fixed-mask detector and $30/1000$ ($[0.021,0.043]$) for the old detector.
The stress-null counterpart is $42/1000$ versus $25/1000$; full intervals, overlaps and all
seven scenarios are in Appendix~\ref{app:r19audit}. These are saved-output reaggregations.
Synthetic independence does not validate real-data exchangeability, restoration or
missing-action selection.

The registered T3 transfer and descriptive cross-family fidelity retain their original scope
(Appendix~\ref{app:integrity}); no new Llama fidelity test is added.
Sign agreement and likelihood-conditioned partial correlations share this sampled reference,
not independent validations (Appendices~\ref{app:supfigs} and~\ref{app:r29}).
The withdrawn reliability correction remains sensitivity-only: its parameter-target validation
failed, and realized-target evaluation does not establish a calibrated latent bound
(Appendices~\ref{app:simval} and~\ref{app:step2}).

\clearpage

\subsection{Association, training and interpretation}
The following historical analyses are retained, with headings demoted for this placement.
They describe the historical evidence; the new independent-bank results qualify
its precision and recovery interpretation in Sections~\ref{sec:independent}--\ref{sec:recovery}.
\subsection{Association structure: gap, credit, replay and likelihood}
\label{sec:mechanism}

For implicit credit, the registered statistic
T3 correlates the credit--contribution gap $\Delta_t = z(\rho_t) - z(A_{\mathrm{replay}}(t))$
--- the within-trajectory z-scored difference between the credit a step receives and the
contribution replay measures there --- with the policy's own mean-token log-probability of the
action. The gap tracks likelihood: median per-trajectory rank correlation $+0.752$, CI
$[0.647, 0.793]$, $n{=}47$ trajectories; raw $p{=}0.0002$, Holm-adjusted $0.0008$ over the six
corrected together (Table~\ref{tab:holm}). This is implicit credit's pre-registered positive
prediction, and it survived multiplicity correction.

\signpost{Three associations on common support.}%
The executed z-score uses population SD and maps constant vectors to zero before the
likelihood mask. Thus T3 reduces to credit--likelihood correlation on constant-replay
trajectories, which cannot enter replay rank fidelity. A post-hoc common-support sensitivity
excludes degenerate vectors and uses identical turns and trajectories for all three
associations within each model, with its previously fixed masks and bootstrap.

\begin{figure}[!htbp]
\centering
\includegraphics[width=5.5in]{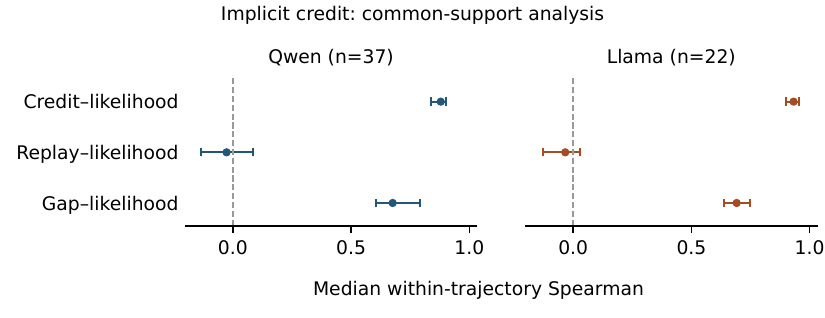}
\caption{\textbf{Implicit credit remains likelihood-associated on common support.}
Median within-trajectory Spearman and $95\%$ trajectory-bootstrap CIs;
the gap is $z(\rho)-z(A_{\mathrm{replay}})$, not raw credit.
All three associations share the same finite, nonconstant support within each model:
Qwen $n{=}37$, Llama $n{=}22$. Llama uses the corrected replay instrument, with its
coverage limitation (footnote~\ref{fn:coverage}). This separately planned post-hoc sensitivity
uses $10{,}000$ draws and the original fixed masks; it is not the registered full-set T3 test
or a direct credit--replay comparison. Source, membership and normalization:
Appendix~\ref{app:analysissets}.}
\label{fig:commonsupport}
\end{figure}

Positive credit--likelihood association persists after excluding degenerate vectors and
harmonizing support, whereas replay--likelihood intervals still span zero.
This does not prove that credit contains no contribution information.
Correlations of a difference do not decompose into differences of correlations; the
original unequal-support results remain in Appendix~\ref{app:analysissets}.

The exploratory regression and likelihood-conditioned partial correlations are distinct
descriptive diagnostics, not a causal decomposition; their definitions, coefficients and
cross-family qualifications remain together in Appendix~\ref{app:r29}.
The complete \texttt{seed017} case window (Table~\ref{tab:seed017}, Appendix~\ref{app:release})
illustrates high-likelihood repetitions with coarse, noisy replay contrasts of both signs.
It does not identify a mechanism. Training utility is a separate empirical question
(Section~\ref{sec:training}).

\subsection{The training layer: a diagnostic under incomplete causal isolation}
\label{sec:training}
\label{sec:dose}

Does alignment with replay matter for training utility? Seven arms---outcome-only, implicit
credit, judge credit, shuffled/inverted controls and reduced-resolution replay reference---used
common random numbers and were evaluated on $128$ held-out tasks.
The full three-seed analysis is a diagnostic, not a causal isolation:
superiority and $\pm3$\,pp equivalence comparisons test distinct nulls and remain
inconclusive (Table~\ref{tab:sevenarm}, Appendix~\ref{app:ks2table}).
At the observed seed-pair variances, one-sided minimum detectable effects span
$7.8$--$36.4$\,pp at $\alpha{=}0.05$, larger still at the Holm entry threshold;
task-level resolution is a different quantity (Appendix~\ref{app:mde}).
The action-format collapse in arm~5 seed~2 stays in the primary analysis because
the pre-registration supplies no exclusion rule; dropping it is an exploratory,
below-minimum-seed sensitivity, not a replacement result.

Credit rules changed both training content and effective optimization dose.
Checkpoint distributions differed, but without a dose-matched comparison those differences
cannot be attributed either to credit content or specifically to dose.
The observed distributional associations and Mantel diagnostics involve a small,
non-independent set of arm pairs; a significant marginal association followed by a
nonsignificant conditional association does not demonstrate mediation
(Appendix~\ref{app:e031b}). The matched comparison was cancelled; the recorded reasons
remain in Appendix~\ref{app:verbatim}.

Finally, the confidence-only routing rule is reported in Appendix~\ref{app:artifacts} as a
retrospective routing statistic: it recovers the resolution-bounded nonzero-contrast label at
chance level --- it is not a detector of causally pivotal steps --- and its routed fraction and
its per-trajectory cost reduction are different quantities, reported there together.

\subsection{Discussion: align controls with the audit question}
\label{sec:artifacts}
When scores cover more steps than a replay or verifier reference can evaluate, the scoring
and evaluation domains are distinct. Our toy shows why a full-domain shuffle need not
preserve the retained multiset; the ALFWorld diagnostic measures this occurrence and
occasional loss of a defined judge control. Its frequency and effect size need not transfer
to another pipeline. Different domains can serve legitimate questions, and the changes in
random-control handling, statistic and test prevent a mask-only causal interpretation.

The judge association concerns ranking a noisy sampled contrast, not calibrated contribution
magnitude, prompt-defined correctness or training utility. The implicit nondetection is not
equivalence, and the two signals have not been tested for superiority. Common-support
likelihood associations describe implicit credit, whose formula already contains likelihood;
they do not establish absence of contribution information or a causal pathway. Known-answer
simulations characterize the named detectors and achieved latent targets under their generator,
not arbitrary weak effects in real agents. Holding the mask fixed addresses neither selection
nor the exchangeability assumption. Dose-matched training and more precise replay remain
different empirical questions.

Controls in a ranking audit must match its declared scoring and evaluation domains.
For assignment information on a retained population, freeze and disclose its IDs, turn mask
and original score normalization, then state the permitted permutations and undefined cases.
Report trajectory-bootstrap uncertainty separately from variation across random controls.
The supplied numeric kit makes these checks concrete: its data contract retains full score
vectors and original turn indices; its reconstruction entrypoint verifies masks and IDs,
computes the observed and archived comparisons, and separates bootstrap intervals from the
retained-control range. Saved synthetic outcomes support detector-rate checks, and the plot
entrypoint rebuilds the displays. These are numerical reconstructions for the supplied inputs,
not an arbitrary-dataset framework or a rerun of the environment. The kit README and
Appendix~\ref{app:artifacts} describe the supported path.

The two historical comparisons are displayed together in
Figure~\ref{fig:signal} in the main text, with the historical presentation retained in Figure~\ref{fig:verdict}.
\clearpage
\section{Scope: what the audit measures, what fixes it, and what it opens}
\label{app:scope}

\paragraph{Coverage depends on the collection protocol and action distribution.}
Failure to sample four distinct admissible alternatives within budget is an observed coverage
property, not a task-capability measurement. Concentrated action mass offers a possible
explanation, but the policy comparison does not isolate that cause. Equation~\ref{eq:estimand}
conditions contribution on the policy, environment and intervention support; the finite
estimate additionally depends on the measurement budget. Changing policy can change the
sampled support, without establishing why every excluded turn lacks alternatives.

It is a property of the (policy, environment, intervention support, measurement budget) tuple --- not an environment constant, and not a statement that either family is uniformly harder to audit; the budget term is the one Section~\ref{sec:map}'s opening subsection measures.

\paragraph{What defined replay does not establish.}
Fidelity is evaluated only where the sampled replay contrast is defined
(Section~\ref{sec:map}). Excluding undefined turns does not eliminate sampling noise or
selection, so a null result can still reflect limited measurement resolution or power.
The implicit scores show substantial likelihood association; this observation neither
establishes the absence of causal information nor identifies a causal mechanism
(Section~\ref{sec:mechanism}).

\paragraph{Correctness against contribution, the two worked cases.}
A correct step can contribute nothing (the
trajectory was already determined) and an incorrect one can be pivotal (it opened the state from
which recovery happened); correctness, contribution ranking, calibrated magnitude and
training utility are distinct properties.

\paragraph{The estimand, and what estimates it.}
Write $h_t$ for the history before action $t$, $R$ for terminal outcome,
$a_t^f$ for the factual action and $\pi$ for the continuation policy. For one
sampled alternative $a'$, define the policy-conditioned contrast
\begin{equation}
\tau_t(a') =
\mathbb E[R\mid h_t,\operatorname{do}(a_t=a_t^f),\pi]
-\mathbb E[R\mid h_t,\operatorname{do}(a_t=a'),\pi].
\label{eq:estimand}
\end{equation}
For the realized alternative set $S_t$ of size $K$, the audit targets
$\mathcal A_t(S_t)=K^{-1}\sum_{a'\in S_t}\tau_t(a')$.
The support is the sampled admissible actions emitted under the collection procedure;
this is a design choice, not a claim that other interventions are scientifically invalid.
Both arms continue under $\pi$, and only terminal outcome is measured.
With equal replication of each alternative, the pooled alternative outcome mean equals
the mean of the per-alternative outcome means. Thus $A_{\mathrm{replay}}(t)$ estimates
$\mathcal A_t(S_t)$, not a single $\tau_t(a')$ and not a policy-weighted expectation
over all alternatives. Unequal replication would instead weight alternatives by their
replica counts unless the estimator were changed.
The sampled set is held fixed for this conditional estimand; drawing a new set adds a
different source of variation (Appendix~\ref{app:resolution}).
We reserve $\Delta_t$ for the normalized credit--replay gap in Section~\ref{sec:mechanism}.
The observed contrast is a Monte Carlo \emph{interventional reference},
not direct access to the latent effect.

\paragraph{What rank fidelity is, and is not, necessary for.}
A history-dependent additive term $b(h)$ leaves the expected policy gradient unchanged
\citep{sutton1999policy} yet can reorder scores across steps. Within-trajectory rank fidelity to
contribution supports an ordering interpretation, not magnitude calibration, and is
\emph{not necessary} for that score to be useful in training: a signal can fail this audit and
still train well. What we test is the interpretation, not the training use.

\paragraph{Three quantities, and which one this is.}
Step-level correctness, causal contribution, and training usefulness are three different things,
and the historical fidelity audit speaks to contribution. The prospective extension
separately measures frozen-policy recovery, not training. Correctness asks whether the step was right; contribution
asks what the step changed about the outcome, which is $\mathcal A_t(S_t)$; usefulness asks whether credit
built on either trains a better policy. Existing step-level benchmarks grade the first
(Section~\ref{sec:intro}) and we measure the second. We do not establish that the second implies
the third: the seven-arm training experiment of Section~\ref{sec:training} was built to test that
implication and returned an inconclusive null without a validated positive control, and licenses
no inference in either direction.

\paragraph{Two noisy instruments, not truth against proxy.}
What Section~\ref{sec:intro} sets up is not truth against proxy but two noisy instruments read
against each other: the judge's error is scoring error, replay's is sampling error, and only the
second can shrink with additional replicas. The attempted reliability correction remains
a sensitivity analysis, with no established calibrated bound (Appendix~\ref{app:simval}).

\paragraph{One direction the audit opens rather than closes.}
Executed replay is expensive as an audit
instrument, but at small $n$ cheap enough to be a candidate selective verifier inside a training
loop --- spent on the turns a cheap signal is least sure about, and used to gate rather than
score. Our resolution measurement bears on how such a budget would be spent: on the turns whose
contrast the environment actually decides, additional replicas of a fixed alternative set buy
stability and a fourth alternative does not (on the $55$ such turns, scored against a full design
that is the reference by construction rather than a converged measurement;
Appendix~\ref{app:resolution}). We have not built this training-loop verifier. The independent-bank recovery study instead
uses independent banks on fixed historical prefixes: its rank agreement remains below
the frozen criterion (Section~\ref{sec:independent}), and selector advantage over
random remains inconclusive (Section~\ref{sec:recovery}). The historical shared-reference subsampling result
does not establish independent-bank stability.

\section{The protocol, in full}
\label{app:protocol}

\paragraph{What the family swap does and does not test.}
For the implicit family the swap changes the measured object, the scoring instrument, and the ruler at once --- HCAPO's scorer \emph{is} the policy, and the replay reference is defined over the policy's own action distribution --- so the cross-family run tests the system-level claim, of observed agreement between a policy's own implicit credit and its sampled replay reference relative to a full-score-domain shuffled control, under a second system; it cannot and does not evaluate the scoring instrument apart from the policy.
\paragraph{The dose-matching table.} One reporting table per credit comparison: surviving
examples, optimizer steps, tokens updated, realised parameter displacement (free for LoRA
adapters via the trace identity, no materialisation), and the measured strength of every
control --- with the behavioural measure in the last column. The reading rule is one sentence:
if the behavioural column is associated with dose, the comparison does not isolate credit
content without dose matching. In our own arms, optimizer steps ranged from $112$ to $8$ per round under identical
budgets.
\paragraph{Measured strength, worked.} Within-trajectory shuffling changed $90.4$\% of
positions for our continuous credit but only $26.0$\% for the tie-heavy discrete judge scores,
so comparing the two families against ``their shuffles'' confounded information destruction with
perturbation magnitude. Every control should ship with its fraction-changed, displacement, and
rank-correlation against the original; our inversion control's assumed $-1$ correlation held
exactly, but held \emph{because we finally checked}.

\section{Instrument details and notes}
One arm-seed carries a format-collapse instrument note alongside its primary result;
the full three-seed analysis retains it. The note and its integrity lesson are below.

\label{app:instrument}
\paragraph{The implicit instrument, in detail.} The ratio $\rho_t$ is computed by
Eqs.~6--7 of the HCAPO paper exactly, with the published constants ($T_{\mathrm{temp}}{=}5.0$,
clip $[0.8, 1.2]$). Hindsight conditioning is a single outcome-injection line appended to the
prompt --- our addition, not HCAPO's; the \emph{policy} scoring mode uses the identical prompt
without it, which keeps Section~\ref{sec:mechanism}'s fluency contrast clean. Deviation note: KS1
rollouts run at temperature 0.7, frozen in the pre-registration; HCAPO's published rollout
temperature is 1.0, and the deviation is a KS1 design choice, logged.

The judge-family adaptation preserves the TARL judge's prompt structure \citep{tarl2509},
response contract, and
$\{1, 0, -1\}$ scale. TARL's published judge receives ground-truth tool-call annotations;
ALFWorld has none, so the judge runs without them --- a disclosed deviation from TARL's regime,
and the condition under which such judges are actually used during training.
The adapted judge's scores show modest positive ranking association with the sampled replay
reference under the fixed-mask test (Section~\ref{sec:audit}). That association does not establish
calibrated contribution measurement: correctness, contribution ranking and contribution
magnitude are distinct targets. Nor does the archived paired nondetection show that the judge
failed its prompt-defined correctness task. Correctness was not our measurand, no correctness
ground truth was collected, and this audit cannot adjudicate that task.

The inverted-credit arm's third seed remains in the primary result and carries an instrument
note. Its near-zero evaluation coincided with degraded output format: $87.7$\% of that seed's actions embedded the
action tag inside the action text, making them inadmissible to the environment, where every
other arm-seed sits at or near zero. The format gate passed it, because the gate tests for
loops and command-likeness, not tag placement --- an integrity lesson
(Appendix~\ref{sec:integrity}) arriving through the training layer.

\section{The scope conditions, in full}
\label{app:limits}

\paragraph{Additional scope conditions.}
One disclosure concerns the estimand's externality: calling the replay contrast an external ground truth implied more independence from return-derived credit than the construction delivers --- the full statement is the paragraph below --- though the audit stands --- none of the three audited families is return-derived.

KS2 lacks a validated positive control, so its null is reported as inconclusive, and three convergent checks --- the strongest an expert-cloning arm at $-10.9$\,pp on held-out evaluation ($n{=}128$) --- bound how badly the harness could be lying without substituting for one; they are itemized below.

We do not run the target-side placebo specified below, which would compare the archived
family-versus-control gap with a signal-free target under the same count structure.
The reported comparisons remain conditional on the sampled reference. They do not calibrate
arbitrary latent fidelity or establish power for an unspecified real-data latent effect.
The generator-specific detector checks in Appendix~\ref{app:r19audit} address a separate,
explicitly modeled question.

At this replication depth the uncorrected intervals do not bound noiseless fidelity.
The attempted correction leaves $15$ of $37$ trajectories undefined and its simulated upper
limit falls below the nominal parameter $0.30$ in $18.7$\% of runs. The realized simulation
target differs from that parameter (Appendix~\ref{app:simval}); additional replication alone
does not resolve the target specification or establish calibration.

Two documented analysis requirements were not met: no MDE is printed beside the
descriptive Llama estimates, and the $0.30$ cut is pre-registered but never swept;
both are named here, in this appendix, as the unmet items.

One mid-run batch-composition
correction (skip undefined turns, keep trajectories) is disclosed with its byte-level regression
verification.
\paragraph{A single environment: relevance and limits.}
ALFWorld is a relevant evaluation setting for the adapted credit instruments, not an arbitrary
negative-control environment.\footnote{The implicit-credit instrument family we adapt
\citep{hcapo2603} and the closest step-credit method \citep{stepopsd2605} report headline results
on ALFWorld. StepOPSD evaluates at 1.7B/3B model scale against our 7B/8B policies; this citation
establishes venue relevance only, not comparability of performance.}
That rationale does not establish external validity. Our model scales, instrument adaptations
and single-environment sample limit transfer to other regimes; those regimes require their
own measurements.

\paragraph{The two truncations, and the axis the realized lengths killed.}
The cross-family analysis set excludes $26.8$\% of intervened turns ($n{=}1{,}082$;
corrected instrument\textsuperscript{\ref{fn:coverage}}, whose per-bucket structure is in
Appendix~\ref{app:integrity}). For Qwen, the included-minus-excluded mean policy
log-probability difference is $-0.70$ (95\% bootstrap interval $[-0.81,-0.58]$;
$1{,}768$ included and $266$ excluded turns). This measures a selection shift, not the direction
of bias in rank fidelity. The Llama counterpart has not been recomputed here (membership
recovery: Appendix~\ref{app:superseded}); the families truncate unequally (Qwen excludes
$13.1$\%, $n{=}2{,}034$).

And the family-by-length interaction our design intended to test is unmeasurable here: the realized length distribution put $42$ of $50$ trajectories in the long bucket (Qwen; B1 kept $2$ and B2 kept $3$), so that motivating axis is dead and inherited by future work.
This appendix expands the scope conditions named in Section~\ref{sec:limitations},
keeping their evidence and qualifications together. Historical interpretations are
identified separately from the paper's current conclusions.

\paragraph{Unmet analysis requirements.}
The cross-family amendment requires a minimum detectable effect printed beside every descriptive
Llama estimate. The archived MDE ladder covers $n{=}25$ and $n{=}30$; the realised fidelity $n$ is
$21$, which the ladder does not reach, so no MDE is printed for those estimates and none is
computed here --- the direction is that an MDE at $21$ would exceed the ladder's $n{=}25$ entry.
Separately, the $0.30$ cut of the anti-attenuation rule is the upper band of a three-band rule
fixed verbatim in the pre-registration before the data existed and marked there as not revisable
after seeing them; no sweep over the cut exists in the record, so its warrant is that provenance
and not a calibration.

\paragraph{The target-side placebo we do not run.}
A target-side placebo is a replay target simulated under a null of no per-step effect, with the
archive's count structure preserved (its outcome-class partition and its per-turn replica counts)
and pushed through the identical pipeline. Its purpose is discriminative: it
would show whether the family-versus-control gap reported in Section~\ref{sec:audit} is larger
than what a target carrying no per-step signal produces under the same counts. We did not run it.

\paragraph{The three convergent checks behind the KS2 null.}
Three convergent checks bound how badly the harness could be lying, without substituting for a
validated positive control: an expert-cloning arm at $-10.9$\,pp on held-out
evaluation (exact McNemar $p{=}0.0488$, $n{=}128$ tasks, $29/15$ discordant), measured
behavioural displacement, and off-policy fit.

\paragraph{The estimand's externality, in full.}
At balanced replica counts, the replay
contrast is a positive rescaling of a return-derived step advantage evaluated at the factual
action --- same sign, same zero set, same rank order --- so calling it an external ground truth implied more independence from return-derived credit than the construction delivers; the audit stands: none of the three audited families is return-derived.

\paragraph{Why the family-by-length interaction is unmeasurable.}
The family-by-length interaction the design intended to test is unmeasurable here because the realized length distribution put $42$ of $50$ trajectories in the long bucket (Qwen; the two shorter buckets kept 2 and 3 after amendments), so that motivating axis is dead and inherited by future work.

\section{Divisibility structure of the replay estimand}
\label{app:divisibility}

This section supports the reading of the nonzero replay-contrast fraction reported in
Section~\ref{sec:map}. It introduces no new result, audits no additional signal, and
concerns only the estimand defined in Section~\ref{sec:method}.

\paragraph{What is established elsewhere, and what is ours.}
The counting argument below is assembled from published results. That a group-relative
advantage under binary rewards takes only two values within a group is established at the
response level \citep[\S5.1]{grpofound2606}. That an episode-level advantage cannot
distinguish the contributions of individual actions within a trajectory is stated in
\citet[\S4.2]{gigpo2505} and again in \citet[\S2.5]{survey2604}. That
anchor-state credit is unreliable at small counts is established in
\citet[\S1]{ecpo2606} by a variance route: an action observed once yields an
empirical success rate of $0$ or $1$, and therefore an unreliable point estimate. Their
subject is the reliability of the estimator; ours is the set of values the estimand can
attain, and the two are distinct arguments about the same regime. Our residual is narrow.
We transpose the counting onto the anchor-state group over which our own replay estimand is
defined, and we measure the resulting classes on an executed-replay archive. We did not find
the statement below made for this estimand family in the sources we read.

\paragraph{The estimand and its zero condition.}
At the design used throughout this paper each intervened turn carries one factual arm and
$K = 4$ alternative arms, each replicated $n$ times. Write $s_f$ for the successes on the
factual arm and $M$ for the successes over the whole group of $K+1$ arms. The replay
contrast of Section~\ref{sec:method} then has the common-denominator form
\begin{equation}
T \;=\; \frac{s_f}{n} \;-\; \frac{M - s_f}{K\,n} \;=\; \frac{(K+1)\,s_f - M}{K\,n}.
\label{eq:lattice}
\end{equation}
Both $s_f$ and $M$ are counts, so the numerator moves in unit steps and $T$ takes values on
a lattice of spacing $1/(Kn)$. In particular $T = 0$ exactly when $(K+1)\,s_f = M$: the
contrast vanishes only when the factual arm's success count equals the group's per-arm
average exactly. Whether a turn is able to register a zero at all is therefore fixed by the
counts available at that turn, before any question about the environment is put. We state
this for the estimand of Section~\ref{sec:method}; we make no claim about the value sets of
other step-credit estimators.

\paragraph{Three classes.}
Applying Equation~\ref{eq:lattice} turn by turn partitions the 1{,}768 complete turns under
Qwen2.5-7B into three classes.

\begin{table}[h]
\centering
\begin{tabular}{lrl}
\toprule
class & turns & contrast at this design \\
\midrule
constant pool (all pooled outcomes identical) & 1{,}204 & forced to zero \\
arithmetically forced  &    509  & forced nonzero \\
genuinely stochastic   &     55  & nonzero with probability $0.5055$ \\
\bottomrule
\end{tabular}
\caption{Divisibility classes of the replay estimand at $K=4$ under Qwen2.5-7B: which turns
could register a zero contrast at all, before the environment is consulted. Read the third
column as a property of the counts, not of the task. $n = 1{,}768$ complete turns;
\texttt{runs/collect\_v2/} class counts, E-B1B3.}
\label{tab:divisibility}
\end{table}

\paragraph{What the observed nonzero replay-contrast fraction is consistent with.}
Under the null in which each turn's contrast is whatever Equation~\ref{eq:lattice} forces it
to be, the expected number of turns with a nonzero contrast is $536.80$. The observed number
is $540$ ($0.3054$ of $1{,}768$; Wilson $[0.2844, 0.3273]$). The excess of $3.20$ turns is
$0.862$ null standard deviations, two-sided $p = 0.3884$. At this sample size the observed
nonzero-label fraction is not distinguishable from the fraction the design forces, which is why
Section~\ref{sec:map} reports it as decision points exhibiting a nonzero replay contrast at the
achieved sampling resolution rather than as a measurement of how sparse causal contribution
is. Non-rejection is not acceptance: these data do not establish that every nonzero contrast
is an artifact of the design, only that the aggregate count carries no evidence against that
reading.

\paragraph{How often a zero is unattainable.}
Restricted to the $564$ turns whose pool is not constant, the fraction whose contrast is
forced nonzero by arithmetic alone is $0.9025$ (Wilson $[0.8752, 0.9243]$, $n = 564$ turns;
at the action level $0.9025$, Wilson $[0.8910, 0.9129]$, $n = 2{,}820$ actions). For nine
turns in ten in that stratum the replay contrast could not have come back zero at this
replica budget, whatever the environment did.

\paragraph{Two pool models give two point estimates.}
The same fraction can be reached analytically, from a budget table that generates outcome
pools independently at the archive's marginal success rate. That route returns a different
point estimate, because outcomes in the archive cluster by trajectory rather than arriving
independently. The two routes support the same reading and disagree on the point estimate;
only the archive value above belongs in a statement about our data, and the analytic cell is
a property of the budget model.

\section{The controls gate, and the same question tested as a difference}
\label{app:paired}
\begin{table}[!htbp]
  \centering
  \caption{\textbf{Archived raw correlations and the original interval-overlap control rule (H3).}
  The control preserves the full scored-vector marginal before the replay mask, not necessarily
  the evaluated-subset marginal. The new fixed-mask comparison is in Table~\ref{tab:r19control}. The families are reported
  separately and are not merged; raw estimates and controls are the rows below, the
  sign-agreement splits beside their reading in Appendix~\ref{app:supfigs}. Source: \texttt{A2\_step2\_output.txt}, Route~1 (correction retracted,
  Appendix~\ref{app:simval}); estimator, controls, instruments and data files:
  Appendix~\ref{app:integrity}.}
  \label{tab:ks1verdict}
  \vspace{0pt} %
  \small\setlength{\tabcolsep}{3pt}
  \begin{tabular}{lccc}
    \toprule
    Family & $\hat{\rho}$ (median) [95\% CI] & $n_{\mathrm{eff}}$ & Shuffle [95\% CI] \\
    \midrule
    Implicit (HCAPO $\rho_t$) & $0.0193$ $[-0.109, 0.081]$ & 37 of 50 & $[0.005, 0.114]$ \\
    Judge (Qwen2.5-72B) & $0.1142$ $[0.027, 0.168]$ & 37 (32 comp.) & $[-0.049, 0.117]$ \\
    \bottomrule
  \end{tabular}
\end{table}

The pre-registered controls gate is an interval-overlap rule. Verbatim, from the frozen
pre-registration's decision-rules section:
\begin{quote}
Controls gate: if the shuffled-credit control's $\hat\rho$ CI overlaps the family's $\hat\rho$ CI,
the family's verdict defaults to H3 regardless of point estimate.
\end{quote}
\noindent Its implementation returns that rule string with the verdict, and both families fired
it: \texttt{verdicts.\{implicit,judge\}.rule} $=$ \texttt{"controls\_gate: shuffled-control CI
overlaps family CI"}, verdict \texttt{H3}. \textbf{Overlap of two intervals is not a test of
difference}: the registered overlap rule does not test the paired-difference null and can
return H3 when a direct paired test detects a difference. The rule
is left exactly as registered --- it is the record of what was decided in advance --- and the
difference is tested directly beside it.

For each trajectory $j$ on which both are defined, $\Delta F(j)$ is the signal's within-trajectory
Spearman minus that of its own full-scored-vector shuffle, followed by the replay mask,
on the same trajectory. The
estimator, the membership rule, the interval, the test and the seeds were fixed and committed
before the statistic was computed, together with the three outcomes that would follow from it.

\begin{table}[h]
\centering
\vspace{-6pt} %
\caption{\textbf{Paired within-trajectory difference between each signal and its own matched
shuffled control.} Both intervals contain zero and the original sign-flip tests do not reject,
conditional on one full-vector shuffle followed by masking. This is not the new fixed-mask
CRT (Table~\ref{tab:r19control}). Seed $20260913$; $10{,}000$ draws for each bootstrap and sign-flip test.}
\label{tab:paired}
\vspace{0pt}
\small\setlength{\tabcolsep}{5pt}
\begin{tabular}{@{}lrrrr@{}}
\toprule
family & $n$ & median $\Delta F$ & paired bootstrap 95\% CI & sign-flip $p$ \\
\midrule
implicit & 37 & $-0.103$ & $[-0.219,\ +0.108]$ & $0.375$ \\
judge    & 32 & $+0.026$ & $[-0.029,\ +0.266]$ & $0.390$ \\
\bottomrule
\end{tabular}
\vspace{-6pt} %
\end{table}

The median of the differences is not the difference of the medians, and the two are reported
separately rather than reconciled: for implicit credit the difference of medians is $-0.023$
against a median difference of $-0.103$, and for the judge $+0.057$ against $+0.026$
on the paired eligible subset. The paired
statistic is the one that answers the controls question, because it is the one computed within
trajectory.

\section{Cross-family descriptive partials}
\label{app:r29}

\paragraph{Implicit-credit exploratory association diagnostics.}
The exploratory regression uses five distinct quantities:
raw credit $\rho_t$; the replay contribution $A_{\mathrm{replay}}(t)$, a success-probability
change; the action's fluency $\ell_t$, its mean-token policy log-probability; the \emph{hindsight
log-probability increment} $g_t = \overline{\log\pi}_{\mathrm{hind}} - \ell_t$, what
seeing the outcome adds to the policy's own log-probability of the action --- a log-probability
change, distinct from $A_{\mathrm{replay}}$ throughout this paper; and the credit error
$\Delta_t$. Ordinary least squares of the credit error $\Delta$ on pooled-standardised fluency
and $g_t$, over the $1{,}768$ complete turns of the registered exploratory decomposition,
gives standardised coefficients $0.955$ on fluency and $0.402$ on the hindsight increment: the
gap has a larger standardized coefficient on fluency. Since
$\log q_t=(\ell_t+g_t)/5$ by construction, this association is not evidence of a newly
identified causal mechanism. A separate calculation gives the Pearson partial correlation
between raw credit and the replay contribution given policy log-probability (both residualised
by OLS) is $-0.004$ ($p{=}0.87$, $n{=}1{,}768$) --- after conditioning on likelihood, no
detectable additional association between credit and measured contribution remains.

The hindsight increment $g_t$ shows no detectable correlation with the replay contribution
on the registered Qwen set; the separate raw-credit/replay partial-correlation estimators disagree for Llama
(Appendix~\ref{app:r29}). These diagnostics do not establish absence of causal information.

\looseness=-1 The coefficient ordering is similar in the two audited systems: the fluency-to-increment
ratio is 2.37 for Qwen and 2.35 for corrected Llama (the pre-correction Llama estimate was 2.29;
Appendix~\ref{app:r29}). This is descriptive agreement across these systems, not a general
robustness result.

\paragraph{The sign, read descriptively.}
Descriptively the family point estimate is negative and its shuffle overlaps it, and sign
agreement is chance-level --- the third measure returning the same answer as the first two; we
note the sign and draw no claim from it, the binding gate being the control comparison, not the
zero crossing. The estimates, their intervals and their $n$'s are in the two paragraphs below.

\paragraph{Corrected-instrument family estimate, descriptive.} The family point estimate is
negative with a confidence interval excluding zero ($-0.043$, $[-0.125, -0.016]$; $n{=}21$
trajectories retained of 28\textsuperscript{\ref{fn:coverage}}). Its interval overlaps that of
its own full-score shuffle followed by masking under the registered overlap gate ($-0.010$, $[-0.102, +0.024]$). Sign agreement between implicit
credit and replay is $0.449$ ($[0.393, 0.505]$; $n{=}301$ turn
pairs\textsuperscript{\ref{fn:coverage}}) --- chance-level.

The partial correlations under the corrected instrument are reported descriptively, both
estimators side by side: after conditioning on fluency, the Pearson partial correlation between
implicit credit and the replay contribution is $-0.086$ ($p{=}0.015$, $n{=}792$; Llama, corrected
instrument), while the Spearman partial is $-0.019$ ($p{=}0.591$,
$n{=}792$)\textsuperscript{\ref{fn:coverage}}.
We draw no confirmatory claim from the nominally
significant Pearson value: the recomputation was not registered, the two estimators disagree, and
the missing turns are concentrated non-randomly in the longest trajectories
(Appendix~\ref{sec:integrity}), so a selection effect cannot be excluded.

The coefficient ordering above is descriptive agreement between the two audited systems,
not general robustness across template regimes or a causal attribution.

\section{Analysis sets and a post-hoc common-support sensitivity}
\label{app:analysissets}
The original T3 computes population-SD z-scores (ddof $=0$, no epsilon) on each
trajectory\textquotesingle s credit/replay intersection, then intersects with policy log-probability.
The implementation maps zero or undefined SD to an all-zero vector; this convention was
not specified by the original z-score wording. Spearman uses average ranks for ties, requires
at least three matched turns and drops undefined correlations before taking the trajectory median.
The fidelity analysis additionally excludes trajectories with fewer than four complete replay
turns or constant replay (the registered amendments); structural tests do not apply these exclusions.
The original source functions are \texttt{compute\_deltas}, \texttt{t3\_policy\_logprob},
\texttt{apply\_v32\_exclusion} and \texttt{median\_bootstrap\_ci}.
Code and collection files were checked against archive commit \texttt{3a3f4ca9}; Qwen replay
summaries matched the sealed replay archive, and Llama uses the corrected replay export.
The accompanying audit retains every eligible trajectory ID, turn mask, missingness reason
and source/code hash in \texttt{analysis\_sets.json}.
\begin{table}[H]\centering\small
\caption{Original analysis sets. Entries are median within-trajectory Spearman, percentile
95\% trajectory-bootstrap interval and $n$. T3 was registered; the two likelihood
descriptives were post hoc. The rows have different support and are not a decomposition.}
\label{tab:analysissets}
\begin{tabular}{llrr}
\toprule Model & Association & Median [95\% interval] & $n$ \\
\midrule
Qwen & Gap--likelihood (T3) & $+0.752$ $[+0.647,+0.793]$ & 47 \\
Qwen & Credit--likelihood & $+0.863$ $[+0.819,+0.885]$ & 47 \\
Qwen & Replay--likelihood & $-0.027$ $[-0.135,+0.086]$ & 37 \\
Qwen & Credit--replay fidelity & $+0.019$ $[-0.109,+0.081]$ & 37 \\
\midrule
Llama & Gap--likelihood (T3) & $+0.701$ $[+0.648,+0.768]$ & 27 \\
Llama & Credit--likelihood & $+0.937$ $[+0.922,+0.956]$ & 27 \\
Llama & Replay--likelihood & $-0.034$ $[-0.129,+0.030]$ & 22 \\
Llama & Credit--replay fidelity & $-0.043$ $[-0.124,-0.016]$ & 21 \\
\midrule
\end{tabular}\end{table}
\begin{samepage}
Constant replay reduces T3 to credit--likelihood correlation on 10 Qwen and 5 Llama trajectories.
This is an algebraic consequence of the executed normalization, not independent evidence
about contribution. The remaining count difference for Llama includes a trajectory with
three complete turns, eligible for replay--likelihood but excluded from fidelity by the four-turn rule.
\end{samepage}
All table entries in this appendix round to nearest at the printed precision; some historical prose used
outward-rounded endpoints. These display differences do not represent different input sets.
\paragraph{New post-hoc sensitivity.} The plan was versioned before this calculation, after
prior results had been seen. Within each model, all three associations below use the same
finite credit/replay/likelihood turn intersection and the same trajectories. At least three
joint turns are required; a constant credit, replay, likelihood or gap excludes the trajectory
from all three estimators. Z-scores are recomputed on that intersection with population SD,
without zero filling. The normalization turn sets happen to be unchanged on these inputs;
trajectory eligibility changes. Matching is preserved in one shared trajectory-bootstrap
index matrix per model: $10{,}000$ draws, seed $20260913$, linear percentile interpolation.
These results do not reconstruct the historical common-support medians whose inputs and
procedure were unavailable. No additional test or causal decomposition is made.
\begin{table}[H]\centering\small
\caption{New common-support sensitivity (post hoc), with paired trajectory-bootstrap 95\% intervals.}
\label{tab:commonsupport}
\begin{tabular}{llrr}
\toprule Model & Association & Median [95\% interval] & $n$ \\
\midrule
Qwen & Gap--likelihood (T3) & $+0.675$ $[+0.606,+0.793]$ & 37 \\
Qwen & Credit--likelihood & $+0.879$ $[+0.838,+0.900]$ & 37 \\
Qwen & Replay--likelihood & $-0.027$ $[-0.135,+0.086]$ & 37 \\
\midrule
Llama & Gap--likelihood (T3) & $+0.691$ $[+0.637,+0.748]$ & 22 \\
Llama & Credit--likelihood & $+0.932$ $[+0.898,+0.955]$ & 22 \\
Llama & Replay--likelihood & $-0.034$ $[-0.129,+0.030]$ & 22 \\
\midrule
\end{tabular}\end{table}

\paragraph{Executed score and prompt details.}
The frozen \texttt{score\_implicit.py} scores action-command tokens only:
the separately tokenized action delimiters are masked out of the teacher-forced
log-probability mean. For each defined turn,
$q_t=\exp(\overline{\log p_{\rm hindsight}}/5)$; the denominator of
$\rho_t=\operatorname{clip}(q_t/\overline q,0.8,1.2)$ is the arithmetic mean over
all defined score turns in that trajectory, before replay eligibility filtering.
There is no added epsilon or smoothing in this score. This is not a policy
importance-sampling ratio. The audit uses the HCAPO-style score, not HCAPO training
or its training-token selection.
The dump builder uses the released template printed below, a two-step history
window and an appended outcome line containing the trajectory's own
\texttt{SUCCESS} or \texttt{FAILURE} and the first 500 characters of the last
record's \texttt{obs\_after}. The policy pass omits this hindsight line.
The same recorded action is scored in both passes. These prompt/outcome choices
are adaptations and can affect the diagnostic.
The archived validation reruns the frozen scorer on all 50 archived Qwen hindsight dumps
and compares every stored score at absolute tolerance $10^{-12}$; the maximum
error and source hashes are recorded in the archived validation report.

\paragraph{One frozen shuffle, not a shuffle-averaged estimand.}
The executed control permutes the full scored credit vector within each trajectory
using the frozen seed, before restricting it to replay-complete turns.
It therefore preserves the full vector's marginal distribution exactly, but
need not preserve the marginal on the subsequently retained turn subset.
The archived validation checks identifier uniqueness before invoking the frozen paired analysis;
the latter matches signal and control, rejects missing pairs, and uses each signal's own eligible trajectory set.
It subtracts the two within-trajectory correlations before taking the median;
it does not subtract two medians. The frozen paired bootstrap resamples
trajectories 10,000 times with seed 20260913 and uses sorted endpoints 250 and
9749 (zero-based), without quantile interpolation. It conditions on the one
stored shuffle realization; no extra permutations are introduced by that validation.

\section{Supplementary figures}
\label{app:supfigs}
\begin{figure}[t!]
  \centering
  \includegraphics[width=\linewidth]{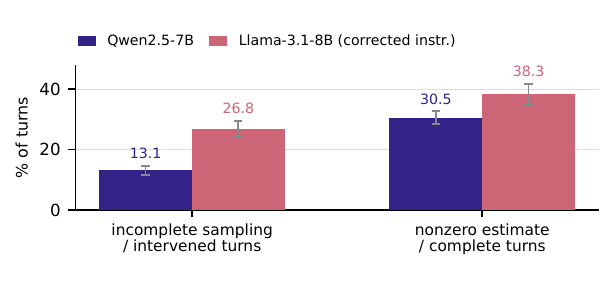}%
\caption{\textbf{Measurability of the replay reference is model-dependent in both directions.}
What is opposite: Llama is the family with the higher rate of intervened turns where four
distinct alternatives cannot be sampled within budget, which makes it the harder family to
audit --- yet where the contrast is defined, Llama registers a nonzero estimate at the higher
rate. Bars carry Wilson 95\% intervals. Rates with intervals, per-rate denominators, the
factual-outcome-variation axis and data paths: Appendix~\ref{app:supfigs}.}
  \label{fig:map}
  \vspace{-8pt} %
\end{figure}

\paragraph{The judge's one above-chance face, and what it does not identify.}
Per-step sign agreement, Wilson 95\%, against the chance line: implicit (HCAPO $\rho_t$)
$50.7$\% $[46.4, 54.9]$ ($267/527$); judge (Qwen2.5-72B) $60.4$\% $[52.1, 68.2]$
($84/139$); chance is $50$\%.

One face of the judge signal does clear chance: its per-step
sign agreement excludes $50$\%, as the counts and intervals above show. It does not identify which steps causally matter: neither family is shown to concentrate credit on nonzero replay-contrast labels above chance (the judge's lift is not shown above the chance line). The registered exploratory lifts, their intervals, and agreement with the anchor judge are below in this appendix.

\paragraph{Precision at the nonzero-contrast label (``pivotal''), by family.}
Throughout this paragraph and the router figures, \emph{pivotal} means the resolution-bounded
label $A_{\mathrm{replay}} \neq 0$ of Section~\ref{sec:method}; the statistics below measure
recovery of that label, and the label's aggregate count is itself consistent with the design's
counting structure (Appendix~\ref{app:divisibility}). The families answer the concentration question
differently (registered exploratory, per family, $n{=}35$ trajectories each, of 50 loaded;
\texttt{runs/collect\_v2/r12\_precision\_at\_pivotal.json}): the implicit family's
precision-at-pivotal lift is $0.940$ $[0.760, 0.997]$, an interval lying entirely below the chance
line of $1.0$, while the judge's $1.000$ $[0.935, 1.001]$ contains it. The judge's sign signal is
not shown to concentrate on the turns that mattered --- its lift interval contains the chance
line; the implicit family's ranking is, if anything, mildly anti-concentrated.

\begin{figure}[h]
  \centering
  \includegraphics[width=\linewidth]{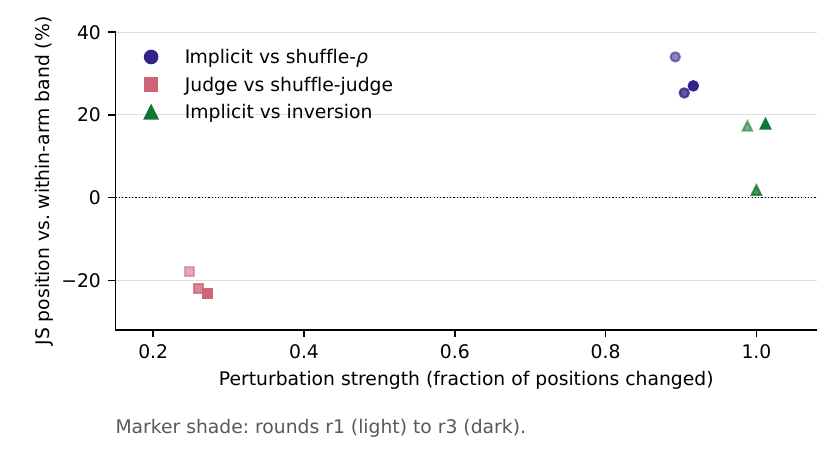}
  \caption{\textbf{Perturbation strength does not order the checkpoints' distributional
  positions: the position readout is not a dose meter for information destruction along the
  magnitude axis.} Each contrast pair's JS position relative to the within-arm band (\%),
  across training rounds r1--r3 (marker shade, lightest r1 to darkest r3), against the
  measured strength of its
  perturbation --- fraction of credit positions actually changed: shuffle-$\rho$ $0.9043$,
  shuffle-judge $0.2602$, inversion $1.0000$ ($n{=}288$ trajectories each;
  \texttt{runs/pc2/a3b\_perturbation\_strength.json}). Position values are the nine
  round-by-contrast cells of \texttt{runs/pc2/a3b\_position\_monotonicity.json}; the
  pre-registered monotonicity test returns $p{=}0.9413$ (10k perm; $n{=}6$ adjacent
  comparisons; grid $3{\times}3$, randomization unit = contrast, $k{=}3$). Four qualifiers
  ride this figure, per its record: the strongest perturbations sit in a saturated regime of
  the readout; separation is monotone in information \emph{destruction}; the non-monotonicity
  is confined to the magnitude axis; and the judge-shuffle pair carries no positional signal.}
  \label{fig:s1}
\end{figure}

\begin{figure}[h]
  \centering
  \includegraphics[width=\linewidth]{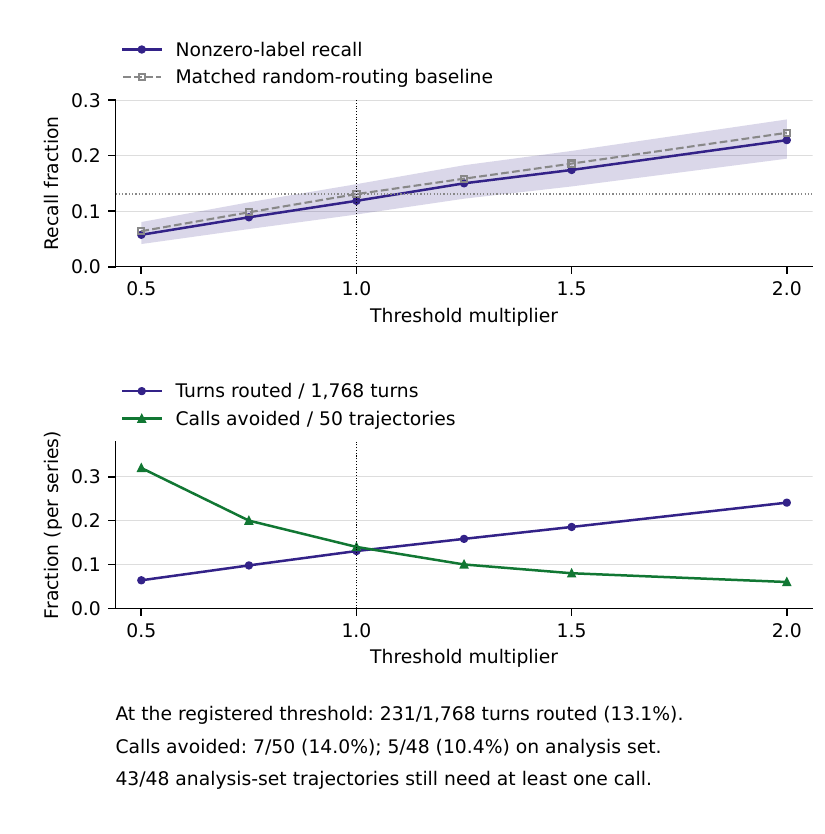}
  \caption{\textbf{[Exploratory --- R\_ROUTE] The confidence router recovers the
  nonzero-contrast label at chance level at every pre-registered threshold --- its recall
  never exceeds the matched random-routing baseline --- and its three cost quantities are
  distinct and never read alone.} Top: recall of labelled turns (label $=$ stored
  $A_{\mathrm{replay}} \neq 0$, resolution-bounded, Section~\ref{sec:method}) across the six
  pre-registered threshold multipliers, with Wilson 95\% band ($n{=}540$ labelled turns at
  $\theta^{*}$); the dashed curve is the matched random-routing baseline (the routed
  fraction), and the dotted line marks the baseline $0.1307$ at the registered threshold
  $\theta^{*}$, against recall $0.1185$ $[0.0939, 0.1485]$ --- chance-level at every
  multiplier. Bottom and separate operating-point account: the three distinct quantities, each labelled with its $\theta^{*}$ value
  and denominator --- turns routed to the judge ($231/1{,}768 = 13.1$\%), judge calls avoided
  per trajectory ($7/50 = 14.0$\% on the registered denominator; $5/48 = 10.4$\% on the
  analysis set), and the $43/48$ trajectories still needing at least one call. Sweep is
  sensitivity only; the single main result is multiplier 1.0. Data:
  \texttt{route\_pack/artifacts/route\_theta\_sweep.json},
  \texttt{route\_pack/route\_retrospective.json} (prereg tag \texttt{1814166\ldots}).}
  \label{fig:s3}
\end{figure}

\paragraph{The cost account.} At the registered threshold, $43$ of $50$ baseline judge calls
remain; the two cost granularities move in opposite directions across the sweep
(Figure~\ref{fig:s3}), so neither substitutes for the other. Two disclosures travel with these
numbers: the per-trajectory figure is stated on the 48-trajectory analysis set (seed010 and
seed020 have no complete turns), because that is the set every other per-trajectory quantity here
is stated on; the frozen denominator of $50$ is the flattering one and gives $7/50 = 0.14$, against
$5/48 = 0.104$ on the analysis set; and the
confidence input is a per-token geometric mean, which runs high for
multi-token actions. Route to save money, not to find
the steps that matter.

\paragraph{The cross-family measurability rates, in prose.} Under the same environment, the same $K{=}4$ alternatives, and the same 15-rollout budget, the
contrast is undefined for want of four distinct alternatives under the sampling budget at
13.1\% of intervened turns for Qwen2.5-7B
($n{=}2{,}034$) but at 26.8\% for Llama-3.1-8B ($n{=}1{,}082$; corrected
instrument\textsuperscript{\ref{fn:coverage}}) --- a factor of 2.05, with non-overlapping
intervals. And where the replay reference \emph{is} defined, the same comparison reverses: the family
that is harder to measure registers a nonzero contrast at a higher rate, 38.3\% of Llama's complete turns carrying the nonzero replay-contrast label
against Qwen's 30.5\%. The full three-rate map is Figure~\ref{fig:mapfull} in
Appendix~\ref{app:supfigs}; the two divergences run in opposite directions.

The Wilson 95\% intervals are $[11.7, 14.6]\%$ (Qwen) and $[24.2, 29.5]\%$ (Llama).

The nonzero-label reversal is the sharper of the two: 38.3\% of Llama's complete
turns ($n{=}792$; corrected instrument\textsuperscript{\ref{fn:coverage}}) against Qwen's
30.5\% ($n{=}1{,}768$). The no-variation structure, by contrast, transfers approximately --- $\sigma_{\mathrm{floor}} = 0$ (no observed factual-outcome variation) at 80.6\% versus 86.0\% of turns. Measurability structure therefore varies with the model in both directions: the family that is harder to define counterfactuals for is the one that registers a nonzero contrast at a higher rate where counterfactuals exist, consistent with its lower rate of turns with no observed factual-outcome variation.

\paragraph{Figure~\ref{fig:map}'s not-plotted apparatus.} The
  factual-outcome-variation axis is near-flat across families and is reported here rather than
  plotted:
  $\sigma_{\mathrm{floor}}{=}0$ at $86.0$\% $[84.3, 87.6]$ of Qwen's and $80.6$\%
  $[77.7, 83.2]$ of Llama's complete turns. The vertical axis is a per-rate percentage: each
  bar is a fraction of its own rate's denominator, given above, not of a denominator shared
  across the two groups. For Qwen's $1{,}184$ both-arms-all-zero turns, Section~\ref{sec:method}
  reports unexcluded $|\Delta p| \leq 0.632$ at one-sided 95\%; that numerical bound
  is not asserted for every exact-zero contrast or for Llama. Data: \texttt{measurability\_counts\_\{ks1,xfam\_redo\}.json}; full three-axis map: Figure~\ref{fig:mapfull} (appendix).

\begin{figure}[h]
  \centering
  \includegraphics[width=\linewidth]{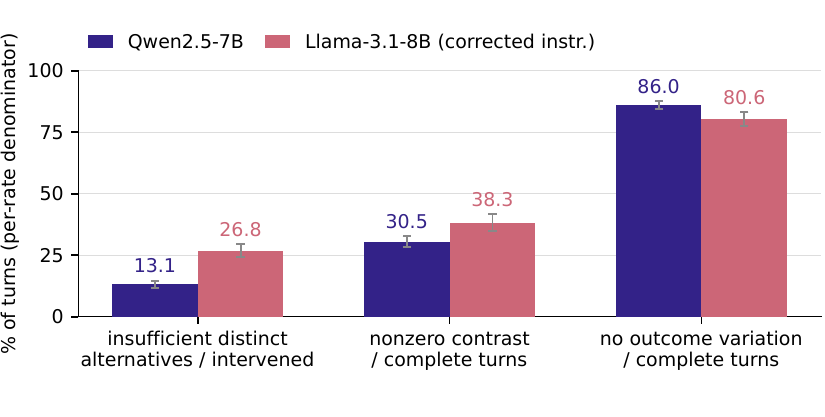}
  \caption{\textbf{Where the per-step replay reference can be measured depends on the model --- in both directions.}
  For each family, three fractions of the replay map: turns with insufficient distinct
  alternatives under the sampling budget
  (Qwen2.5-7B 13.1\%, $n{=}2{,}034$ intervened turns; Llama-3.1-8B 26.8\%, $n{=}1{,}082$, corrected
  instrument\textsuperscript{\ref{fn:coverage}}), turns with a nonzero contrast
  ($A_{\mathrm{replay}} \neq 0$: 30.5\%, $n{=}1{,}768$; 38.3\%, $n{=}792$), and turns with no
  observed factual-outcome variation
  ($\sigma_{\mathrm{floor}}{=}0$: 86.0\%, $n{=}1{,}768$; 80.6\%, $n{=}792$).
  Error bars are Wilson 95\% intervals throughout --- Qwen: $[11.7, 14.6]$, $[28.4, 32.7]$,
  $[84.3, 87.6]$; Llama: $[24.2, 29.5]$, $[34.9, 41.7]$, $[77.7, 83.2]$ respectively.
  The large divergence is the first bar --- Llama lacks four samplable alternatives at twice
  Qwen's rate --- yet the nonzero-contrast fraction runs the \emph{other} way, so neither family is uniformly harder to audit.
  Every zero-valued contrast is a finite-resolution observation, not proof of no effect.
  The $0.632$ bound in Section~\ref{sec:method} applies specifically to Qwen's
  $1{,}184$ both-arms-all-zero turns.
  Data: \texttt{measurability\_counts\_ks1.json}, \texttt{measurability\_counts\_xfam\_redo.json}.}
  \label{fig:mapfull}
\end{figure}

This is a recorded single-family association, not an out-of-sample validation. For Qwen,
observed completion is 0.650 in the lowest and 0.981 in the highest predicted decile, but drops
between the eighth and ninth deciles; it is not monotonically increasing. Across all 2,034
turns, the aggregate prediction misses observed completion by approximately 20 percentage
points. The occupancy model's uniform-residual assumption does not establish calibration on
these observations, and no pre-GPU estimate of measurable coverage is claimed for either family.

\paragraph{The six Holm-corrected confirmatory tests, all reported.} The pre-registration fixes
the confirmatory set as three structural tests in two families --- T1 credit-error against turn
position, T2 against action type, T3 against policy log-probability --- and corrects them together
by Holm--Bonferroni at family-wise $\alpha = 0.05$. The adjustment quoted in
Section~\ref{sec:mechanism} is only checkable if all six are shown, so all six are here. Three
clear the correction and three do not; the one the pre-registration named in advance, T3 for the
implicit family, is among those that clear it and in the registered direction.

\begin{table}[htbp]
  \centering
  \caption{\textbf{The six confirmatory structural tests, Holm--Bonferroni at family-wise
  $\alpha = 0.05$.} The registered prediction was T3-positive for the implicit family; it is the
  only directional prediction the pre-registration made. Verdicts key on the Holm-adjusted column:
  judge T1 and judge T3 have raw $p$ below $0.05$ and do not survive the correction at $m{=}6$,
  which is what the correction is for.}
  \label{tab:holm}
  \footnotesize
  \begin{tabular}{lllll}
    \toprule
    Family & Test & raw $p$ & Holm-adjusted & Outcome \\
    \midrule
    Implicit & T1 $\Delta \times$ turn position   & $0.3014$ & $0.3014$ & not significant \\
    Implicit & T2 $\Delta \times$ action type     & $0.0001$ & $0.0006$ & significant \\
    Implicit & T3 $\Delta \times$ policy log-prob & $0.0002$ & $0.0008$ & \textbf{significant; registered direction} \\
    Judge    & T1 $\Delta \times$ turn position   & $0.0194$ & $0.0582$ & not significant \\
    Judge    & T2 $\Delta \times$ action type     & $0.0001$ & $0.0006$ & significant \\
    Judge    & T3 $\Delta \times$ policy log-prob & $0.0444$ & $0.0888$ & not significant \\
    \bottomrule
  \end{tabular}
\end{table}

\paragraph{Pre-registered secondaries of the judge audit, reported.} Agreement of the primary
judge (Qwen2.5-72B) with the anchor judge (GPT-4.1, the audited method family's own judge, frozen
11-trajectory selection) is exact $74.1$\% (Wilson 95\% $[68.0, 79.4]$, $n{=}224$ turns),
Cohen's $\kappa = 0.496$, pooled Spearman $0.484$
(\texttt{runs/collect\_v2/judge\_anchor\_agreement.json}) --- the judge-family credit signal is
substantially judge-model-sensitive. The self-preference check (self-minus-cross $\rho$ scorer)
detected none: $-0.011$, CI $[-0.077, 0.064]$
(\texttt{runs/collect\_v2/sec7\_self\_preference.json}). The family $\times$ length interaction
registered alongside them is not testable in this run (Section~\ref{sec:limitations}).

\clearpage
\section{Resolution of the replay reference at sub-full budgets}
\label{app:resolution}

Section~\ref{sec:map} reports that at the full design a per-turn interval excludes zero at
$24$ of the $1{,}768$ complete turns. This section gives the design grid that number sits in,
and four cautions about reading it. It runs no new experiment: the archive is frozen, and every
cell is an exhaustive enumeration of the within-turn subsamples of it, so each entry is an exact
expectation over subsamples rather than a resample estimate, and no turn's outcomes are ever
pooled with another turn's.

\paragraph{Why trajectory-level statistics remain readable when single turns are not.}
The fidelity statistics of Section~\ref{sec:audit} are
within-trajectory rank correlations over the complete turns of each retained trajectory: their
power comes from the ordering across turns, not from any one turn's interval. The attenuation
this per-turn unreliability induces is what the reliability correction was designed to remove
--- but that correction failed its own pre-registered validation and is reported as a
sensitivity analysis, so the fidelity estimates of Section~\ref{sec:audit} are read uncorrected
(Appendix~\ref{app:simval}).

\begin{table}[h]
\centering
\footnotesize\setlength{\tabcolsep}{5pt}
\begin{tabular}{rrrrrrr}
  \toprule
  \multicolumn{3}{c}{design} & \multicolumn{3}{c}{P(pivotal label flips vs.\ full data)} & resolvable \\
  \cmidrule(lr){1-3}\cmidrule(lr){4-6}\cmidrule(lr){7-7}
  $n$ & $K$ & $N$ & forced & stochastic & all & P(CI excl.\ 0) \\
  \midrule
    $1$ & $1$ & $2$ & $0.6977$ & $0.4606$ & $0.2152$ & $0.0000$ \\
    $1$ & $2$ & $3$ & $0.5452$ & $0.4347$ & $0.1705$ & $0.0000$ \\
    $1$ & $3$ & $4$ & $0.4380$ & $0.4295$ & $0.1395$ & $0.0342$ \\
    $1$ & $4$ & $5$ & $0.3544$ & $0.4301$ & $0.1154$ & $0.0261$ \\
    $2$ & $1$ & $4$ & $0.5431$ & $0.4197$ & $0.1694$ & $0.0249$ \\
    $2$ & $2$ & $6$ & $0.3618$ & $0.3828$ & $0.1161$ & $0.0083$ \\
    $2$ & $3$ & $8$ & $0.2378$ & $0.3710$ & $0.0800$ & $0.0115$ \\
    $2$ & $4$ & $10$ & $0.1485$ & $0.4036$ & $0.0553$ & $0.0135$ \\
    $3$ & $1$ & $6$ & $0.4450$ & $0.3455$ & $0.1389$ & $0.0095$ \\
    $3$ & $2$ & $9$ & $0.2459$ & $0.3030$ & $0.0802$ & $0.0143$ \\
    $3$ & $3$ & $12$ & $0.1164$ & $0.2455$ & $0.0411$ & $0.0140$ \\
    $3$ & $4$ & $15^{\dagger}$ & $0.0000$ & $0.0000$ & $0.0000$ & $0.0136$ \\
  \bottomrule
\end{tabular}

\caption{Resolution of the replay reference across the twelve $(n, K)$ design cells, Qwen2.5-7B,
$n = 1{,}768$ complete turns. $N = n(1{+}K)$ is the rollout budget per decision point. The three
flip-rate columns give the probability that a subsample at that design assigns a turn the
opposite nonzero replay-contrast label to the one the full data assign it; the constant-pool stratum is
identically $0$ at every cell and is omitted rather than given a column of zeros. The last
column is the fraction of turns whose 95\% Newcombe interval excludes zero.
$^{\dagger}$The full design is the reference, not a measurement: its subsample space has one
element, so its flip rates are $0$ by construction. Generated from
\texttt{resolution\_curve\_summary.json} by \texttt{appendix\_resolution\_table.py}.}
\label{tab:resolution}
\end{table}

\begin{figure}[h]
\centering
\includegraphics{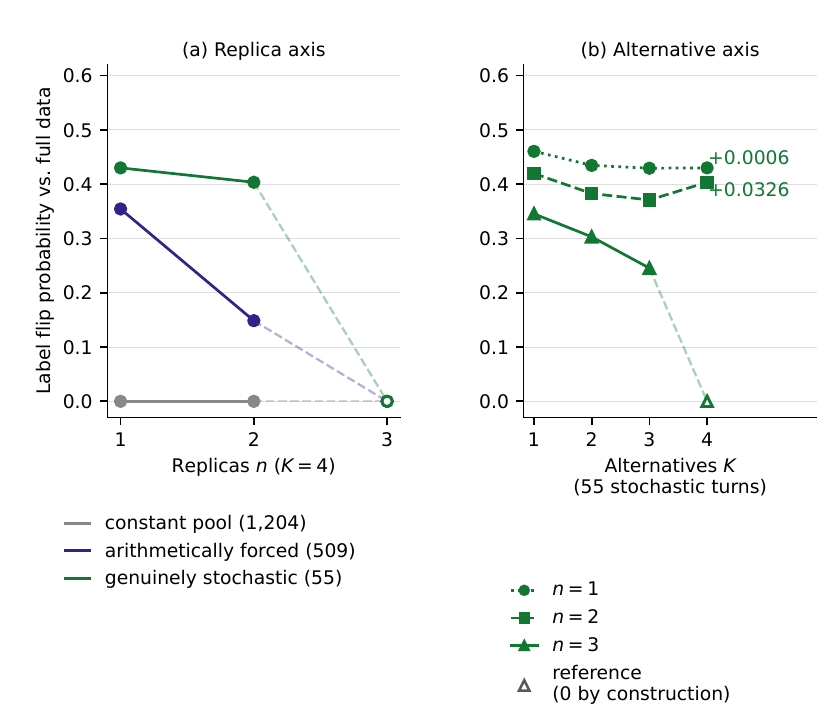}%
\caption{\textbf{Subsampling sensitivity within the archive: replica count buys resolution; a
fourth alternative does not always}, and at $n{=}1,2$ it \emph{raises} the flip rate by the
labelled amounts. Every cell is an exact enumeration of within-turn subsamples of the frozen
archive, not a new experiment. Hollow markers are the $(n{=}3,K{=}4)$ reference, $0$ by
construction. Grid and cautions: Table~\ref{tab:resolution} and the paragraphs below.}
\label{fig:resolution}
\end{figure}

\paragraph{The full design is the reference, not a measurement.}
At $(n{=}3, K{=}4)$ the subsample space has exactly one element --- the full data --- so the flip
rate, the estimate drift and the class-migration rate are all $0$ by construction rather than by
convergence. The rightmost point of every curve in Figure~\ref{fig:resolution} and the daggered
row of Table~\ref{tab:resolution} are the reference the other eleven cells are scored against.
Neither is evidence that the design has converged; read them that way and the quantity inverts.
The enumeration requires the design to be exactly $(n{=}3, K{=}4)$, and it is: all $1{,}768$
complete turns carry three factual replicas and four alternatives of three replicas each, which
we assert rather than assume --- Section~\ref{sec:method}'s ``at least three times'' states the
pre-registered floor, and the realized archive sits on it at every turn.

\paragraph{$N$ is not a sufficient statistic.}
Along the replica axis the flip rate falls monotonically in $n$ at every fixed $K$, in every
stratum, with no violations; that is the clean finite-rollout axis, and
Figure~\ref{fig:resolution}(a) is it at $K{=}4$. Total budget behaves differently. Ordering the
twelve cells by $N$ produces seven pairs in which the larger budget carries the higher flip rate,
and every one of the seven crosses a change in $n$. Two designs that cost the same do not
therefore resolve the same, and a resolution curve must not be collapsed onto a single
rollout-count axis.

\paragraph{More alternatives do not always help.}
Along the $K$ axis the flip rate is monotone for the constant-pool and arithmetically-forced
strata, but not for the $55$ turns whose contrast the environment actually decides: a fourth
alternative raises it, from $0.4295$ to $0.4301$ at $n{=}1$ and from $0.3710$ to $0.4036$ at
$n{=}2$; this is a lattice effect and not sampling noise. The pivotal rule is
$\widehat{\Delta}_t \neq 0$ evaluated exactly on rationals, and $K$ changes which values the
pooled alternative mean can take, so it changes which turns are able to register a zero at all
--- the mechanism of Appendix~\ref{app:divisibility}, met along a second axis. We report it
unsmoothed. The principle that governs is the budget pre-registration's verification tier
T3(ii), which requires a monotonicity violation to be reported as a finding rather than
smoothed; we note that this is the principle and not a literal clause match, since T3(ii) is
written about the forced fraction along $n$, and what fails to be monotone here is the flip rate
along $K$.

\paragraph{The interval is ours, and the last column is sensitive to it.}
The pre-registration fixes the pivotal rule and the classification; it does not fix a per-turn
confidence interval. The 95\% Newcombe hybrid score interval behind the final column of
Table~\ref{tab:resolution} was chosen by this analysis, after the archive was frozen, and the
choice is ours to defend. It is defined at $n{=}1$, which is why that column is populated across
the whole grid, and its width dominates the column's behaviour: the resolvable fraction is
monotone along neither axis, standing at $0.0342$ at $(n{=}1, K{=}3)$ against $0.0136$ at the
full design. We report the column because the sentence in Section~\ref{sec:map} needs it and
because withholding it would make the choice invisible --- not because the ordering across cells
is interpretable. Only the full-design row carries that sentence.

The alternative arm pools $12$ rollouts from four different actions. Conditional on this
sampled set and independent continuations, heterogeneous action probabilities give a sum of
binomials, not a single binomial. Its conditional count variance is no greater than the
binomial variance at the pooled mean. This comparison of variance models does not establish
nesting or coverage of finite-sample Newcombe intervals. Equal replication makes the pooled
mean and the mean of per-alternative means the same random variable, hence they have the
same variance under the same probability model. Resampling the alternative set introduces
additional randomness, which the reported interval does not model. Thus $24/1{,}768=1.36$\%
is the observed fraction resolved by this named interval procedure, not an upper or lower
bound on the fraction another procedure would resolve. Calibration for the heterogeneous
sampled-set design remains unestablished.

\paragraph{Where the resolvable turns sit.}
The $24$ turns whose interval excludes zero are not spread evenly over the partition of
Appendix~\ref{app:divisibility}: none of them is a constant-pool turn, $21$ of $509$ are
arithmetically forced ($4.13$\%, Wilson $[2.71, 6.22]$), and $3$ of $55$ are genuinely stochastic
($5.45$\%, Wilson $[1.87, 14.85]$). The class the design advice of
Section~\ref{sec:artifacts} concerns is the smallest and the least resolved, and its interval is
wide enough that the two nonzero rates are not distinguishable here.

\paragraph{The unstratified median is degenerate.}
Of the $1{,}768$ complete turns, $1{,}204$ are constant-pool, where the contrast is identically
zero at every design; the median absolute estimate over all turns is therefore $0.0000$ at every
cell of the grid. That is a fact about the partition and not a resolution result, and it appears
here only with this caveat attached to it. The per-class curves carry the information, which is
why every panel and every column above is stratified.

\section{Step 2: correcting rank fidelity for replay-target reliability}
\label{app:step2}

We retract the corrected bound: a simulation whose decision rule was fixed before it ran shows the
corrected upper limit falling below the nominal generator parameter $0.30$ in
$18.7$\% of runs, against the $5$\% tolerance that rule allowed. The mean realized target is
different; all target readings appear in Appendix~\ref{app:simval}. The reading rule
does not survive that, and does not reach the uncorrected statistic either. The
pre-registration fixes what the rule is applied to, before any result exists: it defines the
``upper bound'' the rule consumes as the corrected quantity, quoted exactly in
Appendix~\ref{app:step2} (\S3.2, tag \texttt{prereg-reliability-stage0-v3}). That quantity is the
corrected one by definition, so with the correction withdrawn the rule has no input, and we force no band onto a
quantity it does not name. The correction, its scope and the trajectory selection it rests on are
reported as a sensitivity analysis (Appendix~\ref{app:step2}). %
Read mechanically anyway, the uncorrected percentiles fire the middle band, which directs
reporting both readings and recommending nothing; the two values and that reading are in
Appendix~\ref{app:simval}, reported rather than recommended.
The simulation illustrates attenuation under its generator assumptions; it does not calibrate
the latent correlation behind an observed value. No corrected fidelity bound is established
(Route~1, Appendix~\ref{app:simval}).

\noindent\textbf{Status: sensitivity analysis, not a bound.} Everything in this appendix is
reported as measured, but no claim in the paper rests on it. A pre-registered simulation of this
correction (Appendix~\ref{app:simval}) failed the decision rule fixed before it ran, so the
corrected bound and the band reading taken from it are retracted; Section~\ref{sec:audit} reports
the uncorrected estimates instead.

\paragraph{What the rule is applied to, as the pre-registration fixed it.}
Before any result existed, \S3.2 of the reliability pre-registration defined the quantity the
decision rules consume:
\begin{quote}
``Upper bound'' is defined now, before any result: the 97.5th percentile of the bootstrap
distribution of $\rho_{\mathrm{obs}} / \sqrt{\rho_{xx}}$ (two-sided 95\%), propagating uncertainty
in BOTH $\rho_{\mathrm{obs}}$ and $\rho_{xx}$. The decision rules in \S3.4 are applied to that
upper bound.
\end{quote}
\noindent (\S3.2, tag \texttt{prereg-reliability-stage0-v3}.) That quantity is the corrected one,
so the retraction of the correction leaves the rule with no input --- which is what
Section~\ref{sec:audit} reports.

The archived analysis compares each credit family against one full-scored-vector shuffled
control, subsequently masked, against the same sampled target. This paired comparison measures an observed ranking
increment; it does not guarantee a direction of noise bias or bound latent noiseless fidelity.
The pre-registered second step attempted a reliability correction on the Qwen2.5-7B set under the original instrument (37 trajectories;
32 for the judge family after 5 trajectories with undefined Spearman are dropped). Per trajectory,
the target's reliability $\rho_{xx}$ is the within-trajectory intraclass correlation, ICC(2,3), of
the three-replica mean. The three replicates are replicates of the \emph{contrast}, not of one arm:
replicate $r$ is the $r$-th factual rollout minus the mean of the $r$-th replica of each of the four
alternatives, so their mean is $A_{\mathrm{replay}}(t)$ itself and ICC(2,3) is the reliability of
that $15$-rollout contrast, not of a single arm's mean. The trajectory's observed Spearman is
divided by $\sqrt{\rho_{xx}}$; the
family median is taken over trajectories with defined reliability, and trajectories with
$\rho_{xx} \le 0$ are reported undefined and counted, never clamped. The correction is one-sided, and its scope is pre-registered in terms marked liftable;
the following is a historical quotation, whose bound interpretation has been retracted:
\begin{quote}
We disattenuate for measurement error in the \textbf{ground truth only}, never in the credit signal. Formally we set $\rho_{yy} = 1$: the correction applied is $\rho_{\mathrm{obs}} / \sqrt{\rho_{xx}}$ rather than the classical $\rho_{\mathrm{obs}} / \sqrt{\rho_{xx}\,\rho_{yy}}$. This is a \textbf{scoping choice, not an assumption about the signal's reliability.} The credit signal's own noise is not error to be corrected away, because the noisy signal is precisely what a training run consumes: a practitioner who adopts an instrument gets its noise along with it. Correcting for $\rho_{yy}$ would answer a question nobody can act on --- how well an \emph{infinitely reliable} version of the instrument would track truth --- whereas the auditable question is how well \emph{this} instrument, as shipped, tracks truth measured as well as we can measure it. The resulting quantity is therefore an \textbf{upper bound on the fidelity attributable to the instrument as deployed}, and we report it as a bound.

The direction of this choice is stated rather than left to the reader: correcting for target unreliability raises the estimate, and correcting for signal unreliability would raise it further, so $\rho_{yy}=1$ is the stopping point at which the null is stated at its strongest; the verdicts apply to the instrument as deployed, not to a latent signal a more reliable elicitation might recover.
\end{quote}
\noindent (Pre-registration \S B2, tag \texttt{prereg-reliability-stage0-v3}.) The reported bound
is the 97.5th percentile of a $10{,}000$-draw bootstrap of the corrected median, and the
pre-registered rule fires when that bound exceeds $0.30$. The threshold is not chosen here: it is
the upper cut of a three-band rule --- below $0.15$ the null is strengthened, above $0.30$ the
claim is downgraded, between them nothing is recommended and the decision escalates --- copied
verbatim into the pre-registration (\S3.4, tag \texttt{prereg-reliability-stage0-v3}) before the
data it governs existed, and marked there as not revisable after seeing them. Its warrant is that
provenance rather than a calibration: no sweep over the cut exists (Section~\ref{sec:limitations}).

\paragraph{Which band fired, and why that reading is withdrawn.} The rule has three bands and
we quote it rather than paraphrase:
\begin{quote}
If the anti-attenuation upper bound stays below $0.15$ for both signals: the null is
\textbf{strengthened}. If the upper bound exceeds $0.30$ for either signal: the claim must be
\textbf{downgraded} to ``not detectable at this target reliability'', and the abstract must change
with it. If it lands between $0.15$ and $0.30$: \textbf{report both readings, recommend nothing,
escalate.}
\end{quote}
\noindent \textbf{This reading is retracted, and what it was is stated rather than deleted.}
The rule is quantified over the pair. The judge family's primary bound is $0.532$, above $0.30$,
and the rule's second clause reads ``exceeds $0.30$ for \emph{either} signal''; the downgrade band
would therefore require downgrading the claim for the pair. That interpretation was withdrawn
after validation of the correction failed: the simulation of Appendix~\ref{app:simval} places the upper limit below the nominal
generator parameter $0.30$ in $18.7$\% of runs against the $5$\% its rule allowed; the realized
target differs from that parameter. No band is
read in this paper, and Section~\ref{sec:audit} states what the rule does and does not reach.
Three readings of the implicit bound exist and are retained here as sensitivity values only, the
first being the one that was the pre-registered primary: the literal rule gives $0.174$; the
machine-zero sensitivity gives $0.128$; and marginalizing over replica pairings, a robustness
check ordered after the fact and \textbf{not} a registered treatment, gives a median of $0.126$
with a $95$\% range of $0.075$--$0.171$. The band boundaries are as registered and are not
adjusted here; they are no longer applied.

\paragraph{What the selection does before the correction does anything.} The corrected median is
taken over the $22$ trajectories with $\rho_{xx} > 0$, and the $15$ dropped are exactly those
where estimated target reliability is zero or negative. These estimates do not
establish the magnitude of latent attenuation or a corresponding corrected value.
Selection alone changes the uncorrected implicit median from $+0.019$ over all $37$
trajectories to $-0.018$ over the defined $22$, before any disattenuation is applied.
The correction remains withdrawn; the selected population is not interchangeable
with the original one.

\paragraph{The replica pairing, and why it does not carry the result.} Each turn's three
contrast replicates pair the $r$-th factual rollout with the $r$-th replica of each alternative,
one of $1{,}296$ possible pairings; the mean of the three is invariant to that choice but their
spread, and so ICC(2,3), is not, and the pre-registration records that replica index is not a
crossed factor. Over $1{,}000$ independent random pairings, recomputing $\rho_{xx}$ and both
bounds under each, the implicit bound has median $0.1255$ ($[0.0747, 0.1711]$, maximum $0.2956$,
never reaching the rule's threshold) and the judge bound median $0.4489$ ($[0.3352, 0.5916]$); the
asymmetry holds in $998$ of $1{,}000$ draws, with no sign reversal in any of them. The two
exceptions are draws in which the judge bound also falls below threshold, so the asymmetry weakens
only by the judge joining the implicit family, never by the implicit family escaping its bound.
These are historical upper-limit sensitivities, not calibrated bounds. The index-pairing
upper limit $0.1738$ exceeds the reported $97.5$th percentile $0.1711$ of this sweep.
The sweep varies both pairing and bootstrap seeds, so its range includes Monte Carlo variation.

The pre-registration named two routes for $\rho_{xx}$ and required both to be reported if they
disagreed. They disagree, and both are reported (Table~\ref{tab:step2}). Route~2, a bootstrap over
replicas, fails its known-answer test: on synthetic blocks constructed with true reliability exactly
zero it returns $0.4785$, and its floor is analytic rather than empirical --- a bootstrap resample
mean is $m = \bar{x} + e$ with $\mathrm{var}(\bar{x}) = \sigma^2_w/3$ and
$\mathrm{var}(e) = (2/9)\,\sigma^2_w$, so $\mathrm{corr}(m_1, m_2) \to
(1/3)/((1/3)+(2/9)) = 0.600$ as the true between-turn variance goes to zero. Route~2 therefore
cannot return a value near zero and cannot produce an undefined trajectory, which is why it reports
$0$ undefined where Route~1 reports $15$ (implicit) and $11$ (judge). No bound or claim in the paper
rests on a Route~2 value; its rows are reported because the pre-registration requires it.

Route~1 is reported under two treatments. The literal pre-registered rule keeps every trajectory
with $\rho_{xx} > 0$. One trajectory (\texttt{seed007}) has a reliability that prints as zero at six
decimals and is a machine zero (the mean squares are identically equal); under the literal rule it
enters the median with a corrected value of order $10^6$, and under the sensitivity treatment
($\rho_{xx} > 10^{-12}$) it moves to the undefined list, one trajectory's worth of difference in
each family. Treating it as zero moves the implicit family's upper bound from $0.17$ to $0.13$ and
the judge family's from $0.53$ to $0.49$; the judge family's bound exceeds the rule's threshold
under both treatments, so the rule's firing on the judge family does not depend on this
floating-point question. Restricted to the same defined trajectories, the uncorrected medians are
$-0.017588$ (implicit, $n{=}22$) and $0.134952$ (judge, $n{=}21$); corrected and uncorrected values
are different quantities on different $n$ and are never interchanged.

\begin{table}[t]
  \centering
  \caption{\textbf{Step 2: corrected rank fidelity, Qwen2.5-7B, original instrument.} Each row is
  a family median of per-trajectory Spearman with its bootstrap 95\% interval, except the
  separately labelled replica-pairing panel; the upper
  bound is the interval's upper edge (97.5th percentile), on which the pre-registered rule keys.
  Route~1 is the primary; Route~2 rows are reported as the pre-registration requires and are
  disqualified for any bound or claim (known-answer failure, floor $0.600$). Every corrected row
  in this table is a sensitivity value: the pre-registered validation of Appendix~\ref{app:simval}
  failed, so no bound and no band reading is taken from any of them. Uncorrected rows are the
  published medians of Section~\ref{sec:audit} on their own $n$. Undefined = trajectories with
  $\rho_{xx} \le 0$ (or $\le 10^{-12}$ under the sensitivity treatment), counted and excluded from
  the median. Two Route~2 cells coincide with their uncorrected counterparts and are not carry-over:
  Route~2 never returns an undefined trajectory (its floor is $0.600$), so its implicit corrected
  median is taken at a trajectory whose reliability estimate is $1$ and equals the uncorrected
  median exactly, and the judge interval's lower edge is unmoved while its upper edge is not. Both
  are as the run artifact records them. Source: \texttt{A2\_step2\_output.txt} (archived analysis,
  commit \texttt{3e00d92}; sha256 \texttt{aa33b9b1\,\ldots\,8721}).}
  \label{tab:step2}
  \small\setlength{\tabcolsep}{3pt}
  \begin{tabular}{@{}lp{0.30\linewidth}rrrl@{}}
    \toprule
    Family & Estimate & Median & $n$ & Bootstrap 95\% CI & Undefined \\
    \midrule
    Implicit & uncorrected (published) & $0.0193$ & 37 & $[-0.109, 0.081]$ & --- \\
    Judge    & uncorrected (published) & $0.1142$ & 32 & $[0.027, 0.168]$ & --- \\
    \midrule
    Implicit & Route 1, literal rule ($\rho_{xx} > 0$), primary & $-0.034$ & 22 & $[-0.295, 0.174]$ & 15 \\
    Implicit & Route 1, machine-zero sensitivity & $-0.093$ & 21 & $[-0.300, 0.128]$ & 16 \\
    Implicit & uncorrected over the \emph{defined} 22 (selection only) & $-0.018$ & 22 & --- & 15 \\
    Judge    & Route 1, literal rule ($\rho_{xx} > 0$), primary & $0.328$ & 21 & $[-0.081, 0.532]$ & 11 \\
    Judge    & Route 1, machine-zero sensitivity & $0.285$ & 20 & $[-0.162, 0.486]$ & 12 \\
    \midrule
    Implicit & Route 2 (disqualified; reported per pre-registration) & $0.0193$ & 37 & $[-0.137, 0.110]$ & 0 \\
    Judge    & Route 2 (disqualified; reported per pre-registration) & $0.130$ & 32 & $[0.027, 0.201]$ & 0 \\
    \bottomrule
  \end{tabular}
  \par\smallskip\begin{tabular}{lrr}
  \toprule
  Replica-pairing sensitivity & Median upper limit & 2.5--97.5\% range \\
  \midrule
  Implicit, post hoc & $0.126$ & $[0.075,0.171]$ \\
  \bottomrule
  \end{tabular}
  \par\smallskip\raggedright\small
  The separate panel is a distribution of upper limits over $1{,}000$ pairings, not a
  parameter estimate with a confidence interval. Its reported quantiles use sorted index
  $\lfloor qB\rfloor$; pairing-specific defined/undefined memberships were not archived.
  The index-pairing counts $22/15$ therefore do not apply to this panel.
  Source: \texttt{PAIRING\_SWEEP\_output.txt} and \texttt{a2\_step2\_pairing\_sweep.py}.
\end{table}

\section{Validating the reliability correction: a pre-registered simulation}
\label{app:simval}

\paragraph{The mechanical band reading, reported and not recommended.}
Read mechanically anyway, and
reported rather than recommended: the uncorrected $97.5$th percentiles are $0.0808$ (implicit) and
$0.1681$ (judge) --- not below $0.15$ for both, not above $0.30$ for either --- so the band that
would fire is the middle one, which directs reporting both readings and recommending nothing
(Table~\ref{tab:step2}, Appendix~\ref{app:simval}).

The correction of Appendix~\ref{app:step2} divides an observed Spearman by $\sqrt{\hat\rho_{xx}}$,
where $\hat\rho_{xx}$ is an ICC(2,3) estimated from three replicates, and it \emph{selects} on that
estimate: $\hat\rho_{xx} \le 0$ is undefined, which takes $37$ trajectories to $22$. Whether that
procedure recovers a correlation it is given had never been tested. We tested it, and it failed.

\paragraph{What was fixed before anything ran.} The pre-registration was written and committed at
\texttt{2026-09-11T23:37:55Z} (tag \texttt{prereg-correction-validation-2026-09-11}), before the
first simulation run. Two amendments were appended at \texttt{23:38:31Z} and \texttt{23:41:28Z},
both also before the first run and both disclosed in place: the first because a void condition had
been written as an exact equality on a random variable, the second because the void check did its
job and rejected the generator as first specified --- resampling an observed $15$-rollout rate as
if it were the underlying probability inflated the all-zero mass to $0.69945$ against the
archive's $0.66968$, outside the $99$\% interval the amendment had just fixed. Neither amendment
touched the decision rule. The results were committed at \texttt{2026-09-12T00:19:46Z}. The rule,
quoted from \S6 of that document:
\begin{quote}
\textbf{INVALID} if at $\rho_{\mathrm{true}} = 0.30$ the corrected 95\% upper bound falls below
$0.30$ in \textbf{more than 5\%} of runs. Consequence: the abstract \textbf{retracts the corrected
bound} and reports the uncorrected estimate with its interval; the correction is demoted to a
sensitivity analysis, stated with its assumptions and this simulation.
\end{quote}
\noindent The generator draws its turn counts, its tie structure and its latent effect
distribution from the archive itself, and the pipeline's own functions are imported unmodified
rather than reimplemented --- \texttt{icc\_components}, both $\rho_{xx}$ routes, the $37 \to 22$
selection rule, and the bootstrap that produces the $97.5$th percentile. Only the per-trajectory
Spearman is written for the simulation, because the pipeline consumes it from
\texttt{analysis\_summary.json} and never computes it. The zero-signal void checks passed: $1{,}768$
complete turns, mean $15$-rollout success $0.1178$, all-zero fraction $0.66968$, membership $37$.

\paragraph{The registered decision.} At nominal parameter $\rho_{\mathrm{true}} = 0.30$ the
corrected upper limit falls below that parameter in $18.7$\% of runs, against the $5$\%
tolerance the rule allowed. This is upper-limit false exclusion, not two-sided noncoverage;
the upper endpoint is the $97.5$th percentile, whose nominal tail probability is $2.5$\%.
The recorded decision is \textbf{INVALID}: the corrected bound does not support
the registered interpretation; the uncorrected estimate and its uncertainty
remain the reported evidence.

\paragraph{What the uncorrected numbers are consistent with.}
At a
nominal generator parameter of $0.30$ --- mean realized correlation $0.245$ ---
the mean uncorrected estimate over trajectories where it is defined is $0.127$, against the
judge's observed $0.1142$; the uncorrected interval covers the nominal parameter in $0.05$\%
of runs. This simulation illustrates attenuation under its assumptions; it is not an inverse
calibration that identifies the latent fidelity behind the observed judge correlation.

\paragraph{The estimand the validation left under-specified, and both readings.} The
pre-registration's \S3 asserts the generator induces a within-trajectory Spearman equal to
$\rho_{\mathrm{true}}$, and its \S6 rule targets the parameter; the realized column shows the
assertion fails after discretization ($0.245$ at $0.30$), so the validation's estimand was
under-specified --- parameter and realized value were assumed to coincide and do not. Both
readings are therefore reported, at the decision cell ($\rho_{\mathrm{true}}{=}0.30$,
corrected estimator): against the \emph{parameter}, the pre-registered computation, Route~1
false exclusion is $0.1865$ ($373/2000$; coverage $0.811$) --- the rule fired and the
retraction stands. Against each draw's own \emph{mean realized} Spearman, Route~1 false exclusion
is $0.0500$ ($100/2000$; coverage $0.940$), sitting exactly at the rule's boundary --- not
above it --- and against the grand-mean realized value ($0.245$) it is $0.0585$. Route~2
fails under every target ($0.9160$ against the parameter; $0.5600$ against either realized
reading). No reading yields a clear pass, the pre-registered reading fails outright, and the
correction accordingly remains a sensitivity analysis; we do not re-derive the retraction on
the boundary number. The realized target averages within-trajectory correlations over all
generated trajectories where the latent correlation is defined, before observed-correlation
or reliability selection. The corrected estimate is instead a median over the selected set;
these target populations and aggregation rules are not interchangeable. The grand-mean
target further averages the draw-level means and is estimated from these same simulations.
Table~\ref{tab:simtargets} gives all three targets comparable prominence and Wilson Monte
Carlo intervals. One deviation is disclosed: the pre-registration's \S4 specifies
per-replicate hash seeds, while the committed code used one serial generator per
$\rho_{\mathrm{true}}$ cell (\texttt{sim\_correction\_validation.py:158}); reproduction is
unaffected --- the re-run reproduces every committed aggregate exactly.

\begin{table}[ht]\centering\small
\caption{Three target readings at nominal parameter $0.30$. False exclusion means upper
limit below target; coverage means inclusion in the two-sided 95\% interval. Wilson 95\%
Monte Carlo intervals quantify the exclusion rates, conditional on the named fixed target
or draw-level rule; they do not account for estimation of the grand-mean target itself.
The historical registered decision used the parameter row. No row establishes calibration.}
\label{tab:simtargets}
\begin{tabular}{llrrr}
\toprule Route & Target & Exclusions & Rate [MC 95\%] & Coverage \\
\midrule
1 & Nominal parameter (historical) & 373/2000 & $0.1865$ $[0.1700,0.2042]$ & $0.8110$ \\
1 & Draw mean, before selection & 100/2000 & $0.0500$ $[0.0413,0.0604]$ & $0.9400$ \\
1 & Grand mean of draw means & 117/2000 & $0.0585$ $[0.0490,0.0697]$ & $0.9295$ \\
\midrule
2 & Nominal parameter (historical) & 458/500 & $0.9160$ $[0.8884,0.9373]$ & $0.0840$ \\
2 & Draw mean, before selection & 280/500 & $0.5600$ $[0.5162,0.6029]$ & $0.4400$ \\
2 & Grand mean of draw means & 280/500 & $0.5600$ $[0.5162,0.6029]$ & $0.4400$ \\
\midrule
\end{tabular}\end{table}

\paragraph{What the comparison establishes.} Against the nominal parameter $0.30$, upper-limit
false exclusion falls from $0.9995$ uncorrected to $0.1865$ corrected. This is a change in a
selection-and-coverage diagnostic, not a measured fraction of bias removed. It establishes
neither calibrated coverage of a specified population Spearman nor a bound on real-data fidelity.

\paragraph{Three things the simulation shows that the archive alone could not.}
\emph{(i) The corrected statistic is not bounded.} A Spearman correlation cannot exceed $1$;
between $4.59$\% and $8.37$\% of corrected per-trajectory values do, across the five
$\rho_{\mathrm{true}}$ settings, with a maximum of $8.219 \times 10^{7}$ under Route~1. That is the
same pathology the run artifact already displays --- values of order $10^{6}$ entering the median
(Appendix~\ref{app:step2}) --- measured here as a rate rather than noticed as an outlier.
\emph{(ii) The band is not diagnostic at the bottom of its range.} With $\rho_{\mathrm{true}}$ set
to exactly $0.00$, the $0.15$--$0.30$ band fires in $44.7$\% of runs and the $>0.30$ band in
$5.1$\%; the band a zero signal fires is the band the implicit family fired.
\emph{(iii) Selection differs.} The archive's selection is more severe than the
generator reproduces: the archive leaves $22$ of $37$ trajectories defined ($59$\%), the generator
a mean of $25.7$ to $25.9$ out of a mean $35.7$ with a defined Spearman ($72$\%). The
selection populations differ, so the simulated rates need not transfer to the archive.
This discrepancy does not establish a lower or upper bound on the archive's failure rate.

\paragraph{Route 2.} Route~2 --- the replica bootstrap, which Appendix~\ref{app:step2} already
disqualifies on a known-answer failure --- places its upper limit below the nominal parameter $0.30$ in
$91.6$\% of runs. It ran under one pre-registered deviation: $200$ replica-bootstrap draws on
$500$ replicates per $\rho_{\mathrm{true}}$ rather than the paper's $2{,}000$ draws, because the
full design costs $18.9$ hours. The deviation was registered with its own falsifier --- a
spot-check at the paper's $2{,}000$ draws on $50$ replicates at $\rho_{\mathrm{true}} = 0.30$,
invalid if it moved mean $\hat\rho_{xx}$ by more than $0.02$. It moved it by $+0.002833$
($0.665444 \to 0.668277$), so the reduction stands. Route~1 ran at the paper's full $10{,}000$
bootstrap draws with $2{,}000$ replicates per setting and no reduction in bootstrap draws;
the RNG execution deviation is disclosed above. The decision keys on
Route~1, as registered.

\paragraph{Historical simulation table definitions.}\label{app:simdefinitions}
\textbf{Correction-validation simulation: the corrected estimator against a known
  generator parameter.} Route~1 is the pipeline's declared primary and the one the pre-registered decision rule
  keys on. \emph{realized} is the mean Spearman the generator actually induces at the given
  $\rho_{\mathrm{true}}$ (attenuated by the archive's own tie structure). \emph{cov} is the
  fraction of runs whose $95$\% interval contains $\rho_{\mathrm{true}}$; \emph{false excl} is the
  fraction whose upper edge falls below it --- the quantity the rule keys on. \emph{unc(all)} is the
  uncorrected median over every trajectory in the replicate with a defined Spearman --- a mean of
  $35.7$ of the $37$-trajectory membership set, the artifact's key being
  \texttt{mean\_uncorrected\_all37} --- and \emph{unc(def)} is the same restricted to those
  surviving the selection.
  $n$~def is the mean number of trajectories surviving $\hat\rho_{xx} > 0$, and is read against a
  mean of $35.7$ trajectories with a defined Spearman, not against $37$; the archive's own value is
  $22$ of $37$. $\Delta$sel is the shift that selection alone produces in the uncorrected statistic.
  $|\hat\rho_c| > 1$ is the fraction of corrected per-trajectory values outside the range a
  correlation can occupy, and \emph{real.} is \emph{realized}. The boxed cell records the historical decision.
  The printed \emph{cov} and \emph{false excl} columns are the pre-registered computation,
  against the generator parameter; the same decision cell evaluated against the realized
  Spearman (per-draw $0.0500$; grand-mean $0.0585$; Route~2 fails every target) is reported in
  the estimand paragraph above. Source:
  \texttt{sim\_correction\_validation\_out/RESULTS.txt} (archived output; sha256
  \texttt{e37cd80b\,\ldots\,0bb48}); dual-target evaluation:
  \texttt{sim\_target\_sensitivity\_out/}.

\begin{table}[t]
\centering\small\setlength{\tabcolsep}{4pt}
\caption{\textbf{Historical parameter-target simulation, Route 1.}
The upper panel reports corrected statistics; the lower panel separates uncorrected
statistics and selection. All original cells are retained. Definitions and source
are in Appendix~\ref{app:simdefinitions}; comparable target sensitivities are in
Table~\ref{tab:simtargets}.}
\label{tab:simval}
\begin{tabular}{@{}rrrrrrr@{}}
\toprule
Parameter & Realized & Corrected mean & Coverage & False excl. & $n$ def & $|\hat\rho_c|>1$ \\
\midrule
$0.00$ & $0.001$ & $0.0007$ & $0.940$ & $0.0315$ & $25.7$ & $0.0459$ \\
$0.10$ & $0.083$ & $0.0760$ & $0.936$ & $0.0510$ & $25.9$ & $0.0459$ \\
$0.20$ & $0.163$ & $0.1479$ & $0.875$ & $0.1160$ & $25.9$ & $0.0510$ \\
$0.30$ & $0.245$ & $0.2216$ & $0.811$ & \fbox{$0.1865$} & $25.9$ & $0.0587$ \\
$0.50$ & $0.410$ & $0.3660$ & $0.584$ & $0.4160$ & $25.8$ & $0.0837$ \\
\bottomrule
\end{tabular}
\par\medskip
\begin{tabular}{@{}rrrrrr@{}}
\toprule
Parameter & Unc(all) & Unc(def) & Coverage & False excl. & $\Delta$sel \\
\midrule
$0.00$ & $0.0000$ & $0.0006$ & $0.953$ & $0.0240$ & $+0.0006$ \\
$0.10$ & $0.0413$ & $0.0430$ & $0.666$ & $0.3340$ & $+0.0017$ \\
$0.20$ & $0.0821$ & $0.0858$ & $0.130$ & $0.8700$ & $+0.0037$ \\
$0.30$ & $0.1224$ & $0.1270$ & $0.001$ & $0.9995$ & $+0.0045$ \\
$0.50$ & $0.2052$ & $0.2123$ & $0.000$ & $1.0000$ & $+0.0071$ \\
\bottomrule
\end{tabular}
\end{table}

\begin{table}[t]
\centering\small\setlength{\tabcolsep}{4pt}
\caption{\textbf{Historical parameter-target simulation, Route 2.}
The upper panel reports corrected statistics; the lower panel separates uncorrected
statistics and selection. All original cells are retained. Definitions and source
are in Appendix~\ref{app:simdefinitions}; comparable target sensitivities are in
Table~\ref{tab:simtargets}.}
\label{tab:simvalroute2}
\begin{tabular}{@{}rrrrrrr@{}}
\toprule
Parameter & Realized & Corrected mean & Coverage & False excl. & $n$ def & $|\hat\rho_c|>1$ \\
\midrule
$0.00$ & $-0.002$ & $-0.0035$ & $0.954$ & $0.0240$ & $35.7$ & $0.0114$ \\
$0.10$ & $0.081$ & $0.0503$ & $0.814$ & $0.1840$ & $35.7$ & $0.0100$ \\
$0.20$ & $0.165$ & $0.1015$ & $0.442$ & $0.5580$ & $35.7$ & $0.0122$ \\
$0.30$ & $0.243$ & $0.1476$ & $0.084$ & $0.9160$ & $35.8$ & $0.0135$ \\
$0.50$ & $0.410$ & $0.2513$ & $0.000$ & $1.0000$ & $35.8$ & $0.0114$ \\
\bottomrule
\end{tabular}
\par\medskip
\begin{tabular}{@{}rrrrrr@{}}
\toprule
Parameter & Unc(all) & Unc(def) & Coverage & False excl. & $\Delta$sel \\
\midrule
$0.00$ & $-0.0028$ & $-0.0028$ & $0.952$ & $0.0260$ & $-0.0000$ \\
$0.10$ & $0.0411$ & $0.0411$ & $0.674$ & $0.3260$ & $+0.0000$ \\
$0.20$ & $0.0826$ & $0.0826$ & $0.124$ & $0.8760$ & $+0.0000$ \\
$0.30$ & $0.1207$ & $0.1207$ & $0.004$ & $0.9960$ & $-0.0000$ \\
$0.50$ & $0.2054$ & $0.2055$ & $0.000$ & $1.0000$ & $+0.0000$ \\
\bottomrule
\end{tabular}
\end{table}

\paragraph{What we do not claim from this.} The simulation tests the correction and exposes
its limitations; it does not establish calibration. It does not show that implicit latent
fidelity is zero or that judge latent fidelity is nonzero. No corrected bound is licensed.
Greater replication would require a new precision and calibration assessment
(Section~\ref{sec:limitations}).

\clearpage
\section{The seven-arm results, read arm by arm}
\label{app:ks2table}
The primary analysis retains all three seeds; the instrument note and the exploratory
sensitivity do not alter that membership.
\begin{table}[H]
\centering
\vspace{-6pt} %
\caption{\textbf{Every pre-registered confirmatory comparison, on the full three-seed set the
KS2 pre-registration specifies, is inconclusive under the $\pm3$\,pp band, and no arm reliably
beats the untrained base policy (0.422; 54/128).} At the realized seed-pair variances the
one-sided minimum detectable effects are $7.8$--$36.4$\,pp, so moderate effects were
undetectable by design. $\dagger$\,arm\,5 seed\,2 carries the format-collapse instrument note
(Appendix~\ref{app:instrument}) and remains in the primary analysis; estimators, data and the
drop-seed sensitivity: Appendix~\ref{app:ks2table}.}
\label{tab:sevenarm}\label{tab:ks2arms}
\vspace{0pt} %
\small\setlength{\tabcolsep}{3pt}
\begin{tabular}{@{}lccccc@{}}
\toprule
arm & s0 & s1 & s2 & mean & between-seed SD \\
\midrule
1\; outcome-only      & 0.438 & 0.445 & 0.438 & 0.440 & 0.5\,pp \\
2\; implicit ($\rho$) & 0.352 & 0.414 & 0.344 & 0.370 & 3.9\,pp \\
3\; judge             & 0.359 & 0.531 & 0.445 & 0.445 & 8.6\,pp \\
4\; shuffled-$\rho$   & 0.375 & 0.383 & 0.422 & 0.393 & 2.5\,pp \\
5\; inverted-$\rho$   & 0.406 & 0.391 & 0.094$^\dagger$ & 0.297$^\dagger$ & 17.6\,pp$^\dagger$ \\
6\; pivotal-masked    & 0.469 & 0.344 & 0.320 & 0.378 & 8.0\,pp \\
7\; shuffled-judge    & 0.445 & 0.375 & 0.406 & 0.409 & 3.5\,pp \\
\midrule
base policy (no adapter) & \multicolumn{3}{c}{---} & 0.422 & (54/128) \\
\bottomrule
\end{tabular}

\vspace{0.6em}

\begin{tabular}{@{}llccccc@{}}
\toprule
id & comparison & test ($H_0$) & effect & 95\% paired t-CI & raw $p$ & Holm-adj.\ $p$ \\
\midrule
C1 & implicit vs shuffled    & TOST ($|\Delta| \geq 3$\,pp) & $-2.34$\,pp & $[-15.9, +11.2]$ & 0.4273 & 1.00 \\
C2 & judge vs shuffled-judge & TOST ($|\Delta| \geq 3$\,pp) & $+3.65$\,pp & $[-26.4, +33.7]$ & 0.5326 & 1.00 \\
C3 & implicit vs outcome     & TOST ($|\Delta| \geq 3$\,pp) & $-7.03$\,pp & $[-15.5, +1.4]$  & 0.9116 & 1.00 \\
C4 & judge vs outcome        & TOST ($|\Delta| \geq 3$\,pp) & $+0.52$\,pp & $[-19.9, +20.9]$ & 0.3265 & 1.00 \\
C5 & inverted vs implicit    & superiority ($\Delta \geq 0$) & $-7.29$\,pp & $[-46.6, +32.0]$ & 0.2543 & 1.00 \\
C6 & masked vs outcome       & superiority ($\Delta \leq 0$) & $-6.25$\,pp & $[-26.5, +14.0]$ & 0.8422 & 1.00 \\
\bottomrule
\end{tabular}
\vspace{-6pt} %
\end{table}
Read arm by arm, implicit credit trained $2.3$\,pp below its own shuffled control
(raw TOST $p{=}0.43$) and $7.0$\,pp below outcome-only training (raw TOST $p{=}0.91$);
both Holm-adjusted values are $1.00$. Neither comparison establishes equivalence;
the observed uncertainty intervals describe the effects this design leaves unresolved.

Table~\ref{tab:sevenarm} reports the full seven-arm outcome that Section~\ref{sec:training}
summarizes: final-round win rates per arm and seed, and the six
pre-registered confirmatory comparisons under the $\pm 3$\,pp
equivalence band, with the raw $p$ each test produced and the Holm-adjusted $p$ printed
separately. Every comparison is inconclusive; no pre-registered decision criterion is met.
The arm~5 seed~2 evaluation carries the
format-collapse instrument note of Appendix~\ref{app:instrument} and \emph{remains in} the
primary analysis. The KS2 pre-registration contains no exclusion rule; that rule belongs
to the later, non-confirmatory analysis plan. The registered-exploratory sensitivity that
drops seed~2 flips C5's sign: $-7.29$\,pp ($p{=}0.2543$, three seed pairs) becomes
$+1.56$\,pp ($p{=}0.6211$, two seed pairs, 95\% t-CI $[-48.1, +51.2]$\,pp --- below the
pre-registered minimum for any claim, and reported here only as sensitivity).

\paragraph{The table's estimators and flagged seed.}
The table's own title, as the caption gave it, is \emph{Seven-arm training outcome}. The
detection floor that belongs to the printed comparisons is the seed-paired one: one-sided
minimum detectable effects of $7.8$--$36.4$\,pp at $\alpha{=}0.05$ and $19.3$--$89.7$\,pp at
the Holm entry threshold (\texttt{KS2\_PAIRED\_TCI\_MDE\_output.txt}); the
$\approx$11.8\,pp figure is the single-seed 128-task McNemar bound and lives with its ladder
in Appendix~\ref{app:mde}. ``Inconclusive'' is the registered reading, never ``no
difference''.
Top: final-round win rates on
the fixed 128 held-out tasks (greedy; $n{=}128$ per arm-seed;
seed-paired CRN seeds \{0,1,2\}). $\dagger$\,arm\,5 seed\,2 carries the
format-collapse instrument note of Appendix~\ref{app:instrument} and remains in the primary
analysis; the printed arm-5 mean and SD include it and
inherit the flag, and the registered-exploratory sensitivity dropping it
leaves all six comparisons inconclusive. Bottom: paired comparisons (C1--C4 TOST equivalence;
C5--C6 one-sided superiority; Holm over the six, adjusted values printed in the table). Data:
\texttt{runs/ks2/eval/final/<arm>/<seed>/eval.json} (win rates),
\texttt{runs/ks2/eval/r1/base\_control/s0/eval.json} (base),
\texttt{runs/ks2/CONFIRMATORY\_RESULT.json} (comparisons).

\section{The MDE ladder}
\label{app:mde}
Two detection floors live here, at two different analysis units, and they are not
interchangeable. The first is \emph{per-seed, task-level resolution}: exact McNemar on one
seed's 128 (or 96) paired task outcomes, the unit at which a single training run can be read.
The second is the floor of the tests Section~\ref{sec:training} actually reports:
\emph{seed-paired t on three seed pairs}, whose minimum detectable effects at the realized
paired SDs are an order larger. Both carry their $(n, \mathrm{unit})$ labels.
\begin{center}
\begin{tabular}{lcc}
  \toprule
  \multicolumn{3}{l}{\emph{Per-seed, task-level (exact McNemar, single seed)}} \\
  $n$ (split) & discordant pairs & MDE \\
  \midrule
  96 (train, 1 seed) & $\sim$21 & $\sim$13.8\,pp \\
  128 (held-out, 1 seed) & $\sim$28 & $\sim$11.8\,pp \\
  384 (3 seeds pooled)$^{*}$ & $\sim$84 & $\sim$7.0\,pp \\
  \bottomrule
\end{tabular}
\hspace{1.5em}
\begin{tabular}{lccc}
  \toprule
  \multicolumn{4}{l}{\emph{Seed-paired t, 3 pairs (one-sided, 80\% power)}} \\
  id & SD (pp) & MDE @ $.05$ & @ $.05/6$ \\
  \midrule
  C1 & 5.5 & 12.6 & 31.0 \\
  C2 & 12.1 & 27.8 & 68.6 \\
  C3 & 3.4 & 7.8 & 19.3 \\
  C4 & 8.2 & 18.9 & 46.5 \\
  C5 & 15.8 & 36.4 & 89.7 \\
  C6 & 8.2 & 18.7 & 46.2 \\
  \bottomrule
\end{tabular}
\end{center}
$^{*}$The pooled row treats three seeds as $384$ independent pairs; the realized between-seed
variance the training runs exhibited makes that an optimistic reading, and the seed-paired
column is the one the confirmatory tests answer to. Lower block:
\texttt{KS2\_PAIRED\_TCI\_MDE\_output.txt} (\texttt{ks2\_paired\_tci\_mde.py}); realized paired
SDs from \texttt{runs/ks2/CONFIRMATORY\_RESULT.json}.

\section{The pre-registered partition decomposition (E031(b))}
The frozen record's summary of this experiment is one sentence, quoted rather than paraphrased;
it and the two caveats that travel with it wherever it appears are below. Two further
checkpoints with elevated canonicalization-failure counts were flagged per the pre-registration,
not averaged in, and the format-collapsed arm-seed (Appendix~\ref{app:instrument}) was excluded
from all primary analysis \emph{of this family}, per this family's own pre-registration.

\label{app:e031b}

\paragraph{What the checkpoints agree on, and what they do not.}
The checkpoints are not interchangeable. On held-out states the seven arms agree in discrete choices --- measured gaps $+0.41$\,pp ($p{=}0.343$, Stage 1) and $+0.50$\,pp
($p{=}0.185$, Stage 2) against the frozen A3B bar of $\geq 5$\,pp at $p<0.05$ --- while their action \emph{distributions} separate cleanly: between-arm Jensen--Shannon divergence exceeds
within-arm by $2.6\times$ ($0.0520$ against $0.0198$; restricted-permutation $p{=}0.0001$) ---
credit moved where the policies put their probability, without (yet) moving what they do.

\paragraph{Historical summary (interpretation superseded), with its original caveats.}
The following frozen sentence is preserved as a historical quotation, not the current attribution of checkpoint differences: \emph{``Credit source leaves a measurable signature in the policy's output
distribution --- arms cluster by instrument family (rho-based vs judge-based), not by
information content --- yet this signature does not reach greedy action selection: checkpoints
agree on actions at the same rate whether or not they share a credit rule.''} Two caveats travel with it wherever it appears. The masked arm moved far less than any other
(adapter norm $1.37$ against ${\approx}4.7$ for the dense arms), so its position in any
clustering is confounded with how little it moved --- the divergence result survives removing it ($0.0212$ within-arm against $0.0421$ between-arm; $n{=}16$ and $n{=}80$ checkpoint
pairs). And the Stage-2 canonicalizer initially normalized raw generations without extracting the action, producing spuriously unique strings; the error is retained on the record; every number here is from the corrected pass.

The pre-registration fixed two partitions of the seven training arms before computing the
checkpoint-pair matrix: one by credit instrument and one by whether the credit carried task
information. The instrument partition met the registered criteria in both stages:
on the 500-state bank, between-partition minus within-partition JS is $+0.0346$
($p{=}0.0002$; $n_W{=}30$, $n_B{=}141$ cross-arm checkpoint pairs).
The information partition did not show a detectable separation (Table~\ref{tab:e031b}).
This partition-level association does not show that credit content was irrelevant.
Credit rules changed both training content and effective optimization dose.
Checkpoint distributions differed, but without a dose-matched comparison those differences
cannot be attributed either to credit content or specifically to dose.

\begin{table}[h]
  \centering
  \caption{\textbf{The pre-registered instrument partition separates the checkpoints; the
  information partition does not.} Each row reports mean between-partition minus mean
  within-partition JS divergence over cross-arm checkpoint pairs, under the partition named in
  column 1; positive means checkpoints sharing a group are more similar to each other than to
  the other group. Partitions, criterion, and the reading to record were frozen in
  \texttt{a3b-prereg-v1} before the matrix was seen; each partition's $p$ is a restricted
  permutation test under the pre-registered within-seed scheme (10{,}000 draws), the scheme
  attribution following the pre-registration. Both partitions divide the same 171
  cross-arm pairs; the split differs, so the two rows carry different $n_W/n_B$.
  Data: \texttt{a3b\_stage1\_structure.json}, \texttt{a3b\_stage2\_structure.json}.}
  \label{tab:e031b}
  \begin{tabular}{llrrr}
    \toprule
    Partition & Stage / statistic & $\Delta$ (B$-$W) & $p$ & $n_W / n_B$ \\
    \midrule
    Instrument  & Stage 1, turn-1 canonical agreement & $+0.0291$ & $0.0124$ & $30 / 141$ \\
    Information & Stage 1, turn-1 canonical agreement & $-0.0077$ & $0.7205$ & $36 / 135$ \\
    Instrument  & Stage 2, 500-state bank JS          & $+0.0346$ & $0.0002$ & $30 / 141$ \\
    Information & Stage 2, 500-state bank JS          & $-0.0126$ & $0.9777$ & $36 / 135$ \\
    \bottomrule
  \end{tabular}
\end{table}

\paragraph{Dose associations without causal separation.}
Credit rules that zero out many turns drop those turns from the training batch --- one
filter line in the update step --- so sparser credit buys fewer optimizer steps and smaller
parameter displacement. Controlling for realised parameter-change magnitude leaves a residual instrument correspondence statistically indistinguishable from zero in this design (partial Mantel $\rho{=}+0.078$ on realised credit scale,
$p{=}0.774$), alongside the zero-order credit-sparsity association of $+0.912$ ($p{=}0.0001$; $n{=}21$ arm
pairs, non-independent, both tests).

\paragraph{What the association cannot establish.}
The
pair set is small and non-independent, the association is robust to removing the sparsest arm;
the design remains a partial-correlation design --- no $\lVert\Delta W\rVert$-matched
comparison exists. A significant marginal association and a nonsignificant conditional one
are not a causal mediation demonstration.

\section{Post-hoc control sensitivity and known-probability validation}
\label{app:r19audit}

\paragraph{Fixed-mask comparison.}
The plan was versioned after seeing the historical findings but before executing this extension.
The two natural eligible Qwen sets, $37$ implicit and $32$ judge trajectories, exactly match
their archived paired sets. Five additional replay-eligible judge trajectories have constant
retained original scores; they still consume the original full-vector shuffle stream before
the correlation is marked undefined. Original index assignments are reconstructed from the
frozen seed and ordered inputs and checked against archived correlations, not claimed to
have been independently stored as permutations.

Let $r_j$ be the observed within-trajectory Spearman and $s_{jb}$ the retained-score shuffled
Spearman. We report $\operatorname{median}_j(r_j-s_{jb})$ for each of $10{,}000$ draws;
its $2.5/97.5$ percentiles describe random-control sensitivity. They are not trajectory
confidence intervals. For fixed nonconstant score and reference vectors,
$E_P[s_{jP}]=0$: centering average ranks gives zero expected permuted score rank at each
position, while both rank norms remain fixed. This holds with ties and is checked by
exact enumeration. Hence the mean-control diagnostic is exactly median raw Spearman;
its $95\%$ interval resamples trajectories $10{,}000$ times. It is not the old paired estimand.
The positive-alignment CRT compares $\operatorname{median}_j r_j$ with
$\operatorname{median}_j s_{jb}$, using the add-one inclusive tail. Raw $p$ values are
$0.31017/0.00120$ and Holm-adjusted values $0.31017/0.00240$ for implicit/judge.
This tests conditional label exchangeability, not a causal null.

The observed alignment is $T_{\mathrm{obs}}=\operatorname{median}_j r_j$;
each repeated paired increment is $D_b=\operatorname{median}_j(r_j-s_{jb})$.
Table~\ref{tab:r19control} keeps the three objects separate.
\begin{table}[!htbp]
\centering\small
\caption{\textbf{Sampling uncertainty and control variation are distinct.} Estimate brackets are $95\%$ trajectory-bootstrap CIs (unadjusted for observed alignment). The shuffle column is a central $95\%$ range across retained-score controls, not a CI. The judge range starts at zero, not strictly above it. The last column is the positive fixed-mask permutation test, Holm-adjusted over two signals.}
\label{tab:r19control}
\setlength{\tabcolsep}{3pt}
\begin{tabular}{lcccc}
\toprule
Signal ($n$) & \shortstack{Archived paired\\increment} & \shortstack{Observed\\alignment} & \shortstack{Variation across\\retained-score shuffles} & \shortstack{Adjusted\\permutation $p$} \\
\midrule
Implicit (37) & \shortstack{$-0.103$\\$[-0.219,+0.108]$} & \shortstack{$+0.019$\\$[-0.109,+0.081]$} & $[-0.077,0.089]$ & $0.3102$ \\
Judge (32) & \shortstack{$+0.026$\\$[-0.029,+0.266]$} & \shortstack{$+0.114$\\$[+0.027,+0.168]$} & $[0.000,0.192]$ & $0.0024$ \\
\bottomrule
\end{tabular}
\end{table}

\paragraph{Order and single-control sensitivity.}
The additional $2{,}000$ full-vector shuffles preserve the full marginal, but change the
evaluated-subset multiset in $95.9\%$ of implicit and $76.8\%$ of judge trajectory-draws.
Implicit controls remain defined for all $37$ trajectories in every draw; judge retained
counts range from $30$ to $32$, with $0.320\%$ of trajectory-draws constant after masking.
The multiset-change counts are $70{,}964/74{,}000$ and $49{,}183/64{,}000$;
denominators are $37\times2{,}000$ and $32\times2{,}000$ trajectory--permutation draws.
All $205$ undefined judge controls are constant after masking, with no missing-score cases;
these are draws within the fixed eligible set, not newly excluded trajectories.
Missing and constant controls stay undefined, without zero filling or resampling.
Any full-domain paired range therefore describes its varying defined population, not a
fixed-population paired test. The kit contains every repeated correlation, multiset/tie
change count, exact membership, seeds, and conditional-mean Monte Carlo checks.
The old sign-flip remains the historical test; it is not retroactively reinterpreted as
the new conditional randomization test.

\paragraph{Synthetic scope and detector.}
The seven prespecified scenarios use the full Qwen design and observed completeness masks,
conditional on a fixed sampled action set. Known valid probabilities generate independent
Bernoulli outcomes at $3+4\times3$ replicas; score emulators depend on latent quantities and
independent score noise, never on the same sampled reference noise. Continuous/clipped and
three-level scores share each generated reference. The null has nonconstant latent
contribution and independent scores; weak/moderate alignment and a strong latent-score oracle
are separate settings. A no-action-effect stress has constant zero latent contribution and
undefined latent Spearman, not correlation zero. An additional heterogeneous-baseline stress
is evaluated at null and moderate alignment. The exact generator, ranges and seeds are in
the anonymous plan/configuration; no observed three-rollout rate is treated as latent truth.
The old detector validated here is specifically the original two-sided paired sign-flip
with $10{,}000$ signs, not the entire historical H3 overlap rule or training pipeline.
The new positive CRT uses $999$ permutations per dataset and a fixed two-emulator Holm family;
undefined endpoints keep their hypothesis slot. Both use the actual eligibility rules.
Latent associations are medians on explicitly identified full and selected supports.
Reported rate intervals are Wilson Monte Carlo intervals, not uncertainty over ALFWorld tasks.
Restoration, action acquisition and missing-action selection are not validated.

\begin{table}[!t]
\centering\small
\caption{\textbf{Known-probability validation, 1,000 datasets per scenario.} Rejections of each named detector after Holm over the two emulators; brackets are Wilson95 Monte Carlo intervals. Latent $\rho$ is the median across datasets of the selected within-dataset trajectory median for the new detector. No-effect has undefined latent Spearman. Full/old-selected targets, selection distributions, raw $p$ and undefined counts are in the numeric kit.}
\label{tab:r19validation}
\setlength{\tabcolsep}{4pt}
\begin{tabular}{llccc}
\toprule
Scenario & Score & New $k/n$ [MC95] & Old $k/n$ [MC95] & Latent $\rho$ \\
\midrule
Null & Clipped & $30/1000$ $[0.021,0.043]$ & $22/1000$ $[0.015,0.033]$ & +0.002 \\
Null & 3-level & $27/1000$ $[0.019,0.039]$ & $22/1000$ $[0.015,0.033]$ & +0.002 \\
Weak & Clipped & $924/1000$ $[0.906,0.939]$ & $614/1000$ $[0.583,0.644]$ & +0.426 \\
Weak & 3-level & $898/1000$ $[0.878,0.915]$ & $557/1000$ $[0.526,0.588]$ & +0.397 \\
Moderate & Clipped & $1000/1000$ $[0.996,1.000]$ & $994/1000$ $[0.987,0.997]$ & +0.798 \\
Moderate & 3-level & $1000/1000$ $[0.996,1.000]$ & $989/1000$ $[0.980,0.994]$ & +0.751 \\
Oracle & Clipped & $1000/1000$ $[0.996,1.000]$ & $1000/1000$ $[0.996,1.000]$ & +0.999 \\
Oracle & 3-level & $1000/1000$ $[0.996,1.000]$ & $1000/1000$ $[0.996,1.000]$ & +0.935 \\
No effect & Clipped & $34/1000$ $[0.024,0.047]$ & $25/1000$ $[0.017,0.037]$ & undefined \\
No effect & 3-level & $42/1000$ $[0.031,0.056]$ & $19/1000$ $[0.012,0.029]$ & undefined \\
Stress null & Clipped & $28/1000$ $[0.019,0.040]$ & $19/1000$ $[0.012,0.029]$ & -0.000 \\
Stress null & 3-level & $31/1000$ $[0.022,0.044]$ & $18/1000$ $[0.011,0.028]$ & +0.000 \\
Stress moderate & Clipped & $984/1000$ $[0.974,0.990]$ & $731/1000$ $[0.703,0.758]$ & +0.798 \\
Stress moderate & 3-level & $964/1000$ $[0.951,0.974]$ & $692/1000$ $[0.663,0.720]$ & +0.751 \\
\bottomrule
\end{tabular}
\end{table}

\paragraph{Joint false-positive rate from saved outputs.}
For each association-null scenario we align the two score decisions by scenario, dataset ID
and seed, and count the union of their Holm rejections. This is not the sum or maximum of
marginal rejection rates: the null overlap counts are $18/14$ for fixed-mask/archived tests,
and the stress-null counts are $17/12$. The denominator remains $1{,}000$ datasets.
Undefined endpoints would retain their nonrejecting hypothesis slots; none occur in these
four cells. No-action-effect datasets are excluded from this association-null accounting.
The numeric kit verifies all saved raw/Holm pairs and supplies the rejection IDs.
The reported joint rates reaggregate existing outputs without additional simulations.
\begin{table}[!t]
\centering\small
\caption{\textbf{Joint family-wise false-positive rates from saved paired decisions.} A dataset is counted once when either score rejects after Holm over two scores. Brackets are Wilson $95\%$ Monte Carlo intervals. No-effect scenarios are not association-null calibrations. All four cells have 1,000 aligned datasets and no undefined endpoints.}
\label{tab:r20fwer}
\begin{tabular}{llrrc}
\toprule
Scenario & Detector & Both reject & Either rejects & Rate [MC95] \\
\midrule
Null & Fixed-mask & 18 & $39/1000$ & $0.039$ $[0.029,0.053]$ \\
Null & Archived & 14 & $30/1000$ & $0.030$ $[0.021,0.043]$ \\
Stress null & Fixed-mask & 17 & $42/1000$ & $0.042$ $[0.031,0.056]$ \\
Stress null & Archived & 12 & $25/1000$ & $0.025$ $[0.017,0.037]$ \\
\bottomrule
\end{tabular}
\end{table}

\section{The substrate rescue, and why it closed}
\label{app:substrate}
The substrate explanation for the KS2 training nondetection was examined in a separate rescue
experiment. With three diagnosed substrate defects repaired, outcome-only training diverged
at the original hyperparameters, at reduced learning rate ($5\mathrm{e}{-}5 \to 2\mathrm{e}{-}5$),
and at five-fold tighter gradient clipping ($1.0 \to 0.2$). In each configuration, the pre-clip
gradient norm escaped one to several batches before KL breached its guardrail, despite
per-step norm clipping. These observations describe failure of the tested configurations;
they do not identify its cause or establish a structural incompatibility of the method.
No configuration reached evaluation, and the line closed under its pre-specified stopping
criterion. Consequently there is no trained corrected-substrate comparator: the substrate
explanation remains unresolved, neither confirmed nor excluded. These failures cannot turn
the KS2 nondetection into evidence against that explanation.

\section{Practitioner-facing artifacts, in full}
\label{app:artifacts}
\paragraph{Additional reporting practices.} Four named, measured instruments
operationalizing fields of the survey's CA-ID card~\citep{survey2604}: dose matching, measured
perturbation strength for controls, an MDE ladder for training nulls, and a four-dimension
integrity taxonomy and its incident catalogue; concurrent work argues for a matched-concentration
control without replay~\citep{coverage2609}.

The audit's diagnostics translate into named, practitioner-facing artifacts, each of which
exists because one of our own comparisons would have misled us without it: \textbf{dose matching
before any credit comparison} (any update step that drops zero-weight examples lets the credit
rule silently set the training dose --- our arms' optimizer steps ranged from $112$ to $8$ per
round under identical budgets); \textbf{measured perturbation strength for every control} (a
control named ``shuffle'' is not automatically an equal-strength manipulation, and unmeasured it
invalidated our own first family dissociation); \textbf{a routing rule that knows what it is
for} (a cost mechanism, not a detector --- full treatment in
Section~\ref{sec:training}); and \textbf{reading a training-loop null} (every equivalence claim
travels with the registered margin, $\pm 3$\,pp, \emph{and} the minimum detectable effect of
the test actually run --- seed-paired here, one-sided $7.8$--$36.4$\,pp at the realized
variances, the single-seed task-level ${\approx}11.8$\,pp being a different quantity
(Appendix~\ref{app:mde}) --- a nondetection is not an effect-size bound: the MDE describes
the design's detection properties, while effects excluded by the data must be established
by the corresponding interval or equivalence test). The dose-matching specification and its
reading rule are in Appendix~\ref{app:protocol}; the full MDE ladder is in
Appendix~\ref{app:mde}.

\begin{table}[h]
\centering
\small\setlength{\tabcolsep}{4pt}
\caption{\textbf{The audit's results as executable rules.} Each row restates a result of the
named section; no new quantity is introduced here.}
\label{tab:discussion}
\begin{tabular}{@{}p{0.24\linewidth}p{0.24\linewidth}p{0.22\linewidth}p{0.24\linewidth}@{}}
\toprule
to claim & control & report & cannot conclude \\
\midrule
a credit signal aligns with sampled contribution & declared score domain, replay mask and
control order & paired increment, control-sensitivity range and conditional test
(\S\ref{sec:audit}) & noiseless fidelity or causal validity from conditional alignment \\
step-credit \emph{content} \mbox{improves} training & effective optimization dose; seed-paired design
with its MDE & seed-paired effects, intervals, and the MDE of the test run
(\S\ref{sec:training}) & content effects where dose is confounded or the MDE exceeds plausible
effects \\
causal sparsity from a nonzero-label count & the design's counting-structure null & observed
vs.\ forced label counts (\S\ref{sec:map}) & sparsity, when the count matches what the design
forces \\
a reliability correction is valid & a pre-registered simulation with the estimand fully
specified & coverage against the specified target (\S\ref{sec:audit},
Appendix~\ref{app:simval}) & validity, when the target itself is under-specified --- ours is
retracted to a sensitivity analysis \\
\bottomrule
\end{tabular}
\end{table}

\paragraph{Two notes on the routing rule.}
The two $13.1$\% figures coincide by definition: the
routing threshold was set so the routed fraction equals the no-counterfactual rate by
construction (\texttt{ROUTE\_NUMBERS.md:8--9}). That $13.1$\% is the fraction of turns routed to
the judge --- $231$ of $1{,}768$ --- and not a reduction in judge calls; the reduction is the
per-trajectory figure, $43$ of $50$ baseline calls remaining. The generating script names the
routed fraction \texttt{cost\_reduction\_per\_turn}; despite that field name, routed fraction and
call reduction are different quantities. The rule's safety has two layers, only one by construction: on high-confidence steps the
policy-gradient signal is inert ($\nabla\log\pi \approx 0$); the no-outcome-variation half is an
audit finding --- $\sigma_{\mathrm{floor}}$ is replay-derived, invisible to the router ex ante.

\paragraph{The routing statistic, retrospective and in full.}
A confidence-only router --- specified in advance as a replay-free decision rule, its inputs
free at inference time (exploratory, pre-registered as such) --- recovers $11.9$\% of
nonzero-contrast turns (Wilson 95\% $[9.4, 14.9]$; $n{=}540$) against the $13.1$\% of all
turns it routes, which is chance-level recall of that label; and the label it recovers is the
resolution-bounded nonzero-contrast label of Section~\ref{sec:method}, not a certified set of
causally pivotal steps. As a cost mechanism the same threshold routes $13.1$\% of turns to the
judge and cuts judge calls by $10.4$\% per trajectory --- a routed fraction and a reduction,
two different quantities, and neither may be read alone. Prospective deployment of the router
is not evaluated; every number here is retrospective, computed on the archived turns.

The poor recall is descriptively consistent with the high-confidence stuck steps in the
Section~\ref{sec:mechanism} case, but that case does not identify the mechanism of routing
failure or characterize all nonzero-contrast turns. We release
the rule with its negative result attached --- a rule that travels without its evaluation will be
used for the thing it cannot do (sweep and cost account: Appendix~\ref{app:supfigs},
Figure~\ref{fig:s3}). That a distribution-level entropy separation of critical states
(\citealp{carl2512}) and chance-level recall at a cost-matched threshold are not in conflict is
shown in Appendix~\ref{app:concurrent}.

\section{Release inventory}
\label{app:release}
\paragraph{Sources of the \S\ref{sec:mechanism} case table (Table~\ref{tab:seed017}).}
Actions and environment replies: \texttt{trajectories/seed017.jsonl} (sha256
\texttt{dccef883\ldots a100}); policy log-probabilities: \texttt{seed017.pol.json}
(\texttt{f1301dfb\ldots 6073}); implicit credit: \texttt{seed017.rho.json}
(\texttt{a38c977a\ldots 168d}); replay reference: \texttt{seed017/replay\_summary.json} and the
turn directories \texttt{t012}, \texttt{t016}, \texttt{t020} of the sealed replay archive. In
the trajectory dump, the observation stored with record $t$ answers action $t{-}1$; the reply
to action $t$ is the observation of record $t{+}1$ --- the case table prints replies under
that mapping.

\paragraph{One trajectory, in columns.}
Trajectory \texttt{seed017} shows what these aggregates look like on the ground
(Table~\ref{tab:seed017}). The policy returns to the same action --- \emph{take cloth 1 from
toilet 1} --- four times across a take--move--examine loop, at turns 7, 12, 16, and 20; at the
three audited takes its confidence is at least $0.9998$, the environment answers each take with
a successful pick-up, and the replay contrasts are $+0.333$, $-0.167$, and $+0.250$ ---
high-likelihood repetitions whose contrasts are coarse, noisy estimates of both signs, while the
credit column is highest exactly where likelihood is highest. The window also shows the
converse: turn 19, an \emph{examine} issued at confidence $0.114$, carries the same $+0.333$
contrast as the highest-confidence take. This illustrates that high action likelihood and
the observed replay ranking can diverge; one trajectory cannot identify a mechanism.

\begin{table}[t]
\centering
\caption{\textbf{Trajectory \texttt{seed017}, the audited window: credit tracks the policy's
likelihood of the action; the replay contrasts are coarse, noisy estimates of both signs.} Feedback is the
environment's reply to the printed action, from the trajectory dump; $x/n$ are success counts;
$A_{\mathrm{replay}} = x_{\mathrm{fact}}/3 - x_{\mathrm{alt}}/12$. Sources and hashes:
Appendix~\ref{app:release}.}
\label{tab:seed017}
\vspace{0pt}
\small\setlength{\tabcolsep}{2.5pt}
\begin{tabular}{@{}rp{0.17\linewidth}p{0.20\linewidth}rrccr@{}}
\toprule
turn & action & environment reply & $\ell_t$ & $\rho_t$ & fact. & alt. & $\widehat A_t$ \\
\midrule
11 & examine cloth 1 & ``Nothing happens.'' & $-2.732$ & $0.814$ & $0/3$ & $0/12$ & $0.000$ \\
12 & take cloth 1 from toilet 1 & ``You pick up the cloth 1 \ldots'' & $-0.0001$ & $1.142$ & $1/3$ & $0/12$ & $+0.333$ \\
16 & take cloth 1 from toilet 1 & ``You pick up the cloth 1 \ldots'' & $-0.0001$ & $1.142$ & $0/3$ & $2/12$ & $-0.167$ \\
19 & examine toilet 1 & ``On the toilet 1, you see \ldots'' & $-2.176$ & $1.053$ & $1/3$ & $0/12$ & $+0.333$ \\
20 & take cloth 1 from toilet 1 & ``You pick up the cloth 1 \ldots'' & $-0.0001$ & $1.142$ & $1/3$ & $1/12$ & $+0.250$ \\
21 & go to countertop 1 & ``You arrive at countertop 1 \ldots'' & $-3.828$ & $0.800$ & $0/3$ & $1/12$ & $-0.083$ \\
\bottomrule
\end{tabular}
\end{table}

Historical release inventory (counts from the sealed final-state record; numerical inputs accompany this candidate and the larger archive is staged for de-anonymization):
\begin{center}
\begin{tabular}{lrr}
  \toprule
  Item & Count & Size \\
  \midrule
  KS1 \texttt{collect\_v2} replay archive & 29{,}402 files & 4.7\,G \\
  per-trajectory \texttt{replay\_summary.json} & 1{,}026 & --- \\
  per-turn \texttt{summary.json} & 17{,}682 & --- \\
  credit files \texttt{*.credit.json} & 6{,}058 & --- \\
  $\rho$ files / judge files & 2{,}275 / 1{,}596 & --- \\
  \texttt{eval.json} & 118 & --- \\
  adapters & 87 & 13.1\,G \\
  logprob dumps (\texttt{.pol}/\texttt{.hind}) & 4{,}601 & --- \\
  500-state bank (\texttt{a3b\_state\_bank.jsonl}) & 1 & 0.8\,M \\
  \midrule
  \texttt{resolution\_curve\_summary.json} (App.~\ref{app:resolution}) & 1 & --- \\
  its generator and the two renderers (\texttt{*.py}) & 3 & --- \\
  \bottomrule
\end{tabular}
\end{center}

\noindent The resolution summary and the three scripts that produce and render it are released
with the archive, so Table~\ref{tab:resolution} and Figure~\ref{fig:resolution} can be rebuilt
rather than taken on trust. One quantity in Section~\ref{sec:audit} does not sit in this
inventory: the reliability correction's own output, \texttt{A2\_step2\_output.txt}, lives on the
archived analysis commit \texttt{3e00d92} with its sha256 recorded in
Table~\ref{tab:step2}'s caption. It is the artifact behind the corrected bound, and we name its
location here rather than leave it implicit.

\section{Analysis plans, prompts and diagnostic records}
\label{app:stubs}
\paragraph{Source and rendering conventions.}
Original analysis plans and diagnostic records accompany this PDF with source
hashes; release hashes separately verify anonymized copies. Prompt files below
preserve their frozen content, with recorded glyph transliterations for the PDF
engine. The original files remain authoritative. Chinese-language records are
preserved in the supplement; their scientific content is summarized in English
where needed. Author and repository identifiers are redacted for double-blind
review. The numerical kit retains source plans and incident narratives with their hashes. Broader original context remains in the versioned manuscript and archival repository staged for de-anonymization; historical interpretations in those records are not adopted as current claims.

\subsection{Pre-registrations: index and current-version map}
Pre-registrations accompany this PDF in the supplement. Original versions are identified by
\texttt{SHA256SUMS.txt} and the hashes below; anonymous redactions are recorded in
\texttt{REDACTION\_NOTE.md}, and \texttt{SHA256SUMS\_RELEASE.txt} verifies shipped bytes.
Amendments are append-only; the \emph{current version} column names the version governing
these analyses. This index identifies the historical plans verified against the
canonical numerical kit; the independent-bank experiment is separately documented
in Appendix~\ref{app:postexperiment}. Display aliases map to archived names and hashes
in \texttt{PREREG\_ALIASES.json}.

\begin{center}
\small\setlength{\tabcolsep}{3pt}
\begin{tabular}{@{}p{0.29\linewidth}p{0.24\linewidth}p{0.25\linewidth}p{0.17\linewidth}@{}}
\toprule
document & governs & current version / notes & sha256 (12) \\
\midrule
KS1 audit preregistration & KS1 audit: T1--T3, controls, decision order & original, tag-signed & \texttt{9b90cfd081cf} \\
KS1\_XFAM\_PREREG.md & cross-family replication (sole primary: T3) & original & \texttt{9972cb3710ff} \\
KS2\_PREREG.md & seven-arm diagnostic: C1--C6, $\pm3$\,pp, Holm & original; contains no exclusion rule; 5-seed adaptive rule never triggered & \texttt{d85e5ec31680} \\
PREREG\_reliability\_stage0.md & reliability correction, Phase A & v3, tag \texttt{prereg-\allowbreak reliability-\allowbreak stage0-v3} & \texttt{24a45c965c99} \\
PREREG\_\allowbreak correction\_\allowbreak validation\_\allowbreak 2026-09-11.md & the validation that retracted the corrected bound & original $+$ 2 pre-run amendments; \S6 rule untouched by either & \texttt{f3a295361f8e} \\
PREREG\_\allowbreak paired\_\allowbreak controls\_\allowbreak 2026-09-13.md & the paired-increment control test (\S\ref{sec:audit}) & original, committed before the statistic was computed & \texttt{963dec974a99} \\
A3B\_PREREG.md ($+\Delta$, amendment, addenda 3--4) & checkpoint-divergence family (exploratory); owns the arm-5 seed-2 exclusion & original $+$ 4 append-only companions (hashes in the manifest) & \texttt{b654adf30ae0} \\
R\_ROUTE\_PREREG.md & confidence router (exploratory) & original, tag \texttt{1814166\ldots} & \texttt{1814166559b4} \\
\bottomrule
\end{tabular}
\end{center}

\paragraph{R\_ROUTE (the confidence router).} Chinese-major source; byte-exact in the
supplementary pack.
Structure, for the reader: frozen under tag \texttt{1814166\ldots}; registered as
EXPLORATORY; the single main result is threshold multiplier $1.0$, with the six-multiplier
grid registered as sensitivity only; recall and both cost granularities are required to be
reported together.
\paragraph{PC3 (substrate-repair experiment).}
No standalone PC3 pre-registration is available in the archived release.
The stopping criterion (S6: same-type divergence implies no further
configurations) is recorded in \texttt{PC3\_VERDICT.md}, preserved in the
supplement. Appendix~\ref{app:pc3table} summarizes its observations and current
claim boundaries; Appendix~\ref{app:substrate} gives the scientific interpretation.

\subsection{The four frozen prompt files}
The first file below is the HCAPO ALFWorld agent template, extracted verbatim from that
paper's own appendix (Appendix C.2.1), not reconstructed, and printed first because it is the
template the collection and scoring pipelines loaded at run time
(\texttt{data/hcapo\_alfworld\_template.txt}, loader
\texttt{scripts/collect\_trajectories.py}; hindsight scoring appends a single
outcome-injection line to it and the base template is untouched,
Appendix~\ref{app:instrument}). Two copies of the template are retained for distinct roles: the
annotated working copy printed here, whose header documents a dangling section pointer in the
original extraction note, and the byte-exact bundle extraction in \texttt{supplementary/}.
The judge audit used the judge prompt exactly as printed second below, fixed before any
scoring.
\paragraph{HCAPO ALFWorld agent template --- extracted verbatim from the paper's appendix, not
a reconstruction; the run-time template.}
\wrapverb{appendix_sources/hcapo_alfworld_template.txt}

\paragraph{Judge prompt v1 --- the prompt the judge audit ran.}
\wrapverb{appendix_sources/judge_credit_v1.txt}

\paragraph{Judge rubric v1.} The 5-point rubric variant is the exploratory
rubric-format prompt (PC\_PREREG Step 3); it was never used in any reported number and ships for
provenance completeness only.
\wrapverb{appendix_sources/judge_credit_rubric_v1.txt}

\paragraph{TARL judge prompt, original.}
\wrapverb{appendix_sources/tarl_judge_prompt_original.txt}

\subsection{PC/PC2 diagnostics}
The registered diagnostics are specified in the $\Delta$ addenda above:
perturbation strength, position monotonicity and the 500-state bank.
Measured outputs appear in Section~\ref{sec:artifacts} and Figure~\ref{fig:s1};
the full versioned experiment records accompany the replay archive.

\subsection{Substrate-repair observations and claim boundaries}
\label{app:pc3table}
With the G=4 correction, three configurations of the v1.4 stabiliser were tested:
the original settings, reduced learning rate ($5\times10^{-5}\to2\times10^{-5}$),
and tighter gradient clipping ($1.0\to0.2$). All diverged before evaluation.
The recorded precursor was a pre-clip gradient spike one to two batches before
the KL breach; the warning fired in all three configurations.
Thus no held-out outcome-only effect is available from these runs.

The earlier weak-signal G=1 run completed $63$ updates without divergence.
The corrected configurations improved during their initial training phase;
the clipping variant's intermediate win rate reached $0.604$, with declining
loss and smooth KL before failure. Clipping was active throughout; the tighter
variant recorded clip-fraction $1.00$. These observations do not identify the
cause of divergence, show that outcome-only credit is ineffective, or establish
that the G=4 correction is harmful. In particular, the original directional-drift
and signal--stabiliser-mismatch diagnoses are historical hypotheses, not identified
mechanisms. The complete original interpretation remains in the archival records.

\subsection{Archived decision records and scope limits}
\label{app:verbatim}
The separate archival supplement preserves the original decision entries,
incident narratives, editorial placement instructions, identifier corrections,
and checker postmortems, with their source identities. These records document
how the analysis developed; they introduce no additional scientific endpoint.
The PDF retains the scientific consequences below and in
Appendices~\ref{app:integrity} and~\ref{app:gatemiss}.

\paragraph{Why a matched-update comparison was not run.}
The archived decision cited the partial-correlation result, a design limitation,
and cost. Its numerical motivation was the instrument-family partial correlation
$+0.078$ ($p=0.774$), with positive-control association $+0.912$ ($p=0.0001$).
The proposed $\lVert\Delta W\rVert$ overlap was across different training rounds,
so caliper matching would confound round and training volume. The estimated cost
was $\$25$--$30$ (about $3$ GPU-hours).
The comparison remains unrun. The archival assertion that partial correlation
already answered the causal question is not adopted: a significant marginal
association and nonsignificant conditional association do not establish mediation.
The recorded sensitivity associations remain $r=+0.9250$ ($p=0.0004$) after
dropping arm 6 and $r=+0.9274$ ($p=0.0013$) after dropping arms 6 and 1;
neither supplies the missing matched design.
The unchanged record is \texttt{verbatim\_records/02\_R3\_matched\_dW\_ruling.md}
in \texttt{ICLR\_ARCHIVAL\_RECORDS.tgz}.

\paragraph{Partial-coverage analysis and subsequent claim review.}
Collection stopped with insufficient complete trajectories under the recorded
stop-time interpretation. Analysis was then allowed with explicit partial-coverage
labels; the affected claim text was cleared in a second, later review on the same
day. These are separate events. The executed inclusion rule retained $26$ of $28$
trajectories, correcting the earlier expectation that all $28$ would remain.
All coverage counts, nonrandom missingness, the ineffective imbalance flag and
cost overrun remain in Appendix~\ref{app:integrity}; the threshold discrepancy is
in Appendix~\ref{app:gatemiss}. The complete original record is
\texttt{verbatim\_records/03\_S3\_gate\_withheld\_released\_ruling\_and\_diff.md}
in the archival supplement. The Llama fidelity arm remains descriptive, with no
confirmatory verdict under its amendment.

\section{Concurrent work, reconciled}
\label{app:concurrent}

Three concurrent papers touch this paper's objects. The main text carries one-sentence positions
(\S\S1--3, 6); the full reconciliations are recorded here.

A similar sparsity pattern is reported by CARL; ours differs in kind: zeros are
noise-floored and resolution-bounded, and measurability itself is bidirectionally
model-dependent.

\paragraph{CARL (multi-hop search).} CARL reports that entropy separates critical from
non-critical states at the distribution level (Cliff's $\delta = 0.42$, $n = 294$,
Brunner--Munzel), in multi-hop search, with criticality estimated from resampled continuations.
Their criticality signal does not enter training through a classification threshold. It enters
as an action-density term --- expansion count per unit entropy at a state --- with the update set
restricted to states that have more than one child; the operating point at which their method
works is therefore an emergent property of the rollout budget, not a stated rule whose recall was
measured. We record the absence explicitly, because it is the substance of the difference: the
paper reports no precision, recall, F1, or AUC for criticality identification, and no threshold,
quantile, or top-$k$ selection rule. Two independent passes over the v3 text were made, over the
sections named in the provenance record; the absence is a finding of those passes, not an
inference from silence in a summary. Nothing in our data speaks to multi-hop search, and nothing
in theirs measures a router's recall; the designs answer different questions. We reconcile
against v3 (2026-05); the bibliography entry carries the v1 date. The revision, not the first
posting, is what ``concurrent'' refers to here.

\paragraph{CAR (formalization).} CAR formalizes per-step causal effect measurement as
do-operations with re-execution and validates the estimator on synthetic SCMs with planted
effects --- ground truth known by construction. That validation and this audit address
complementary questions: estimator recovery of planted effects, and the precision and
reproducibility of finite measurements in a live environment without planted step-level truth.
We audit deployed scores against a sampled intervention contrast and evaluate their recovery
decisions independently; neither replay bank becomes oracle truth. CAR places real tools
with side effects ``out of scope'' (\S7). Our evidence remains specific to the fixed policy,
action support and ALFWorld cohort rather than validating all live-tool interventions.

\paragraph{CSO.} The phrase ``policy reachability'' appears in CSO's abstract \citep{cso2602}; the method section
describes the requirement differently, as branches that remain within the policy's capability. We
quote the abstract and say so, rather than attributing to their method a term their method does
not use. Our coverage measurement is narrower: under our fixed sampling protocol, four
distinct admissible alternatives were not obtained at 13.1\% of intervened turns for Qwen
and 26.8\% for Llama (\S\ref{sec:map}). These budget-dependent acquisition failures do not
identify whether the policy could support a counterfactual under another protocol, and are
not a verification rate for CSO's capability requirement.

\paragraph{C3.} C3 defines replay-based credit diagnostics for cooperative LLM protocols
\citep[\S4.1]{c3exact2603}: credit fidelity compares assigned advantages with a reference
advantage under a fixed continuation distribution. Its experiments include C3, MAPPO and
MAGRPO on math and code tasks. It provides a relevant ranking audit. Our audited objects
are adapted implicit and judge scores in single-agent stateful ALFWorld. We examine the
scoring, evaluation and permutation domains, alongside coverage and finite-replica resolution.

The credit-assignment survey~\citep[\S2.3--2.4]{survey2604} is the nearest methodological neighbour: its replay hygiene and replica noise floor are the constructs our replay method uses, its Proposition~1 is the identification argument the estimand rests on, and its Proposition~2 (sign unidentifiability) bounds a claim this paper does not make; none of the three audits a deployed signal against executed replay.

\subsection{Source integrity}
The twenty files in \texttt{SHA256SUMS.txt} retain their original-source hash identities;
per-item hashes in this appendix's \LaTeX{} comments refer to the same originals.
The supplement differs only by the recorded identity redactions, verified separately in
\texttt{SHA256SUMS\_RELEASE.txt}. Glyph transliteration applies only to the PDF rendering.

\section{Instrument corrections, data integrity and missingness}
\label{app:integrity}

\paragraph{Raw correlations behind Table~\ref{tab:ks1verdict}.}
Figure~\ref{fig:verdict} shows both the archived paired increments and observed alignment
for Qwen, with their distinct statistical comparisons (Appendices~\ref{app:paired}
and~\ref{app:r19audit}); it has no Llama row. The original paired fields derive from
\texttt{analysis\_summary.json} and \texttt{paired\_controls\_test.py}.
The raw results are Qwen2.5-7B implicit $0.0193$ $[-0.109, 0.081]$ ($n_{\mathrm{eff}}{=}37$ of 50;
own control $[0.005, 0.114]$) and judge $0.1142$ $[0.027, 0.168]$ ($n{=}37$, 32 computable; own
control $[-0.049, 0.117]$), data \texttt{runs/collect\_v2/analysis\_summary.json};
Llama-3.1-8B implicit under the corrected instrument $-0.043$ $[-0.125, -0.016]$ ($n{=}21$
trajectories retained of 28; own control $[-0.102, +0.024]$), data
\texttt{runs/xfam\_ext/redo/redo\_family\_analysis.json}, coverage
88.3\%\textsuperscript{\ref{fn:coverage}}.

\paragraph{Estimator, controls, instruments and data for the archived raw comparison (Table~\ref{tab:ks1verdict}).} Estimator: median
within-trajectory Spearman vs.\ $A_{\mathrm{replay}}$, 10k bootstrap; controls: full-score-domain-matched
within-trajectory shuffles followed by replay masks; instruments: original (Qwen), corrected (Llama; coverage
88.3\%\textsuperscript{\ref{fn:coverage}}); data: \texttt{runs/collect\_v2/analysis\_summary.json},
\texttt{runs/xfam\_ext/redo/redo\_family\_analysis.json}. Sign agreement: per-step vs
trajectory-median splits, Wilson 95\%; implicit $267/527 = 50.7\%$ $[46.4, 54.9]$, judge
$84/139 = 60.4\%$ $[52.1, 68.2]$; $n_{\mathrm{eff}}$ = trajectories retained by the frozen
exclusion rules.

This appendix records instrument corrections, data availability and analysis
limitations. Complete historical process records are preserved separately in
\texttt{ICLR\_ARCHIVAL\_RECORDS.tgz}.
\paragraph{The determinism verification, as executed.}
One trajectory hash, four independent environments: verified before collection and
re-verified on every host that touched the data.

\subsection{Semantic instrument correction and validation limits}
\label{sec:integrity}
A metadata inconsistency revealed a chat-template error that byte-identity checks
had missed. A static trace and A/B comparison preceded a full re-replay under the
corrected instrument: the map-rate change was $12.0$ pp predicted and $12.3$ pp
measured (\texttt{REDO\_REPORT.md:19}). No Qwen verdict class changed; the
cross-family arm remains descriptive and receives no confirmatory verdict.
The measurability rates nevertheless moved materially.

All affected replicas had matching hashes. The problem was semantic instrument
identity, including the tokenizer and template, rather than different file bytes.
The re-replay also failed its recorded completeness threshold. Partial-coverage
analysis proceeded only through an explicit exception, followed by separate
review of the affected claims (Appendix~\ref{app:gatemiss}). The counts and
missingness are retained below. Complete original incident narratives are in
\texttt{ICLR\_ARCHIVAL\_RECORDS.tgz}.

\paragraph{Validation limits.}
The full catalogue of integrity checks remains in
\texttt{supplementary/integrity\_taxonomy.md}, verified by the numerical kit's
checksum manifest. Two checks had known-positive validation; two held by design
and had not been tested. The template correction shows why byte identity alone
does not establish semantic instrument identity: tokenizer and chat-template
behavior must also agree.

\paragraph{The Llama replication, and what it was run for.}
The Llama replication was run twice, the second time in full under the corrected instrument
(Appendix~\ref{sec:integrity}).
Under the corrected instrument the same descriptive pattern holds: the family estimate and its
own full-score-shuffle control overlap (Appendix~\ref{app:r29}, which carries both
raw intervals). Overlap is not a test of equality. The arm is replication-descriptive and is issued no verdict, as its own amendment
requires (Appendix~\ref{app:verbatim}).

\paragraph{Comparing the two instrument generations.}
The template repair moved the measurability map materially
(Appendix~\ref{sec:integrity}). The Llama fidelity interval overlaps its own
shuffled control under both templates, at pilot power and at $n{=}21$;
sign agreement remains consistent with chance, and the transfer criterion holds
in both. These descriptive observations do not establish equivalence between
instrument generations or validate either as latent truth.

\paragraph{The transfer criterion, with its margins.} T3
replicates across model families under the corrected instrument: on the registered primary (full)
set, the cross-family median ($+0.7008$ $[0.6482, 0.7683]$, $n{=}27$, coverage
88.3\%\textsuperscript{\ref{fn:coverage}}) satisfies the pre-registered point criterion --- it
lies within KS1's registered CI $[0.647, 0.793]$. The corrected interval's lower edge, $0.6482$, clears the
registered band's exact lower bound ($0.6474$) by $0.0008$ --- the cross-family replication holds
on both the point criterion and the full interval, with essentially zero margin, which we state
proactively. The second, independent
criterion also holds: re-run under a script frozen before execution, the shuffled control for T3
is $-0.0155$ ($[-0.1290, +0.0844]$, $n{=}27$), non-overlapping with the observed interval, so the
control criterion fires; we note its shuffling procedure executes in a remote frozen script and
its control value is read from the archived artifact rather than recomputed locally. And the
full-set family difference, re-run by the recovered original bootstrap on the corrected inputs,
spans zero for 40/40 seeds (seed-11 primary $-0.0188$ $[-0.1178, +0.0954]$;
$n_{\mathrm{xfam}}{=}27$ vs $n_{\mathrm{KS1}}{=}47$)\textsuperscript{\ref{fn:coverage}} --- the
families do not measurably differ on the statistic whose replication the criterion certifies.

\paragraph{Additional validation failures.}
Some internal checks initially tested a release against its own manifest rather
than its source, or confused a configuration failure with an untraceable result.
The full records and checker postmortems are preserved in the archival supplement;
they are not additional empirical contributions of the step-credit audit.

\paragraph{Archive readability.} An inventory check
passes at 29{,}402 files discovered against 29{,}402 expected while one file is unreadable
[\texttt{ALL\_EXPERIMENTS\_RECORD.md:2037}]; the archive verification run records
29{,}401 match / 0 mismatch / 0 missing / 1 named unreadable [
\texttt{ALL\_EXPERIMENTS\_RECORD.md:2206-2207}]. The derived summary survives, so the number
stays traceable while its re-derivation is lost; the verified tarball is accordingly the
\emph{primary} copy of the replay evidence and the volume secondary.

\paragraph{Partial coverage, corrected inclusion count and cost.} The gate paragraph of
Appendix~\ref{sec:integrity} has its supporting counts here: this run covered
$1{,}082/1{,}225 = 88.3\%$ of parseable turns [\texttt{REDO\_STATUS.md:20}]; the
original XFAM-v1 analysis had itself proceeded at $1{,}129/1{,}225 = 92.1\%$ [
\texttt{REDO\_STATUS.md:41}]. The stop-time claim that the
frozen v3.2 rule ``would retain all 28'' trajectories [\texttt{REDO\_STATUS.md:38--41}]
was corrected by the executed analysis, which retained 26 of 28 [
\texttt{REDO\_REPORT.md:27}]; the record's claim overstates by two. The release stands either way,
since 26 clears the 20-trajectory inclusion count used for the exception; this does
not mean the original full-coverage stopping criterion passed. There were two releases, not one, on the same day and
in sequence: the exception allowed the \emph{analysis}; a separate subsequent
review cleared the affected claim text after its own condition passed. Missingness is
reported with its structure visible: per-bucket coverage is B1 $11/11 = 1.000$ $[0.741, 1.000]$,
B2 $17/17 = 1.000$ $[0.816, 1.000]$, B3 $0.881$ $[0.861, 0.898]$, overall $0.883$, with all 142
missing and 1 unparseable turns falling exclusively in 13 long-bucket (B3) trajectories of length
$\geq 44$ [\texttt{REDO\_REPORT.md:55--70}]. The stated flag criterion --- a bucket is
marked insufficient-coverage when its Wilson 95\% CI excludes the overall rate --- fires for no
bucket; that is not evidence of balanced coverage, because B3 dominates the universe and its CI
contains the overall rate by construction, so the flag is structurally incapable of firing for
the only bucket with missing data. The missing turns sit at the cap-trip end of long trajectories
and are plausibly harder than average; the B3-level numbers inherit that caveat. Context of the
stop: generation halted at the 13 GPU-hour cap
trip with 1{,}083 summaries written, 13.17 GPU-hours metered against a \$40 cap [
\texttt{REDO\_STATUS.md:14,16}].

The archived pilot fidelity estimate was $-0.0800$ $[-0.2086,+0.1424]$,
$n_{\rm eff}=18$; the corrected estimate is $-0.0431$ $[-0.1245,-0.0164]$,
$n_{\rm eff}=21$ (two inclusion-rule and five constant-reference exclusions
out of 28). Its negative raw interval does not replace the comparison with its
own control or the descriptive-only status of this arm.
The missing tail was not rerun: an earlier retry backlog yielded one complete
turn from 101 attempts (about $1\%$). A non-atomic write left one summary
unparseable; the proposed atomic-write change was recorded but not retroactively
applied to the collected data. These engineering facts do not supply scientific
evidence about the missing turns.

\paragraph{Unavailable procedures and checker errors.}
A requested exact rerun could not proceed because the original procedure was not
archived; no replacement was inferred from prose. Superseded values and the
subsequently recovered computations are distinguished in
Appendix~\ref{app:superseded}. Internal comparator errors and their correction
logs remain in the archival records.

\paragraph{A frozen power proxy, inverted.} The KS2 adaptive rule computed power from the
outcome-only arm's between-seed spread, treated as an upper bound on the paired spread; realised
paired spreads ran $7.6$ to $35$ times that proxy --- the anchor arm happened to be the most
stable one, so the bound was inverted, not conservative, and the rule reported full power for a
study underpowered at the registered band. It was executed verbatim and escalated rather than
adapted mid-run: a power calculation frozen before the data is only as conservative as its
proxy, and ours taught us which proxy not to freeze.

The archival records also retain the packaging and typography postmortems.
They do not change the scientific values or the limitations reported here.

\paragraph{Estimator note (Table~\ref{tab:ks1verdict}).} The shuffled control's own interval may
exclude zero without carrying signal: under marginal matching, the median of within-trajectory
rank correlations over small turn counts is not constrained to centre on zero --- which is why
the verdict gate is the family-versus-its-own-control comparison, not either interval's position
relative to zero.

\section{Superseded and disclosed values}
\label{app:superseded}
\paragraph{Historical full-set difference, as published.} The originally published value
$+0.0432$ $[-0.0903, +0.2044]$ has no artifact, no script, and therefore no recorded procedure
and no recorded seed; it is disclosed here and not printed in the main text. The main-text
disclosed re-computation is the seeded one ($+0.0319$ $[-0.0903, +0.2000]$, script in repo, seed 11)
--- a re-runnable computation of the same quantity, not a reproduction of the original run ---
whose script was later shown to regenerate its artifact byte-identically, and whose corrected-%
instrument successor ($-0.0188$ $[-0.1178, +0.0954]$) is what Section~\ref{sec:audit} reports.
That the two historical intervals share a lower bound suggests procedural similarity but does
not constitute proof, and no same-procedure comparison between them is computable.
\paragraph{Common-support medians.} The common-support medians ($+0.7673$ cross-family,
$+0.6757$ KS1) remain disclosed but unrecomputed: the generating procedure was
never archived and cannot be rebuilt, and the cross-family side's inputs predate the instrument
correction. They remain unrecomputed, and no claim rests on them.
\paragraph{The Llama selection check.} The included-minus-excluded log-probability check of
Section~\ref{sec:map} was reported for Qwen only. The historical counts artifact did not
separately record the corrected Llama membership. The archived membership audit recovers the $792$ complete
turn IDs from the corrected replay summaries, but does not recompute a Llama selection
contrast. Historical absence of that result is not a claim that membership remains unrecoverable.
\paragraph{Pre-redo instrument values.} All pre-redo Llama-side quantities (the pre-redo T3
values, the pre-redo measurability rates, and their derivatives) are retained in the ledger
under explicit SUPERSEDED / NOT-citable labels and appear nowhere in this paper's claims, except
where a pre-correction value is quoted, explicitly labeled, solely to demonstrate robustness across
the instrument fix.
\paragraph{Voided residual observation.} The contaminated-era family-residual
observation is void and does not support a claim.

\section{Stopping-rule deviation and two-stage review}
\label{app:gatemiss}

The corrected-instrument re-replay failed the recorded stopping criterion.
The protocol required at least $20$ of $27$ trajectories to be complete before
analysis; the run finished with $15$ of $28$ complete, and analysis was withheld.
An explicit exception subsequently allowed partial-coverage analysis, with every
derived quantity labeled accordingly. A second review later cleared the affected
claim text; permission to analyze did not itself clear those claims.
The counts and nonrandom missingness are in Appendix~\ref{app:integrity}.

The decision record identifies a discrepancy between the task brief's
trajectory-completeness threshold and the frozen pipeline's turn-level inclusion
rule. The stop-time threshold was not itself a validated scientific criterion.
That discrepancy explains the exception but does not make the failed threshold
pass. The original stop and both subsequent review steps remain preserved in the
archival records, so the partial-coverage label cannot conceal the deviation.

\end{document}